%% file: main.tex
\documentclass[10pt,journal,compsoc]{IEEEtran}
\ifCLASSOPTIONcompsoc
  \usepackage[nocompress]{cite}
\else
  \usepackage{cite}
\fi
\ifCLASSINFOpdf
\else
\fi
\usepackage{url}

\usepackage[author={Reviewer}]{pdfcomment} 

\usepackage{amssymb}
\usepackage{graphicx}
\usepackage{xcolor}
\usepackage{amsmath}
\usepackage{siunitx}
\usepackage{algorithm}
\usepackage{algpseudocode}
\usepackage{multirow}
\usepackage{array}
\usepackage{caption}
\usepackage{subcaption}
\usepackage{hhline}

\usepackage{tikz}
\newcommand*\annotatedFigureBoxCustom[8]{\draw[#5,thick,rounded corners] (#1) rectangle (#2);\node at (#4) [fill=#6,thick,shape=circle,draw=#7,inner sep=2pt,font=\sffamily,text=#8] {\textbf{#3}};}
\newcommand*\annotatedFigureBox[4]{\annotatedFigureBoxCustom{#1}{#2}{#3}{#4}{white}{white}{black}{black}}

\newenvironment {annotatedFigure}[1]{\centering\begin{tikzpicture}
\node[anchor=south west,inner sep=0] (image) at (0,0) { #1};\begin{scope}[x={(image.south east)},y={(image.north west)}]}{\end{scope}\end{tikzpicture}}

\newcolumntype{P}[1]{>{\centering\arraybackslash}p{#1}}

\DeclareMathOperator{\simmatch}{sim}

\newcommand{\oldtext}[1]{\ignorespaces}
\newcommand{\newtext}[1]{\textcolor{black}{#1}}

\begin{document}
%
\title{Convolutional Cross-View Pose Estimation}
%
%
%
%

\author{Zimin~Xia,
        Olaf~Booij,
        and~Julian~F.~P.~Kooij,~\IEEEmembership{Member,~IEEE}
\thanks{This work is part of the research programme Efficient Deep Learning (EDL) with project number P16-25, which is (partly) financed by the Dutch Research Council (NWO).}
\IEEEcompsocitemizethanks{
\IEEEcompsocthanksitem J.F.P. Kooij and Z. Xia are with the Intelligent Vehicles Group, Department
of Cognitive Robotics, Delft University of Technology, Leeghwaterstraat, 2628 CN Delft, Netherlands \protect\\
E-mail: {z.xia,j.f.p.kooij}@tudelft.nl
\IEEEcompsocthanksitem O. Booij is affiliated with the Computer Vision Lab, Delft University of Technology, Mekelweg 4, 2628 CD Delft, Netherlands \protect\\
E-mail: olaf.booij@xs4all.nl}
}

%
%

\markboth{Journal of \LaTeX\ Class Files,~Vol.~14, No.~8, August~2015}%
{Shell \MakeLowercase{\textit{et al.}}: Bare Demo of IEEEtran.cls for Computer Society Journals}
%



\IEEEtitleabstractindextext{%
\begin{abstract}
We propose a novel end-to-end method for cross-view pose estimation.
Given a ground-level query image and an aerial image that covers the query's local neighborhood, the 3 Degrees-of-Freedom camera pose of the query is estimated by matching its image descriptor to descriptors of local regions within the aerial image.
The orientation-aware descriptors are obtained by using a \oldtext{translational} \newtext{translationally} equivariant convolutional ground image encoder and contrastive learning.
The Localization Decoder produces a dense probability distribution in a coarse-to-fine manner with a novel Localization Matching Upsampling module.
A smaller Orientation Decoder produces a vector field to condition the orientation estimate on the localization.
Our method is validated on the VIGOR and KITTI datasets, where it surpasses the state-of-the-art baseline by 72\% and 36\% in median localization error for comparable orientation estimation accuracy.
The predicted probability distribution can represent localization ambiguity, and enables rejecting possible erroneous predictions.
Without re-training, the model can infer on ground images with different field of views and utilize orientation priors if available.
On the Oxford RobotCar dataset, our method can reliably estimate the ego-vehicle's pose over time,
achieving a median localization error under 1 meter and a median orientation error of around 1 degree at 14 FPS.

\end{abstract}

}

\maketitle

\IEEEdisplaynontitleabstractindextext

%
\IEEEpeerreviewmaketitle

\input{1-introduction.tex}
\input{2-relatedwork.tex}
\input{3-methodology.tex}

\input{4-experiments.tex}
\input{5-conclusion.tex}


%

\appendices



\ifCLASSOPTIONcaptionsoff
  \newpage
\fi



%
\bibliographystyle{IEEEtran}
\bibliography{ref}

%

\begin{IEEEbiography}
[{\includegraphics[width=1in,height=1.25in,clip,keepaspectratio]{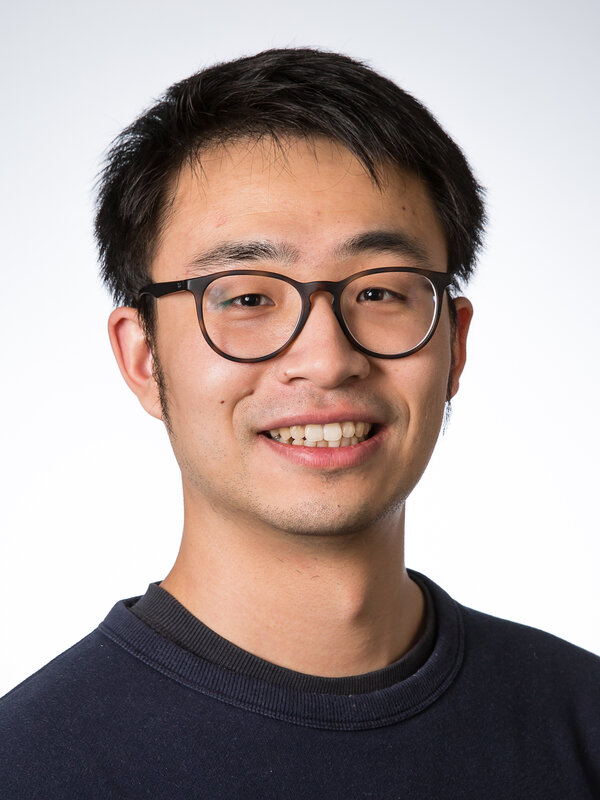}}]{Zimin Xia}
received a B.Sc. in Geomatics Engineering in 2016 at Wuhan University, and an M.Sc. in Geomatics Engineering in 2019 at the University of Stuttgart. Currently, he is a Ph.D. candidate in the Intelligent Vehicles Group at TU Delft.
His research interests include visual localization, ground-aerial image matching, representation learning, and self/semi-supervised learning.
\end{IEEEbiography}

\begin{IEEEbiography}
[{\includegraphics[width=1in,height=1.25in,clip,keepaspectratio]{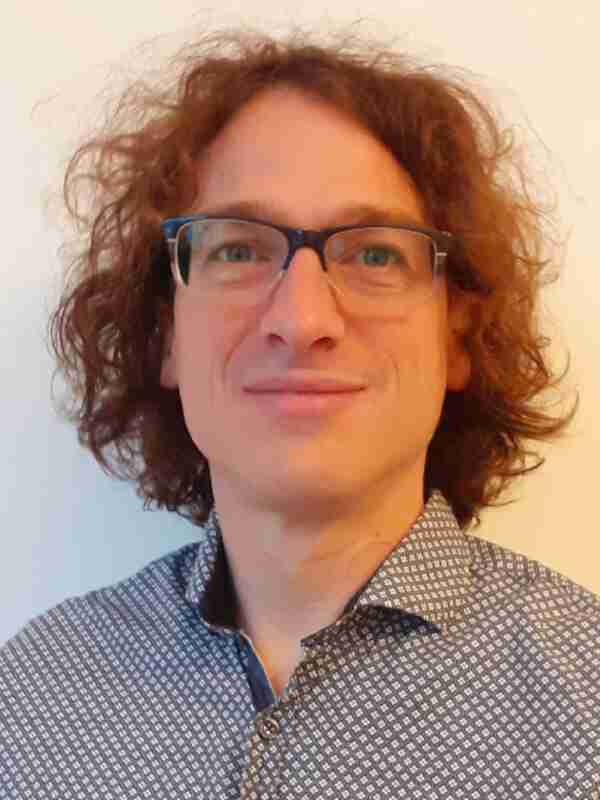}}]{Olaf Booij}
received an M.Sc. in Artificial Intelligence on the topic of Spiking Neural Networks in 2004 and a PhD on vision-based mapping in 2011, both at the University of Amsterdam. Afterwards, he worked as a researcher and team lead at Fugro Intersite, developing robust computer vision methods for offshore and sub-sea and at TomTom as deep learning expert and manager software engineering for automated map making. Currently, he's a guest researcher at the Computer Vision lab at the TU Delft.
\end{IEEEbiography}


\begin{IEEEbiography}[{\includegraphics[width=1in,height=1.25in,clip,keepaspectratio]{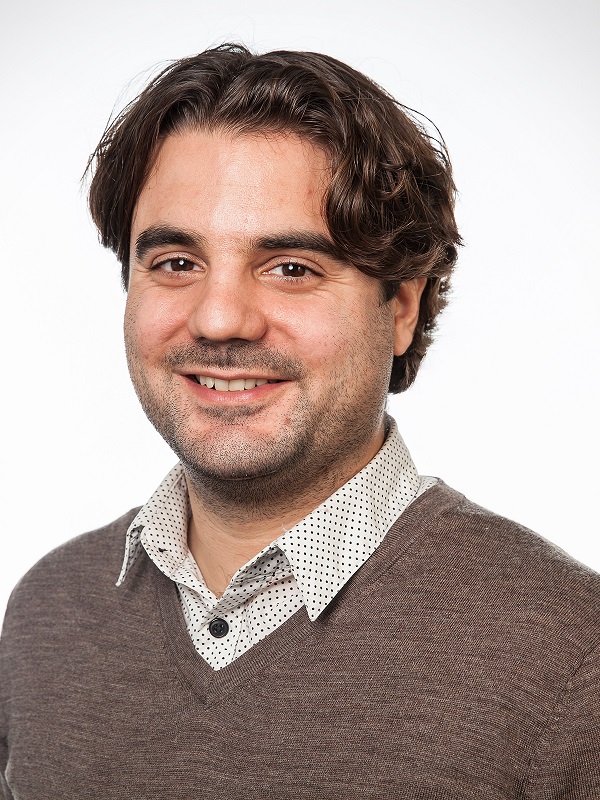}}]{Julian F. P. Kooij}
obtained the Ph.D. degree in 2015 at the University of Amsterdam
on visual detection and path prediction for vulnerable road users.
Afterwards, he joined TU Delft, first in the Computer Vision lab
and since 2016 in the Intelligent Vehicles group
where he is currently an Associate Professor.
His research interests include deep representation learning and probabilistic models for multi-sensor localisation, object detection, and forecasting of urban traffic.
\end{IEEEbiography}


\vfill


\end{document}

%% file: 1-introduction.tex
\IEEEraisesectionheading{\section{Introduction}\label{sec:introduction}}

%
%
%
%

\IEEEPARstart{L}{ocalization} is a core task in autonomous driving and outdoor robotics~\cite{thrun2002probabilistic}.
In urban canyons~\cite{benmoshe2011urbangnss}, Global Navigation Satellite System (GNSS), such as GPS, often has positioning errors of up to tens of meters due to the multipath effect.
Thus, other sensors~\cite{chen2020survey}, such as camera~\cite{lowry2015visual,janai2020computer,sarlin2019coarse} and LiDAR~\cite{barsan2020learning,wei2019learning,lu2019l3}, are used in combination with detailed HD maps~\cite{9561459,9635923} to enhance the localization accuracy and robustness.
In practice, most commercial vehicles are not equipped with expensive LiDAR sensors.
Besides, maintaining an up-to-date HD map is laborious and expensive, especially for areas in fast development.
Hence, exploring alternative map sources for camera-based methods is an important and practical task.
One promising map source is aerial imagery as it provides rich \oldtext{semantic} \newtext{appearance} information with global coverage.


We consider the task of cross-view camera pose estimation, namely, estimating the camera's location and orientation from a given ground-level query image by matching it to geo-referenced aerial imagery.
Previous deep learning-based works \cite{ground-to-aerialgeolocalization,WideAreaImageGeolocalization,workman2015location,SAFA,CVM-Net,regmi2019bridging,cai2019ground,shi2020optimal,zhu2021revisiting,Toker_2021_CVPR,zhu2021vigor,yang2021cross,zhu2022transgeo,rodrigues2022global} successfully performed coarse city- or even country-scale localization, by formulating the localization as image retrieval, i.e. to find the aerial image from a reference database that contains the location of the ground query.
More recently, there has been increasing interest in applying cross-view image retrieval for autonomous driving by zooming into a smaller geographically local region~\cite{Geolocal_feature,xia2021cross}, especially in the urban canyon where GNSS is prone to have large positioning errors.
A few pioneer works \cite{zhu2021vigor,shi2022beyond,xia2022visual,wang2022satellite,hou2022road,lentsch2022slicematch,fervers2022uncertainty} demonstrated the feasibility of pinpointing the 2D location, sometimes together with the orientation, of the ground camera within a known aerial image.
Similar to \cite{shi2022beyond,xia2022visual,lentsch2022slicematch}, we are interested in the 3-Degrees-of-Freedom (3-DoF) camera pose, i.e. planar location and orientation (yaw), instead of the full 6-DoF pose, since the change in camera height, pitch, and roll are \oldtext{often very small in autonomous driving}
\newtext{less important in downstream tasks in autonomous driving, such as motion prediction and planning}.

However, several gaps must be filled before large-scale real-world deployment of cross-view camera pose estimation methods is a realistic possibility for self-driving.
So far, the localization accuracy of existing methods is not yet good enough for autonomous driving requirements, e.g. the lateral and longitudinal error should be below \SI{0.29}{\meter}~\cite{reid2019localization}.
Besides, many methods cannot be run \oldtext{in real-time} \newtext{at sufficiently low latency}, i.e. ${\sim}15$ frames per second (FPS), \oldtext{in self-driving datasets}\newtext{on datasets for self-driving}\cite{OxfordRobotCar1,caesar2020nuscenes,Geiger2013IJRR}.
\oldtext{because of using expensive iterative optimization \cite{shi2022beyond,wang2022satellite} or computationally heavy Transformers~\cite{fervers2022uncertainty}.}
\newtext{For example, \cite{shi2022beyond} relies on iterative optimization to estimate the ground camera's pose.
In \cite{fervers2022uncertainty}, computationally heavy Transformers are used to construct Birds Eye View (BEV) feature representations, and then the BEV representations from ground and aerial views are compared densely at each of the location-orientation combinations (i.e. 3-DoF poses).
Both methods~\cite{shi2022beyond,fervers2022uncertainty} run at a low frame rate, e.g. 2 to 3 FPS.}
\newtext{SliceMatch~\cite{lentsch2022slicematch} fulfills the runtime requirement.
However, it requires pre-computed slice masks for each possible pose it considers, which means that memory overhead limits the number of possible poses and therefore its accuracy.}
We also observe that when the aerial view contains a symmetric scene layout, e.g. at crossroads, single-mode regression-based methods~\cite{zhu2021vigor,hou2022road} might regress to a midpoint between visually similar locations, and optimization-based methods~\cite{shi2022beyond} might get stuck at a wrong local optimum.

\begin{figure*}[t] 
    \centering
    \includegraphics[height=6.5cm]{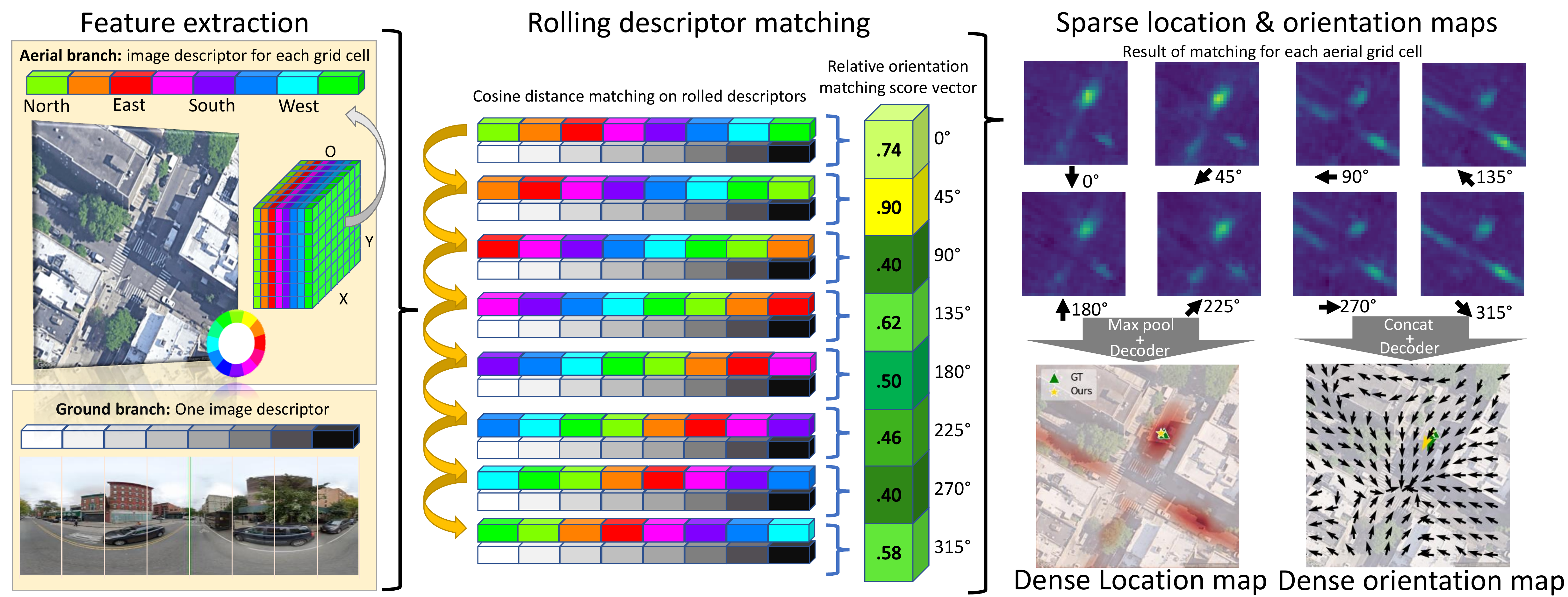}
    \caption{In Convolutional Cross-View Pose Estimation (CCVPE), ground and aerial images are encoded into orientation-aware image descriptors. For the aerial image, we create a grid of descriptors. Efficient \newtext{joint localization and} orientation prediction are enabled by matching rolled \oldtext{ground (or equivalently aerial) image descriptors} \newtext{aerial descriptors with the ground descriptor}. Sparse location and orientation maps are up-sampled into dense maps using decoders with coarse-to-fine matching. We max-pool descriptor matching scores over orientation channels to predict the most probable location considering different orientations. We concatenate the matching scores from different orientations to gather information for accurate orientation prediction. The final orientation prediction is conditioned on localization, i.e. it is selected at the predicted location in the dense orientation map.}
    \label{fig:fig1}
\end{figure*}

\newtext{To improve the pose estimation accuracy over prior works and meanwhile achieve fast runtime, 
}
\oldtext{In this work,} 
we propose a novel method that predicts a multi-modal distribution for localization and jointly considers the orientation of the ground camera.
As shown in Fig.~\ref{fig:fig1}, we exploit the translational equivariance property of convolutional networks to construct orientation-aware image descriptors that represent visual information in both ground and aerial views at different locations with a particular viewing direction.
\newtext{
Joint localization and orientation estimation are achieved by convolving the ground descriptor on the aerial descriptors with circular padding, i.e. matching the ground descriptor to different rolled/shifted versions of the aerial descriptor.
Then, our model regresses the fine-grained orientation based on discrete orientation matching scores and follows a coarse-to-fine formulation to gradually refine a sparse location map into dense output.
The final output orientation is conditioned on the predicted location.
}
\oldtext{We implement the commonly used coarse-to-fine formulation to gradually up-sample a sparse location map to a dense output and condition the output orientation on the predicted location.}

%

Our main contributions are:
(i)~We propose a novel method for end-to-end cross-view camera pose estimation, Convolutional Cross-View Pose Estimation (CCVPE)\footnote{Code is available at \url{https://github.com/tudelft-iv/CCVPE}}.
It surpasses the previous state-of-the-art baselines by a large margin in localization and achieves comparable orientation estimation accuracy on VIGOR and KITTI datasets when testing generalization to new measurements within the same area and across different areas.
\oldtext{
(ii)~In contrast to baselines that output a single location estimate, CCVPE can output a multi-modal distribution that covers all probable locations and avoids regressing or converging to a wrong location between modes.
The predicted probability can also be used to filter out predictions that potentially have large localization or orientation errors.
}
\newtext{
(ii)~CCVPE constructs a multi-modal distribution for localization and uniquely associates each location with its most probable orientation.
We avoid a dense search over all 3-DoF poses (localization+orientation) by discretizing the orientation sparsely and performing additional regression.
This formulation is efficient for fine-grained pose estimation.
We also show the predicted probability can be used to filter out predictions that potentially have large localization or orientation errors.
}
(iii)~Our designed architecture exploits the strength of a translational equivariant feature encoder and contrastive learning. 
Our ground image encoder maintains the spatial scene layout information relative to camera's viewing direction in the ground image descriptor and the contrastive loss enforces aerial descriptors to encode global orientation information. 
These descriptors enable 
\oldtext{orientation estimation} \newtext{jointly localization and orientation estimation} with negligible extra computational cost.
Without re-training, our model can infer the camera pose on images with different horizontal Field of Views (FoVs).
In addition, it can utilize a coarse orientation prior, if available, to improve the localization without re-training.

An earlier version of our method was presented as a conference paper~\cite{xia2022visual}. 
This article extends~\cite{xia2022visual} by the following.
First, we extended our method for end-to-end orientation estimation and jointly consider both location and orientation.
Second, we re-designed our ground and aerial descriptors matching module and exploit coarse-to-fine matching.
Third, we extended our experiments to include the recent KITTI cross-view localization dataset~\cite{shi2022beyond} and compared our method to additional recent state-of-the-art cross-view camera pose estimation baselines~\cite{shi2022beyond,lentsch2022slicematch}.
Fourth, we demonstrated that our method can estimate the ego-vehicle pose at $14$ FPS on the Oxford RobotCar dataset with a median lateral and longitudinal error below $1$ meter, and a median orientation error around $1^\circ$.

%% file: 2-relatedwork.tex
\section{Related Work}
In this section, we review the work related to cross-view camera pose estimation.

\textbf{Cross-view image retrieval} has shown great progress in the past years.
It enjoys the advantage of the widely available geo-referenced aerial images and aims for rough geo-localization by retrieving the aerial image patch that covers the location of the ground-level query image.
The first deep networks for this task date back to 2015 \cite{ground-to-aerialgeolocalization,WideAreaImageGeolocalization,workman2015location}.
Since then, the common practice of using Siamese-like architecture was established.
The ground and aerial images are encoded into image descriptors by two network branches.
Usually, these branches do not share weights~\cite{CVM-Net,SAFA,CVACT}, because two input images are from different domains.
This domain gap is also one of the main challenges in the cross-view setting.
Subsequent works seek to bridge the domain gap between the learned ground and aerial representations via various approaches.


An effective way for minimizing the domain gap is to construct visually similar inputs~\cite{SAFA,regmi2019bridging,Toker_2021_CVPR,shi2022accurate,li2022multi}.
SAFA \cite{SAFA} observes that the polar rays in the aerial image correspond to the vertical lines in the ground image, and proposes to use a polar transformation on the aerial image to build an image that is visually similar to the ground view.
In~\cite{li2022multi}, an inverse polar transformation is used on ground-level panoramas to generate synthetic aerial images.
In~\cite{regmi2019bridging}, the authors bridge the domain gap between the ground and aerial images by generating synthetic aerial images using GANs \cite{regmi2018cross}.
In \cite{Toker_2021_CVPR}, ground-level images are generated from aerial images using GANs, and the features for image generation are shared for cross-view image retrieval.

Besides constructing visually similar inputs, several works try to optimize the learned image feature for retrieval in different ways.
CVM-Net~\cite{CVM-Net} adopts the powerful image descriptor, NetVLAD~\cite{NetVLAD}, to learn how to gather local image features for building global image descriptors.
In~\cite{CVACT}, the authors propose to use the orientation information of both views to guide the model to find more discriminative features across views.
In~\cite{Geolocal_feature}, a rough localization prior from GNSS is considered during training to encourage the model to learn geographically locally discriminative features.
CVFT~\cite{shi2020optimal} considers Optimal Transport theory to facilitate the
feature alignment between ground and aerial images.
Global-assists-local~\cite{rodrigues2022global} addresses the case of retrieving a ground-level query with a limited horizontal FoV and proposes to embed the aerial feature outside the query's FoV into the aerial descriptor to aid the retrieval.
In~\cite{wang2021each}, the feature locality is explicitly enforced when building global image descriptors by partitioning the encoded features.
CVLNet~\cite{shi2022cvlnet} gathers temporal information into the ground descriptor by making use of a ground-level query video.
Recently, transformers are also used.
L2LTR~\cite{yang2021cross} introduces self-cross attention to flow effective information into the descriptors.
TransGeo~\cite{zhu2022transgeo} proposes an attention-guided non-uniform cropping method to attend to and zoom in the informative local image patches.
Apart from retrieval, several works also estimate the orientation of the ground camera~\cite{LocOriStreet,zhai2017predicting,shi2020looking,zhu2021revisiting}.

However, the major limitation of cross-view image retrieval is that the ground query is assumed to be located at the center of the matched aerial image patch, but we may not have aerial image patches whose center is in fact at the unknown test location.
Densifying reference aerial patches reduces the influence of this assumption but increases the computation cost.
In~\cite{shi2022accurate}, the authors propose to zoom into the initial retrieved aerial image and crop smaller aerial patches at a set of candidate locations in the initial retrieved image for second-stage retrieval.
A few works~\cite{CVMJournal,xia2021cross,downes2022city,downes2022wide} fuse the image retrieval results with temporal filters for more accurate localization.
Still, estimating the accurate location and orientation of a single frame ground-level query within a reference aerial image patch that covers it remains an open yet important task.


\textbf{Cross-view camera pose estimation} can be seen as a follow-up task after image retrieval or other coarse localization techniques.
Given a ground-level query and an aerial image that covers the local surrounding of the query, the objective is to estimate the exact location and the orientation of the query within the given aerial image.
In~\cite{zhu2021vigor}, a large-scale dataset for this task is introduced, and the authors propose a model that first retrieves an aerial image given the ground query with a known orientation and then regresses the location offset between them.
Later, \cite{hou2022road} also formulates the localization as a regression problem and includes an additional road extraction training objective.
In~\cite{hu2022beyond}, the orientation of the ground camera is estimated by assuming the location of the ground camera in the aerial image is known.
Instead of regression, \cite{shi2022beyond} solves the query ground camera pose by iterative optimization.
It first warps the feature from the aerial image to a ground view using homography and then uses a multi-level Levenberg-Marquardt algorithm to estimate the 3-DoF ground camera pose using the warped aerial feature and extracted ground-level feature.
SliceMatch~\cite{lentsch2022slicematch} generates aerial descriptors at a set of candidate ground camera poses by pooling aerial features within the geometric extent of the viewing frustum of each pose.
Then the ground descriptor is compared to all aerial descriptors for pose estimation.
Vision Transformers~\cite{dosovitskiy2020image} are used in~\cite{fervers2022uncertainty} to map the features of the ground-level surrounding views to \oldtext{Bird's Eye View (BEV)} \newtext{BEV}, and the mapped BEV \oldtext{features} \newtext{feature maps} are \newtext{densely} compared to \oldtext{features} \newtext{feature maps} extracted from the aerial image for pose estimation.
In~\cite{wang2022satellite}, LiDAR measurements are fused with camera images for cross-view pose estimation.

However, there are several limitations in the above methods.
Some methods only estimate the location~\cite{zhu2021vigor,hou2022road} or orientation~\cite{hu2022beyond} of the ground camera.
Current regression~\cite{zhu2021vigor,hou2022road} or optimization~\cite{shi2022beyond,wang2022satellite} formulation for localization restricts the output to a single mode without uncertainty estimation.
When there are several visually similar locations in the aerial view, regression-based methods~\cite{zhu2021vigor,hou2022road} might regress to the midpoint between those locations, and optimization-based methods~\cite{shi2022beyond} might converge to a wrong local optimum.
More importantly, these methods lack uncertainty estimation to reflect the quality of the outputs.
Besides, the runtime is also a bottleneck in many existing methods, e.g. $2$ to $3$ FPS in~\cite{shi2022beyond,wang2022satellite,fervers2022uncertainty}.
SliceMatch~\cite{lentsch2022slicematch} has fast runtime but it constructs descriptors only at the deepest level in the network and might miss fine-grained details for more accurate pose estimation.
\newtext{Besides, it needs pre-computed slice masks for aggregating features in different orientations.
The memory overhead brought by the masks limits the number of tested poses and thus accuracy.}

\textbf{Floor map-based localization} is a related task to cross-view camera pose estimation.
Instead of aerial images, the goal is indoor localization of sensor measurements, e.g. camera images, in a building's BEV floor map of walls and rooms.
LaLaLoc~\cite{howard2021lalaloc} renders ground-view floor layouts from a BEV floor map, and learns a shared descriptor space for query images and rendered layouts for end-to-end retrieval and pose refinement.
LaLaLoc++~\cite{howard2022lalaloc++} removes the need for the rendering step in LaLaLoc~\cite{howard2021lalaloc} and uses a UNet-like architecture~\cite{ronneberger2015u} to build a descriptor at each candidate location.
Localization is achieved by looking for locations whose local map descriptor is similar to the descriptor of the query.
Laser~\cite{min2022laser} renders ground descriptors from a floor plan in an efficient way and formulates localization as metric learning.
The above methods have many properties in common with cross-view camera pose estimation works.
In particular, our method is similar to Laser~\cite{min2022laser} in constructing orientation-aware image descriptors for orientation estimation.
However, Laser~\cite{min2022laser} cannot directly be used for cross-view camera pose estimation because it relies on explicit map occupancy boundary information, whereas the cross-view visual relation between outdoor ground and aerial images is not explicitly modeled but has to be learned.

%% file: 3-methodology.tex
\section{Methodology} 
Given a ground-level color image $G$ of size $H \times W \times 3$ and an aerial color image $A$ of size $L \times L \times 3$ that covers the local surrounding of $G$, we aim to estimate the 3 Degrees-of-Freedom (DoF) pose, $\hat{\textbf{P}} \in \mathbb{R}^2 \times \mathbb{SO}(2)$, of the camera that took $G$. 
Specifically, $\hat{\textbf{P}} = [\hat{x}, \hat{o}]$.
$\hat{x}=(\hat{u},\hat{v})$ denotes the image coordinates of the location of the camera of $G$ in the aerial image $A$.
$\hat{o} \in [0^\circ,360^\circ)$ denotes the orientation of the camera in the 2D aerial image plane: $0^\circ$ means heading North, i.e. the up direction in the aerial image, and the orientation angle increases in the clockwise direction.
Similar to other cross-view camera pose estimation methods~\cite{shi2022beyond,lentsch2022slicematch}, we assume the pitch and roll angle of the ground camera are small, which is often the case for a vehicle-mounted camera.

\begin{figure*}[t] 
    \centering
    \includegraphics[height=6cm]{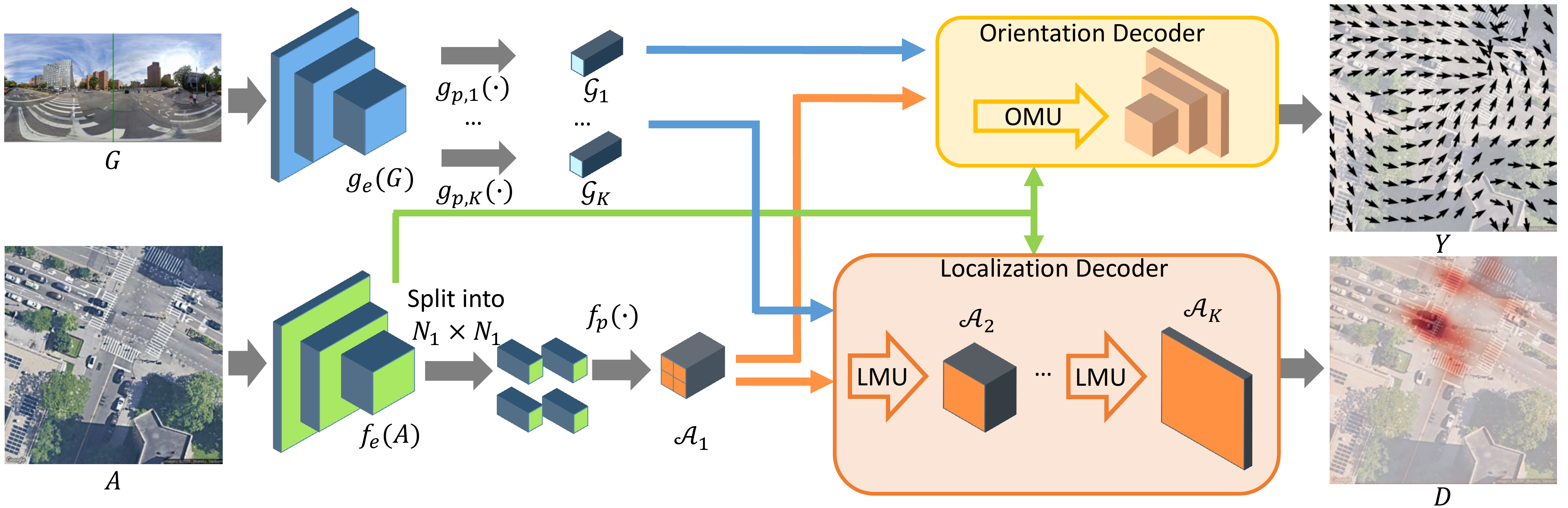}
    \caption{An overview of our proposed Convolutional Cross-View Pose Estimation method, CCVPE. 
    We overlay the output localization distribution (in red) and orientation vector field (black arrows) on top of the input aerial image for intuition.
    }
    \label{fig:network_architecture}
\end{figure*}

\subsection{Methodological design considerations}
\label{sec:design_considerations}
Existing cross-view camera pose estimation methods~\cite{zhu2021vigor,hou2022road,xia2022visual} use a Siamese network with two image encoders without weight-sharing,
fuse the encoder's descriptors at the bottleneck, and finally, have a decoder provide the output.
Our model follows a similar approach with a few novel modifications.

\textbf{1. Multi-modal prediction:}
Instead of treating localization as a uni-modal estimation problem~\cite{zhu2021vigor,hou2022road},
we propose to predict location with a discrete probability distribution $D$ over the pixels in the $L \times L$ aerial image $A$, and formulate the learning as multi-class classification.
This way, the output can capture the potential multi-modal localization ambiguity,
and assign high probability to multiple distinct aerial locations that match the observed ground image $G$.
The probabilistic output could be provided to a downstream robot localization stack for fusion with other sensors, 
or the \textit{Maximum A-Posteriori} (MAP) location can be taken as a single localization estimate.
Furthermore, the probability provides a confidence estimate suitable to reject unreliable predictions, as our experiments will demonstrate.

\textbf{2. Coarse-to-fine descriptor matching:}
To obtain a high-resolution localization distribution,
we propose to \textit{match} a single ground descriptor to local regions in the aerial feature map, e.g.~using the cosine similarity.
The concept of learning a shared feature space where descriptors from different views are compared is also encountered in cross-view image retrieval~\cite{CVM-Net,SAFA,zhu2022transgeo},
but we apply it for dense localization prediction.
Our approach can therefore benefit from the contrastive learning loss to learn discriminative feature spaces for matching.

Furthermore, we observe that the discriminative visual information that distinguishes one aerial region from another depends on the aerial resolution and scale.
We therefore propose to apply this descriptor matching approach in a coarse-to-fine manner,
starting at the low-resolution bottleneck, doubling the feature map resolution each time
until the full target resolution is reached.
At each subsequent level, our approach will match the ground and aerial information and use the resulting matching score to guide the upsampling of the aerial feature to a higher spatial resolution.
We will show that this improves localization accuracy.

\textbf{3. Joint location and orientation matching:}
Location and orientation should be considered jointly. 
Estimating one, while ignoring the other could lead to sub-optimal estimation since the observed layout of the scene in the ground image $G$ only relates to the BEV layout when both location and orientation of the ground camera are correct.
Meanwhile, exploring a prior in one, e.g. orientation, should also benefit the estimation of the other, e.g. localization.
This leads us to two considerations:

First, the image descriptors should \textit{not} be invariant to different orientations. Instead, 
we construct ground descriptors where the elements correspond to information for specific viewing directions \textit{relative} to the camera's unknown orientation,
and aerial descriptors where dimensions capture information in specific \textit{global} viewing directions
(see Fig.~\ref{fig:fig1} left).
An aerial descriptor should only match the ground descriptor if the locations are similar, and if the viewing directions are aligned.
By constructing ground descriptors that are \textit{equivariant} with the camera's viewing direction (i.e.~the horizontal image direction), 
the correct global orientation of the ground camera can be found by reordering its descriptor's feature dimensions (`rolling' the descriptors, see Fig.~\ref{fig:fig1} middle) to match the local aerial descriptor.

Second, in addition to the Localization Decoder, we add an Orientation Decoder 
that predicts orientation as a function of the predicted location, i.e. it predicts a 2D vector field $Y$ over the aerial view that maps each aerial location to the ground camera's most probable orientation if it would be located there.
For instance, if the ground image shows the camera oriented towards a crossing, the localization uncertainty in the aerial view could be spread across the streets approaching the crossing,
and each location would suggest a different global orientation
(see Fig.~\ref{fig:fig1} right).
Uncertainty in the localization output thus also captures uncertainty over the global orientation.

\textbf{4. Generalize to different horizontal FoVs:}
We aim for a model that can be used to match panoramic ground images, as well as images with a limited horizontal FoV without re-training, and can be trained with images of different FoVs for data augmentation.
Therefore, other than constructing descriptors with a fixed length, our ground descriptors have a flexible length that depends on the horizontal FoV of the ground image $G$.

\subsection{Architecture overview}
\label{sec:architecture_overview}
The design considerations from Sec.~\ref{sec:design_considerations} motivate our proposed Convolutional Cross-View Pose Estimation (CCVPE ) architecture, shown in Fig.~\ref{fig:network_architecture}.
One branch of the network, $g(\cdot)$, encodes the ground image $G$, and another branch, $f(\cdot)$, encodes the aerial image $A$.
The descriptors from both encoders are matched in two specialized decoder branches:
the Localization Decoder predicts the 2D spatial distribution $D$,
the Orientation Decoder outputs the dense orientation vector field $Y$.

To match descriptors in a coarse-to-fine manner at $K$ levels,
the ground image $G$ will be encoded into $K$ ground descriptors $\mathcal{G}_k, k \in \{1, \cdots, K\}$, each of a different length $C^G_k$
and capturing the relevant information to distinguish poses at that level's spatial resolution.
Similarly, $K$ aerial descriptor maps $\mathcal{A}_k$ are constructed to represent the relevant matching information of each local aerial region at level $k$.
Each aerial descriptor $\mathcal{A}_k^{i,j}$ at spatial location $(i,j)$ in the descriptor map $\mathcal{A}_k$ has a length of $C^A_k$,
which represents all $360^\circ$ viewing directions at that local region.
When the ground descriptor $\mathcal{G}_k$ is encoded from a $360^\circ$ panoramic ground image, it similarly has $C^G_k = C^A_k$.
If the ground image instead has a limited horizontal FoV, then $C^G_k < C^A_k$
and its descriptors will later be matched to only $C^G_k$ of the $C^A_k$ aerial descriptor dimensions.
The spatial resolution of the aerial descriptor map at level $k$ is $N_k \times N_k = 2N_{k-1} \times 2N_{k-1}$, where $N_1 \times N_1$ is the lowest resolution at the bottleneck,
and $N_{K} \times N_{K} = L/2 \times L/2$ is the last matching level $K$ before the final output.
The first aerial descriptor map, $\mathcal{A}_1$,
is shared between the Localization Decoder and the Orientation Decoder.

\begin{figure}[t] 
    \centering
    \includegraphics[height=4cm]{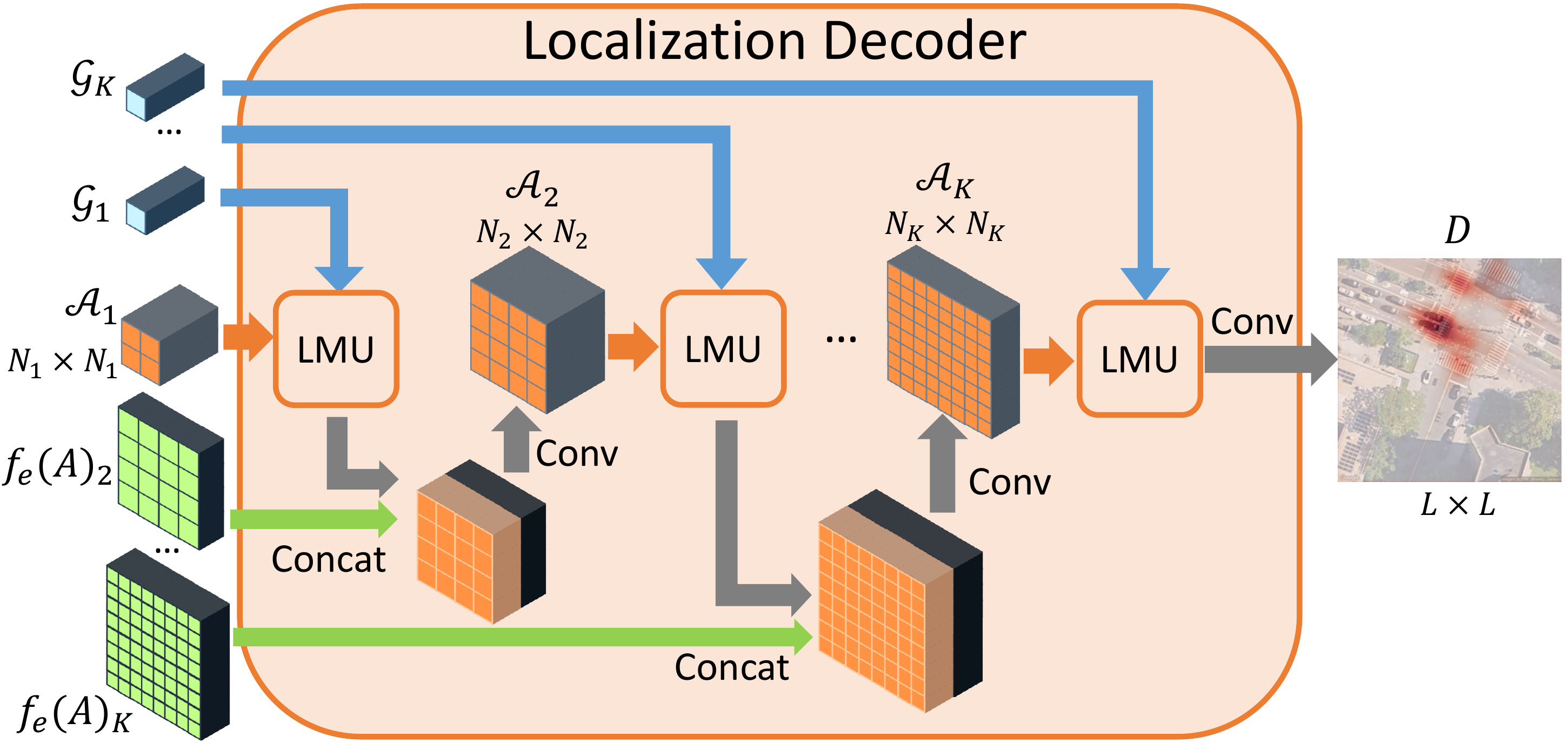}
    \caption{Our proposed Localization Decoder.}
    \label{fig:localization_decoder}
\end{figure}

In the Localization Decoder, see Fig.~\ref{fig:localization_decoder},
the ground and aerial descriptors are compared at multiple resolution levels in a coarse-to-fine manner with our novel Localization Matching Upsampling (LMU) module.
In the Orientation Decoder, we employ a similar
Orientation Matching Upsampling (OMU) module,
though only once after the bottleneck (our experiments will demonstrate this decoder does not benefit from coarse-to-fine matching).
Similar to UNet~\cite{ronneberger2015u} and other models for dense prediction tasks~\cite{dosovitskiy2015flownet,mayer2016large,sun2018pwc}, we furthermore add skip connections from the aerial encoder to the two decoders between feature maps of same spatial resolution.

In the following, we provide details on our proposed descriptor construction, descriptor matching modules, localization and orientation decoders, and used loss functions.

\subsection{Descriptors construction}
\label{sec:descriptors_construction}
Both ground and aerial encoders $g(\cdot)$ and $f(\cdot)$ first apply their own feature extractor, $g_e(\cdot)$ and $f_e(\cdot)$, respectively.
For the ground branch, we then use $K$ ground feature projectors $g_{p,k}(\cdot), k \in \{1, \cdots, K\}$
to extract from the encoder's feature map
the descriptors for the different coarse-to-fine levels, i.e.~$\mathcal{G}_k = g_{p,k}(g_e(G))$.
For the aerial branch, we split the aerial feature volume $f_e(A)$ into $N_1 \times N_1$ sub-volumes and use a shared aerial feature projector $f_p(\cdot)$ on each sub-volume $f_e(A)^{i,j}$ to generate the $N_1 \times N_1$ aerial descriptor map at level $1$, $\mathcal{A}_1^{i,j}=f_p(f_e(A)^{i,j})$.
The aerial descriptor maps $\mathcal{A}_k$ for $k > 1$ will be constructed within the Localization Decoder.

Our model will assume that ground images follow a cylindrical projection,
namely that each column of pixels in the image represents the same number of degrees in the horizontal FoV.
While cylindrical projections are commonly used for panoramic images, regular images with a limited horizontal FoV typically do not use this projection.
We still model all ground images as such, since we find this approximation still works well in practice\footnote{\newtext{See supporting experiments in the Supplementary Material.}}.
We use a translational equivariant ground encoder $g(\cdot)$,
therefore the length $C^G_k = C^A_k \times \frac{F}{360}$ of the ground descriptors $\mathcal{G}_k$ reflects the $F$ degrees horizontal FoV of the ground image $G$.



\textbf{Feature extractors:}
We use a regular convolutional network backbone as the ground and aerial feature extractors $g_e(\cdot)$ and $f_e(\cdot)$ on the input $H \times W \times 3$ ground image $G$ and input $L \times L \times 3$ aerial image $A$, without sharing weights between these branches.
We denote the shape of the encoded ground feature maps as $H' \times W' \times C'$, and of the aerial feature maps as $L' \times L' \times C'$.

\textbf{Ground feature projector:}
A projector $g_{p,k}(\cdot)$ produces a single ground descriptor of length $C^G_k$.
To reduce the computational cost of matching at increasingly higher spatial resolutions, the length of the ground descriptor at the next level is half of that of the ground descriptor at the current level, i.e. $C^G_k = 2 C^G_{k+1}$.

Each projector consists of a $1 \times 1$ convolution to reduce the $C'$ feature channels of the extracted ground feature $g_e(G)$ to $C'_k < C'$.
To summarize the information along the vertical (height) direction in the scene while keeping it equivariant with the horizontal direction (relative viewing direction),
we apply a fully-connected operation along the columns and squeeze the column dimension from $H'$ to $1$, resulting a $1 \times W' \times C'_k$ feature map.
Finally, the ground descriptor $\mathcal{G}_k$ is created by reshaping this feature map into a 1D vector of length $C^G_k = W'  C'_k$.
These ground descriptors $\mathcal{G}_k$ are explicitly orientation-aware, as every block of $C'_k$ elements captures the semantic content in a specific horizontal viewing direction relative to the camera's orientation.

\textbf{Aerial feature projector:}
We create spatial granularity for localization by splitting the $L' \times L' \times C'$ aerial feature volume into $N_1 \times N_1$ feature sub-volumes.
Then, a shared fully connected layer is used as our aerial feature projector $f_p(\cdot)$ to map each of the sub-volumes into an aerial descriptor $\mathcal{A}_1^{i,j}$.
The orientation awareness of our aerial descriptors is encouraged by our loss function, which will align the aerial descriptors with the orientation-aware ground descriptors, see details in Sec.~\ref{sec:loss_functions}.

\begin{figure}[t] 
    \centering
    \includegraphics[height=3.5cm]{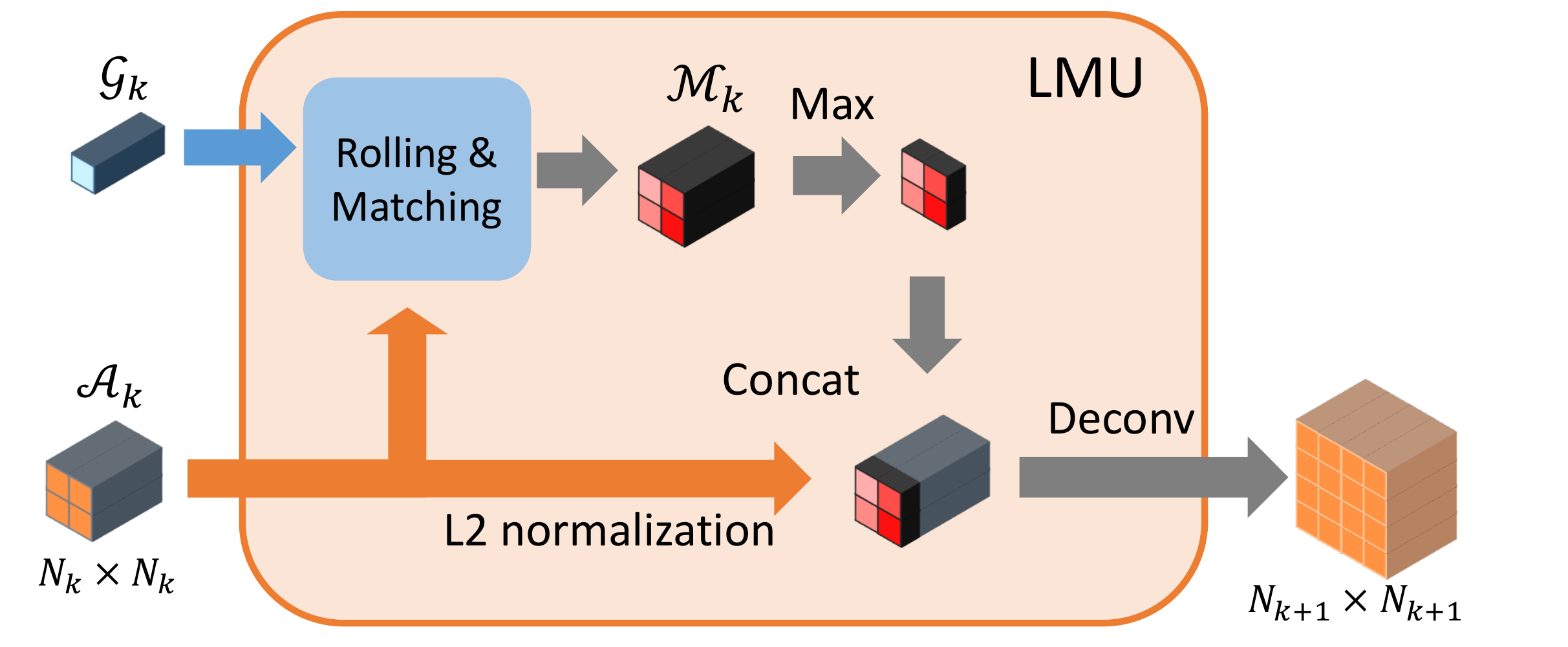}
    \includegraphics[height=3.5cm]{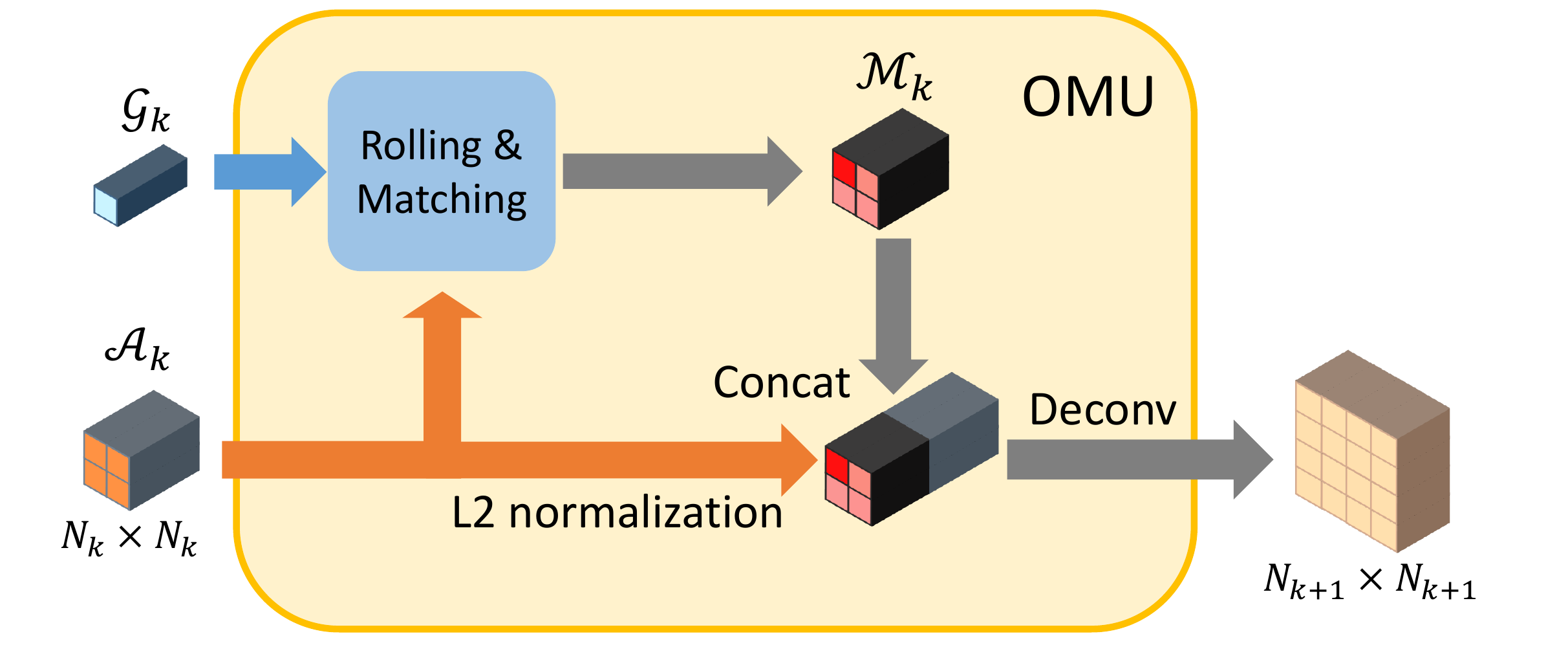}   
    \caption{Our proposed Localization Matching Upsampling (LMU) and Orientation Matching Upsampling (OMU) modules.
    Both generate an aerial feature map at a higher resolution than its input by matching the input aerial features to the ground features.
    }
    \label{fig:LMU_OMU}
\end{figure}

\subsection{Descriptor matching modules}
\label{sec:descriptor_matching_modules}
To jointly consider location and orientation, we match ground descriptors at different locations in the aerial image and consider $R$ different global orientations.
In particular, the matching is done inside our proposed descriptor matching modules, the Localization Matching Upsampling (LMU) module, and the Orientation Matching Upsampling (OMU) module.
As seen in Fig.~\ref{fig:LMU_OMU},
both modules rely on a Rolling \& Matching strategy to compute descriptor matching scores.

\textbf{Rolling \& Matching:}
Both LMU and OMU use the ground descriptor $\mathcal{G}_k$ to `match' the aerial descriptors $\mathcal{A}_k$ at each candidate location $(i,j)$ with a defined global orientation $\frac{r}{R} 360^\circ, r \in \{1, \cdots, R\}$, and output a feature volume with the higher spatial resolution of the next level $k+1$.
To create $R$ global orientations, $[0^\circ, \frac{1}{R} 360^\circ, \cdots,  \frac{R-1}{R} 360^\circ]$, we `roll' our orientation-aware aerial descriptor $\mathcal{A}_k^{i,j}$ at each candidate location $(i, j)$ $R$ times.
Specifically, each `rolling' is achieved by shifting all elements in $\mathcal{A}_k^{i,j}$ by a step length of $\frac{C^A_k}{R}$ to the front, and moving the $C'_k$ front-most elements to the back.
Note that we select $R$ such that the rolling step length $\frac{C^A_k}{R}$ is a multiple of $C'_k$.
The resulting aerial descriptors $\mathcal{A}_k^{i,j,r}$ 
 each represents `what the ground descriptor at level $k$ should contain' at a particular location and global orientation combination $(i,j,\frac{r}{R} 360^\circ)$.

Matching each aerial descriptors $\mathcal{A}_k^{i,j,r}$ to the ground descriptor $\mathcal{G}_k$ is then done by the cosine similarity.
Whereas each $\mathcal{A}_k^{i,j,r}$ captures the environment's appearance in all global directions with a $C^A_k$-dimensional vector,
the $C^G_k$-dimensional ground descriptor $\mathcal{G}_k$ may represent images with a limited horizontal FoV, i.e.~$C^G_k < C^A_k$.
Therefore, we crop the middle $C^G_k$ elements from $\mathcal{A}_k^{i,j,r}$, denoted as $\overline{\mathcal{A}}_k^{i,j,r}$,
to match the same-sized descriptors of the same FoV using,

\begin{align}
    \mathcal{M}_k^{i,j,r} =
    \simmatch(\mathcal{G}_k,\mathcal{A}_k^{i,j,r}) = \frac{\mathcal{G}_k \cdot \overline{\mathcal{A}}_k^{i,j,r}}{\|\mathcal{G}_k\|_2 \times \|\overline{\mathcal{A}}_k^{i,j,r}\|_2}.
    \label{eq:cosine_similarity}
\end{align}
\newtext{The Rolling \& Matching can be seen as convolving a kernel $\mathcal{G}_k$ over $\mathcal{A}_k^{i,j}$ with a stride of $\frac{C^A_k}{R}$, circular padding, and extra normalization.}
The resulting $N_k \times N_k \times R$ matching score volume $\mathcal{M}_k$ expresses how similar the ground descriptor is to the aerial descriptor at each candidate location and orientation.

\textbf{LMU:}
The LMU summarizes the localization cues from $\mathcal{M}_k$ in an invariant manner to the different global orientations. We therefore take for each location the maximum matching score over the $R$ orientations.
These $N_k \times N_k \times 1$ max scores are concatenated to the L2-normalized $N_k \times N_k$ aerial descriptors to guide the upsampling of the aerial feature through a deconvolution.
We show in our ablation study that the L2-normalization before feature concatenation is crucial for good pose estimation performance.

Notably, a prior in the ground camera's orientation is often available for vehicle localization, e.g. indicated by the driving direction. 
Incorporating such an orientation prior 
is straightforward in the LMU
by removing the non-corresponding orientation channels in the matching score volume $\mathcal{M}_k$. This does not require any retraining.



\textbf{OMU:}
Instead of extracting features that are orientation-invariant, OMU explicitly maintains the orientation information in $\mathcal{M}_k$.
It has a similar design as LMU other than that the $\mathcal{M}_k$ is directly concatenated to the L2-normalized aerial descriptors.
Thus the deconvolution layer can make use of information on how the ground descriptor $\mathcal{G}_k$ matches aerial descriptors $\mathcal{A}_k$ at all $R$ orientations.
We apply OMU only to matching level 1 in the Orientation Decoder (we explored other settings, but did not observe a clear benefit).




\subsection{Decoders}
\label{sec:decoders}
We have two separate decoders for localization and orientation estimation.

\textbf{Localization Decoder:} 
Our Localization Decoder contains $K$ LMU modules to gradually increase the spatial resolution of ground and aerial descriptor matching and finally generates a discrete distribution $D$ over the pixels of $L \times L$ aerial image $A$ for localization, see Fig.~\ref{fig:localization_decoder}.

At each level $k$, 
the output feature from LMU is concatenated with the skip-connected aerial feature $f_e(A)_{k+1}$ of the same spatial resolution from the aerial feature extractor $f_e(\cdot)$ to access the scene layout information.
Then, 2D convolution is applied to generate the aerial descriptors $\mathcal{A}_{k+1}$ for level $k+1$.
After the LMU at level $K$, the output feature volume would have a spatial resolution $L \times L$, where $L=2N_K$.
Next, 2D convolution with a softmax activation is applied to convert the feature volume into a $L \times L \times 1$ discrete distribution $D$, in which the values denote how probable the ground camera is located at each pixel location $(i,j)$.
The \textit{Maximum A-Posteriori} (MAP) pixel location $(\hat{i}, \hat{j})$ in $D$ is taken as our final localization estimation, and the image coordinate of its center is our final prediction,  $(\hat{u}, \hat{v}) = (u_{\hat{i}}, v_{\hat{j}})$.

\textbf{Orientation Decoder:}
Our Orientation Decoder up-samples the coarse orientation information into a dense orientation vector field.
It contains an OMU module at the beginning to match the ground descriptor $\mathcal{G}_1$ and aerial descriptors $\mathcal{A}_1^{i,j}$ at level $1$ and upsamples the resulting matching score volume together with the L2-normalized aerial descriptors to spatial resolution $N_2 \times N_2$.
The remainder of our Orientation Decoder uses a series of deconvolutions and convolutions to further upsample the feature volume to the target resolution $L \times L$.
Similar to our Localization Decoder, there is a skip connection that passes aerial feature $f_e(A)_k$ from the aerial encoder to the Orientation Decoder.
The final output of our Orientation Decoder is an $L \times L \times 2$ vector field $Y$ denoting the predicted orientation at each pixel location $(i,j)$ in the aerial image $A$.
The feature channel is L2-normalized and we use the first and second channels to represent the cosine and sine of the predicted orientation angle.
The final orientation prediction $\hat{o}$ is selected in $Y$ at the predicted pixel location $(\hat{i}, \hat{j})$, i.e. $\hat{o}=Y^{(\hat{i}, \hat{j})}$.


\subsection{Loss functions}
\label{sec:loss_functions}
Our loss $\mathcal{L}$ consists of three parts: a contrastive learning loss $\mathcal{L}_{\mathcal{M}}$, a classification loss $\mathcal{L}_{D}$ for localization, and a regression loss $\mathcal{L}_{Y}$ for orientation estimation.
The ground truth location is represented by a discrete distribution $D_{gt}$ of size $L \times L$.
In practice, we place a 2D Gaussian distribution at the ground truth pixel coordinates $(i_{gt},j_{gt})$ to form a smooth ground truth distribution $D_{gt}$.

\textbf{The contrastive learning loss:} $\mathcal{L}_\mathcal{M}$ is an average over contrastive learning losses $\mathcal{L}_{\mathcal{M}k}$, at $K$ levels, i.e.
$\mathcal{L}_\mathcal{M} = \frac{1}{K}\sum_{k=1}^{k=K}{\mathcal{L}_{\mathcal{M}k}}$.
At each level $k$, ${\mathcal{L}_{\mathcal{M}k}}$ is applied on the matching score volume $\mathcal{M}_k$
to encourage the aerial descriptors 
for locations and orientations close to the ground truth poses
to match the ground descriptor $\mathcal{G}_k$.
Since our ground descriptors are orientation equivariant, training with $\mathcal{L}_{\mathcal{M}k}$ enforces the aerial descriptors at the correct locations to be orientation equivariant as well.

The location and orientation space
of $\mathcal{M}_k$ is discretized into $N_k \times N_k \times R$, 
and the ground truth location and orientation would never exactly be centered at a grid point.
We therefore express the closeness of  indices $(i,j,r)$ at level $k$ to the true pose with weights $w_k^{i,j,r} = w_k^{i,j} \times w_k^{r}$.
To obtain spatial weights $w_k^{i,j}$,
we reduce the spatial dimensions of $D_{gt}$ from $L \times L$ to $N_k \times N_k$ by max pooling.
To compute weights $w_k^{r}$ over the $R$ orientation channels,
we first find the orientation indices $r_1$ and $r_2$ of the discrete angles closest to the true orientation, 
$360^\circ\times ({r_1}/{R}) < o_{gt} < 360^\circ\times ({r_2}/{R})$.
We only assigning non-zero weight to ${r_1}$ and ${r_2}$,
where their weight is inversely proportional to their relative angular distance to $o_{gt}$,
and $w^{r_1}_k + w^{r_2}_k = 1$.
%
We can now finally define $\mathcal{L}_{\mathcal{M}k}$
as a weighted sum $\mathcal{L}_{\mathcal{M}k} = \sum_{i,j,r} w_k^{i,j,r} \mathcal{L'}_{\mathcal{M}k}(i,j,r)$
of infoNCE losses~\cite{oord2018representation}
on the cosine similarity of
Eq.~\eqref{eq:cosine_similarity},

\begin{align}
    \mathcal{L'}_{\mathcal{M}k}(i,j,r) = -\log\frac{\exp(\simmatch(\mathcal{G}_k, \mathcal{A}_k^{i,j,r})/\tau)}{\sum_{i',j',r'} \exp(\simmatch(\mathcal{G}_k, \mathcal{A}_k^{i',j',r'})/\tau)}.
    \label{eq:infoNCE_loss}
\end{align}

%

\textbf{The localization loss:}
We formulate the localization problem as a multi-class classification. 
In our main setting, the localization loss $\mathcal{L}_D$ is a cross-entropy loss,

\begin{align}
    \mathcal{L}_D = - \sum_{i=1}^{L} \sum_{j=1}^{L} {D_{gt}^{i,j} \log D^{i,j}},
    \label{eq:CE_loss}
\end{align}
where $(i,j)$ are pixel coordinates. 

As an alternative, we also consider training the localization distribution $D$ by minimizing the transported mass from $D$ to $D_{gt}$ based on Optimal Transport theory~\cite{monge1781memoire}.
For this, we use a Wasserstein distance-based loss~\cite{frogner2015learning} as our $\mathcal{L}_D$.
Unlike cross-entropy, Wasserstein distance considers the distance between the mass in the source and target distributions.
To compute the Wasserstein distance loss $\mathcal{L}_D$ between $D$ and $D_{gt}$ efficiently,
we here define $D_{gt}$ as a one-hot distribution.
This loss then becomes,
\begin{align}
    \mathcal{L}_D = - \sum_{i=1}^{L} \sum_{j=1}^{L} {d(i,j) \cdot D^{i,j}}.
    \label{eq:W_loss}
\end{align}

In Eq.~\eqref{eq:W_loss}, $d(i,j)$ denotes the L2-distance in pixels between the pixel location $(i,j)$ in $D$ to the ground truth pixel location $(i_{gt},j_{gt})$, i.e. $d(i,j) = \sqrt{(i-i_{gt})^2 + (j-j_{gt})^2}$.
%
We compare the cross-entropy loss and Wasserstein distance-based loss in our ablation study, but will use the cross-entropy loss as $\mathcal{L}_D$ in our main experiments.

\textbf{The orientation loss:} 
Instead of treating the orientation prediction as a discrete classification problem, which would result in a large number of classes for joint localization and orientation supervision, we formulate this problem as regression.
Since we use a Gaussian smoothed ground truth $D_{gt}$, we sum the contributions from smoothed ground truth locations and define our orientation loss $\mathcal{L}_{Y}$ as,

\begin{align}
    \mathcal{L}_{Y} = \sum_{i=1}^{L} \sum_{j=1}^{L} D_{gt}^{i,j} \left(\left(\cos{\left(o_{gt}\right)} - Y_1^{i,j}\right)^2 + \left(\sin{\left(o_{gt}\right)} - Y_2^{i,j}\right)^2\right).
    \label{eq:ori_loss}
\end{align}

In Eq.~\eqref{eq:ori_loss}, $o_{gt}$ is the ground truth orientation, $Y$ is the $L \times L \times 2$ predicted orientation vector field.
$Y_1$ and $Y_2$ denote the first and second channel of $Y$.
Multiplying with $D_{gt}$ removes the contribution to the orientation loss $\mathcal{L}_{Y}$ at wrong locations.

\textbf{The total loss:} $\mathcal{L}$ is a weighted combination of the localization loss $\mathcal{L}_D$, orientation loss $\mathcal{L}_Y$, and contrastive learning loss $\mathcal{L}_M$:

\begin{align}
    \mathcal{L} = \mathcal{L}_D + \alpha \mathcal{L}_Y + \beta \mathcal{L}_M,
    \label{eq:loss}
\end{align}

where the $\alpha$ and $\beta$ are hyperparameters that weigh the importance of $\mathcal{L}_Y$ and $\mathcal{L}_M$ during training.




%% file: 4-experiments.tex
\section{Experiments}
In this section, we first introduce the three used datasets, followed by our evaluation metrics.
After this, we provide our implementation details.
Then, our CCVPE method is compared to previous state-of-the-art baselines w.r.t. generalization to new measurements within the same areas and across different areas.
Next, we study how our method works with an orientation prior and on images with different horizontal FoVs.
Then, we use our method to estimate the pose of the ego-vehicle along test traversals using sequences of ground images.
Finally, an extensive ablation study \newtext{and an analysis of runtime are provided}.

\subsection{Datasets}
We test the generalization of all models to new measurements within the same areas and across different areas on the VIGOR~\cite{zhu2021vigor} and KITTI~\cite{Geiger2013IJRR} datasets.
On the Oxford RobotCar dataset~\cite{OxfordRobotCar1,OxfordRobotCar2}, our method is used to estimate the ego-vehicle pose frame-by-frame using the ground image sequence collected in test traversals.
The original KITTI and Oxford RobotCar datasets do not contain any aerial images, therefore we make use of the collected aerial images from \cite{shi2022accurate} for KITTI, and \cite{xia2021cross,xia2022visual} for Oxford RobotCar.

\textbf{The VIGOR dataset}~\cite{zhu2021vigor} contains ground-level panoramic images and aerial patches collected in four cities in the US.
The aerial patches are distributed regularly as a grid, providing seamless coverage of the 4 target cities.
Each aerial patch covers a ${\sim}70$~m $\times$ $70$~m ground region.
The orientation of the panorama and aerial patch is aligned such that the center vertical line in the panorama corresponds to the up direction (North) in the aerial patch.
In our experiments, changing the orientation of the ground panorama is achieved by shifting the image along the horizontal axis.
Reducing the horizontal FoV is achieved by dropping the image columns at the left and right borders.
Since the ground truth labels are improved by~\cite{lentsch2022slicematch}, we use those more accurate labels.
The VIGOR dataset defines the aerial patches as either positive or semi-positive for each ground image.
An aerial patch is positive if its center $1/4$ region contains the ground camera's location, otherwise, it is semi-positive.
In our experiments, we use positive aerial images for training and testing all models.
We adopt the Same-Area and Cross-Area split from~\cite{zhu2021vigor}.
On the Same-Area split, we train models on images from all four cities and test models on images from the same cities.
Training and test sets do not share any ground images but may share aerial patches.
On the Cross-Area split, models are trained on image pairs from New York and Seattle and tested on pairs from Chicago and San Francisco.
For validation and hyperparameter tuning, we randomly select $20\%$ of the data from the training set.

\textbf{The KITTI dataset}~\cite{Geiger2013IJRR} is collected by a vehicle platform in Karlsruhe, Germany, covering city, rural area, and highway scenarios.
The stereo camera faces the driving direction and has a horizontal FoV of $90^\circ$.
In~\cite{shi2022beyond}, the authors make use of the images from the left camera of the stereo camera and collected aerial images with ground resolution ${\sim}0.20$~m/pixel to enable cross-view pose estimation.
Each aerial patch covers a ${\sim}100$~m $\times$ $100$~m ground area.
The data is split into Training, Test 1, and Test 2 sets.
Images in Training and Test 1 sets are from the same regions.
Images in Test 2 set are from different areas than those in the Training set.
In our experiments, we refer to Test 1 and Test 2 sets as Same-Area and Cross-Area.
As assumed in~\cite{shi2022beyond}, ground images are located within a $40$~m $\times$ $40$~m area in the corresponding aerial patches' center, and there is an orientation prior with noise between $-10^\circ$ and $10^\circ$.
\newtext{In this case, a random rotation between $-10^\circ$ and $10^\circ$ is applied on each aerial image whose `East' orientation was aligned with the ground image.}
In our experiments, we adopt the same setting, and \newtext{also} provide extra results \oldtext{where we assume an unknown orientation}
\newtext{for unknown orientations where a random rotation from the $360^\circ$ circular domain is applied on each orientation-aligned aerial image}.

\textbf{The Oxford RobotCar dataset}~\cite{OxfordRobotCar1,OxfordRobotCar2} contains videos \newtext{with a limited horizontal FoV} collected over multiple traversals at different times, seasons, and weather conditions, along the same route in Oxford, UK.
In~\cite{Geolocal_feature,xia2021cross}, the authors collected aerial patches for retrieval, and later \cite{xia2022visual} stitched those aerial patches with their collected extra ones into a continuous aerial image that covers the Oxford area.
We follow the same setting as in~\cite{xia2021cross,xia2022visual} that the training, validation, and test data are from different traversals to test our model's generalization to different dynamic objects, weather, and lighting conditions across time.
Instead of directly using the sparse test images used in~\cite{xia2021cross,xia2022visual}, we sample test ground images from the original Oxford RobotCar dataset~\cite{OxfordRobotCar1,OxfordRobotCar2} at a higher frame rate, $\sim 1.6$ FPS, for our experiment of ego-vehicle following.
In total, there are three test traversals, enabling testing in Summer and Winter.
During training, aerial patches that cover ${\sim}74$~m $\times$ $74$~m ground area are randomly cropped from the continuous map around a location that is less than ${\sim}26$~m away from the vehicle's location.
For validation and testing, we use the same set of aerial patches as in~\cite{xia2022visual}.
\newtext{In our experiment, the orientation of the ground camera is always assumed unknown, so we simply use the ground images and north-aligned aerial images as input pairs.}

\subsection{Baselines methods}
We compare CCVPE to two types of baselines.

First, we include the state-of-the-art (SOTA) cross-view pose estimation baselines: the cross-view regression method (CVR)~\cite{zhu2021vigor}, iterative optimization method LM~\cite{shi2022beyond}, and SliceMatch~\cite{lentsch2022slicematch}.
CVR~\cite{zhu2021vigor} is originally designed for joint image retrieval and location regression.
For a fair comparison, we train it for localization within a given aerial image (we find it achieves better localization error than training it for retrieval + localization).
We also trained a CVR model using the same EfficientNet-B0~\cite{tan2019efficientnet} as its feature extractor, denoted as Eff-CVR.
LM~\cite{shi2022beyond} uses an iterative method to estimate the location and orientation of the ground camera on the aerial image. 
We use the provided model by its author and the same setting on its prior~\cite{shi2022beyond} on KITTI dataset, namely, the ground images are located in a $40m \times 40m$ area in the input aerial image center, and a rough orientation prior with noise between $- 10^\circ$ and $10^\circ$ is available during training and test time. 
For completeness, we also include the model trained and tested without any orientation prior. 

On the KITTI dataset, \cite{shi2022accurate} evaluated several image retrieval or retrieval with orientation estimation baselines by limiting their searching area to a region of $40m \times 40m$.
We include the same fine-grained cross-view image retrieval baselines from \cite{shi2022accurate}.

\subsection{Evaluation metrics}
We use mean and median error over all test samples as our main evaluation metrics for both localization and orientation prediction.
For localization, the error is the distance in meters between the predicted location and the ground truth location.
For orientation, the error is the angular difference ($^\circ$) between the predicted camera orientation at the predicted location and ground truth camera orientation.
In addition, we report the percentage of test samples that has an error below certain thresholds,
namely $1$ m, $3$ m, and $5$ m for localization, and $1^\circ$, $3^\circ$, and $5^\circ$ for orientation.
For localization, longitudinal and lateral error w.r.t.~the vehicle's driving direction are given separately~\cite{shi2022beyond}.
For image retrieval methods, the localization error is calculated by measuring the distance between the center of the retrieved aerial image patch and the ground truth location.

To measure if the true location receives probability mass, and thus would not be discarded if used in a probabilistic temporal filter, we also measure the predicted probability at the ground truth pixel.
For the baseline method that regresses~\cite{zhu2021vigor} to a single location without uncertainty estimates, we assume their prediction is the peak of an isotropic Gaussian distribution, and estimate the standard deviation of this Gaussian distribution on the validation set.
SliceMatch~\cite{lentsch2022slicematch} measures descriptors' similarity scores on their candidate poses.
We use the scores at the candidate location that is closest to the ground truth location to derive the predicted probability at the ground truth pixel.
Finally, we measure our model's runtime on a Tesla V100 GPU.

\subsection{Implementation details}
\label{sec:implementation_details}
EfficientNet-B0~\cite{tan2019efficientnet} is used as our ground and aerial feature extractors, $g_e(\cdot)$ and $f_e(\cdot)$.
There is no weight-sharing between them.
When the ground image $G$ is panoramic, circular padding in the horizontal direction is used inside the ground encoder $g(\cdot)$, and zero padding otherwise.
For other model components, and the vertical direction padding in $g(\cdot)$, we use zero padding.
During training, the feature extractor is initialized from ImageNet~\cite{deng2009imagenet} pre-trained weights, and other components are initialized randomly.
Our model is trained using the Adam optimizer~\cite{Adam} with a learning rate of $0.0001$.
We use the default drop connect~\cite{wan2013regularization} rate, 0.2, from EfficientNet~\cite{tan2019efficientnet}, and the default $\tau = 0.1$ in Eq.~\eqref{eq:infoNCE_loss}, from infoNCE loss~\cite{oord2018representation}. 
Different weights $\alpha =\num{1e-2}, \cdots, \num{1e2}$ and $\beta = 10, \cdots ,\num{1e5}$  were tested for weighing the orientation loss $\mathcal{L}_Y$ and contrastive learning loss $\mathcal{L}_M$, and $\alpha = 10$ and $\beta=\num{1e4}$ are selected since they provide best validation performance.
The model bottleneck size $N_1 \times N_1$ is set to $8 \times 8$, and consequently there are $K=6$ levels in our coarse-to-fine descriptor matching.

\newtext{Note that, even though our method assumes ground images follow cylindrical projection, for KITTI and Oxford RobotCar datasets we directly input their perspective images.
Our detailed study in the Supplementary Material shows that both projections work equally well in practice.}

\subsection{Generalization to new measurements in same area}
First, we compare our method, CCVPE, to baselines for generalizing to new measurements (i.e.~ground images) in the same area.
This corresponds to use cases that target operation in a predetermined area, such as driving in one city, so models can be trained on data from that specific area.
For this task, we report the evaluation results on VIGOR Same-Area test set and KITTI Same-Area (Test 1) set.

\begin{table*}[ht]
\caption{Evaluation on VIGOR Same-Area and Cross-Area test set. \textbf{Best in bold.} We report mean and median localization and orientation error, as well as the probability at the ground truth pixel in the aerial image, denoted as `P@GT'. \oldtext{In the left column,}\newtext{The left-most column indicates orientation uncertainty:} `$0^\circ$' means testing with \newtext{known} orientation, \newtext{such that the center vertical line in the panorama corresponds to the North direction in the aerial image, and the known orientation is used to remove the non-corresponding orientation channels in the matching score volume}; \oldtext{`w/o.' means testing without orientation prior.} \newtext{`$360^\circ$' means the test orientation is unknown and then the panoramic ground image was horizontally shifted corresponding to a random angle in the $360^\circ$ circular domain.}}
    \label{tab:VIGOR_test}
    \centering
    \begin{tabular}{|P{0.15cm}|P{1.85cm}|P{0.8cm}P{0.8cm}|P{0.8cm}P{0.8cm}|P{0.8cm}P{0.8cm}|P{0.8cm}P{0.8cm}|P{0.8cm}P{0.8cm}|P{0.8cm}P{0.8cm}|}
    \hline
    \multicolumn{2}{|c|}{\multirow{3}{*}{VIGOR test}}  & \multicolumn{6}{c|}{Same-Area} & \multicolumn{6}{c|}{Cross-Area} \\ 
    \cline{3-14}
    \multicolumn{2}{|c|}{} & \multicolumn{2}{c|}{$\downarrow$ Localization (m)}   & \multicolumn{2}{c|}{$\downarrow$ Orientation ($^\circ$)}  & \multicolumn{2}{c|}{$\uparrow$ P@GT (\num{e-3})} & \multicolumn{2}{c|}{$\downarrow$ Localization (m)}   & \multicolumn{2}{c|}{$\downarrow$ Orientation ($^\circ$)}  & \multicolumn{2}{c|}{$\uparrow$ P@GT (\num{e-3})} \\ 
    \multicolumn{2}{|c|}{} & mean & median  & mean & median & mean & median & mean & median  & mean & median & mean & median \\
    \hline
    \multirow{4}{*}{\rotatebox{90}{$0^\circ$}} & CVR \cite{zhu2021vigor} & 8.82 & 7.68 & - &  - & 0.02 & 0.02 & 9.45 & 8.33 & - &  - & 0.02 & 0.02 \\
    \cline{2-14}
    & Eff-CVR & 7.89 & 6.25 & - &  - & 0.02 & 0.03 & 8.27 & 6.60 & - &  - & 0.02 & 0.03\\
    \cline{2-14}
    & SliceMatch~\cite{lentsch2022slicematch} & 5.18 & 2.58 & - & - & 0.06 & 0.05 & 5.53 & 2.55 & - & - & 0.06 & 0.06 \\
    \cline{2-14}
    & \textit{CCVPE} (ours) & \textbf{3.60} & \textbf{1.36} & - & - & \textbf{1.60} & \textbf{1.12} & \textbf{4.97} & \textbf{1.68} & - & - & \textbf{1.08} & \textbf{0.71} \\
    \hline
    \multirow{2}{*}{\rotatebox{90}{\newtext{$360^\circ$}}} 
    & SliceMatch~\cite{lentsch2022slicematch} & 8.41 & 5.07 & 28.43 & \textbf{5.15} & 0.02 & 0.02 & 8.48 & 5.64 & \textbf{26.20} & \textbf{5.18} & 0.02 & 0.02 \\
    \cline{2-14}
    & \textit{CCVPE} (ours) & \textbf{3.74} & \textbf{1.42} & \textbf{12.83} & 6.62 & \textbf{1.47} & \textbf{1.00} & \textbf{5.41} & \textbf{1.89} & 27.78 & 13.58 & \textbf{0.93} & \textbf{0.58} \\
    \hline
    \end{tabular}
\end{table*}

\textbf{Pose estimation on VIGOR Same-Area:}
As shown in Tab.~\ref{tab:VIGOR_test} Same-Area, when testing with images with known orientation, our method surpasses all baselines, CVR~\cite{zhu2021vigor}, Eff-CVR, and SliceMatch~\cite{lentsch2022slicematch}, w.r.t. mean and median localization errors.
Replacing the VGG~\cite{VGG} backbone with EfficientNet-B0~\cite{tan2019efficientnet} for CVR improves localization performance, but Eff-CVR still has significantly higher localization errors than ours.
When the orientation of test images is unknown, our method beats the previous SOTA SliceMatch by a large margin in localization, i.e. 56\% in the mean error and 72\% in the median error.
Regarding orientation estimation, CVR could not infer the orientation of the ground camera, and thus it is not included in the comparison.
SliceMatch and our method address orientation prediction differently.
SliceMatch divides the $360^\circ$ orientation space into 64 bins and selects the most probable one based on descriptors matching, while our method creates $R=20$ orientation scores and regress the true orientation after the grid-based matching.
Quantitatively, our method has a lower mean orientation error but a slightly higher median orientation error than SliceMatch.
We demonstrate in our Supplementary Video that our method can smoothly track the change in the orientation of the ground camera.
Grid-based solutions, such as SliceMatch, would need to densify their grid, resulting in more memory and computation needs.

\begin{figure*}[t] 
\centering
 \includegraphics[width=0.24\textwidth]{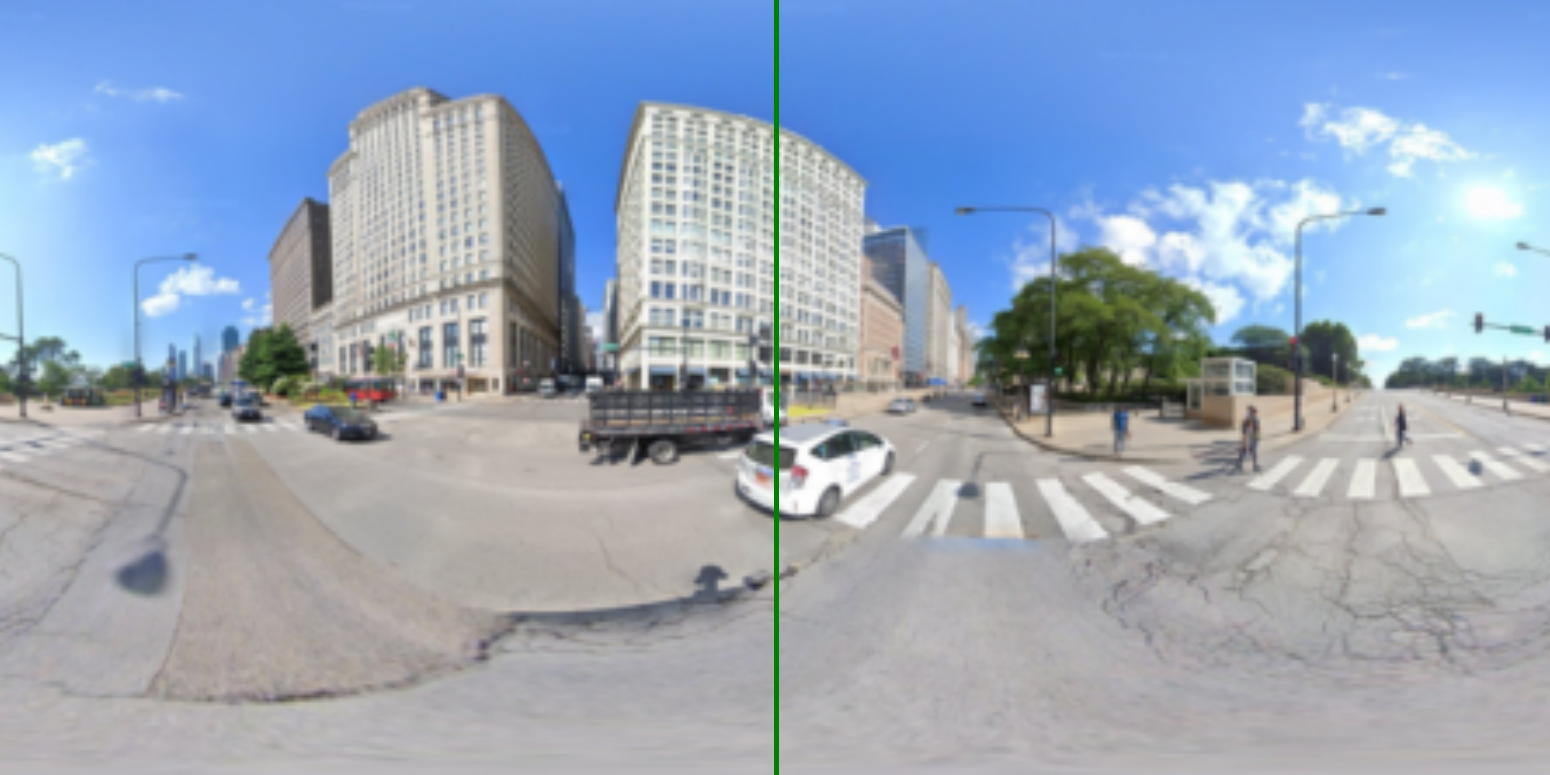}
 \includegraphics[width=0.24\textwidth]{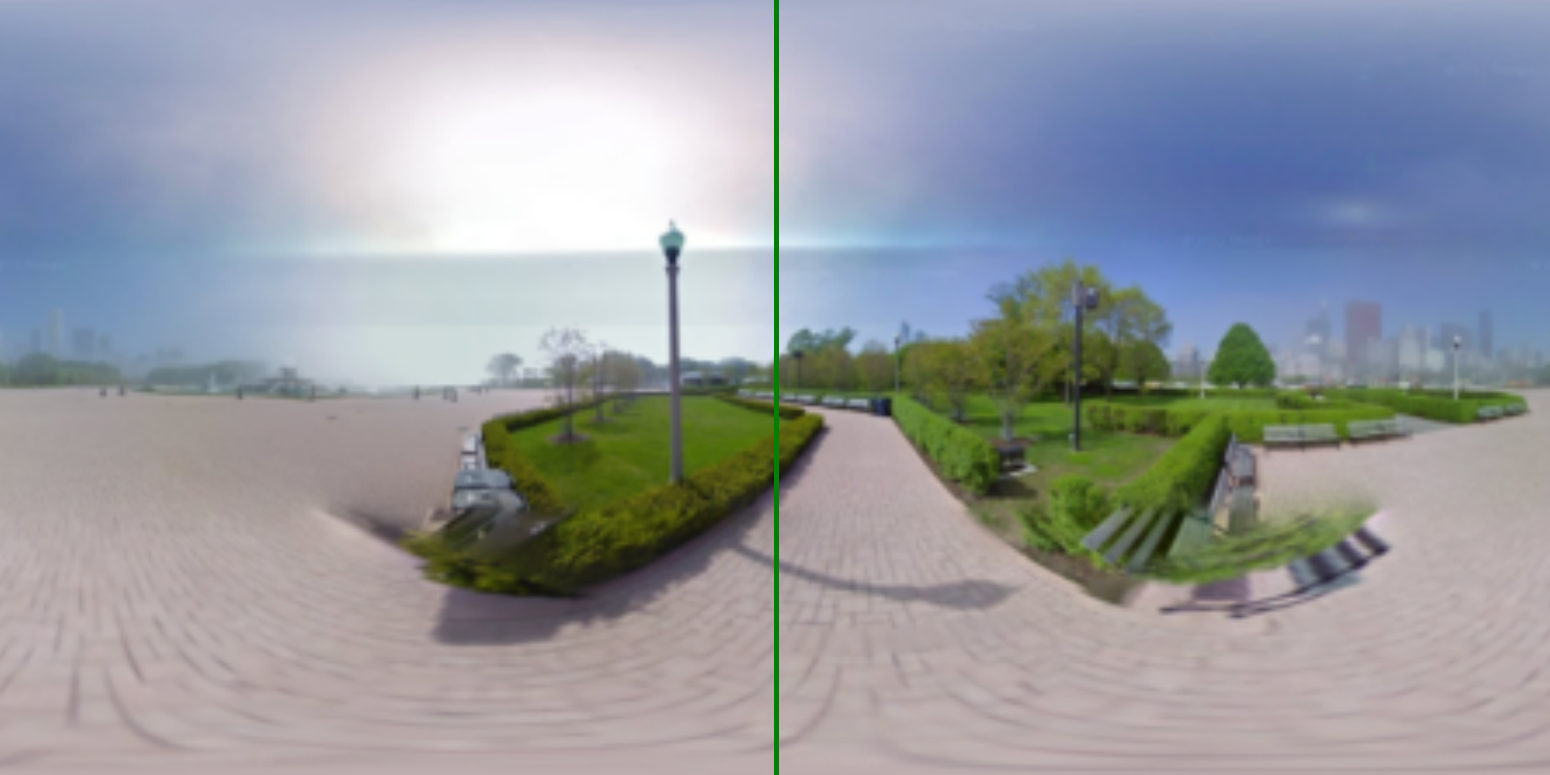}
 \includegraphics[width=0.24\textwidth]{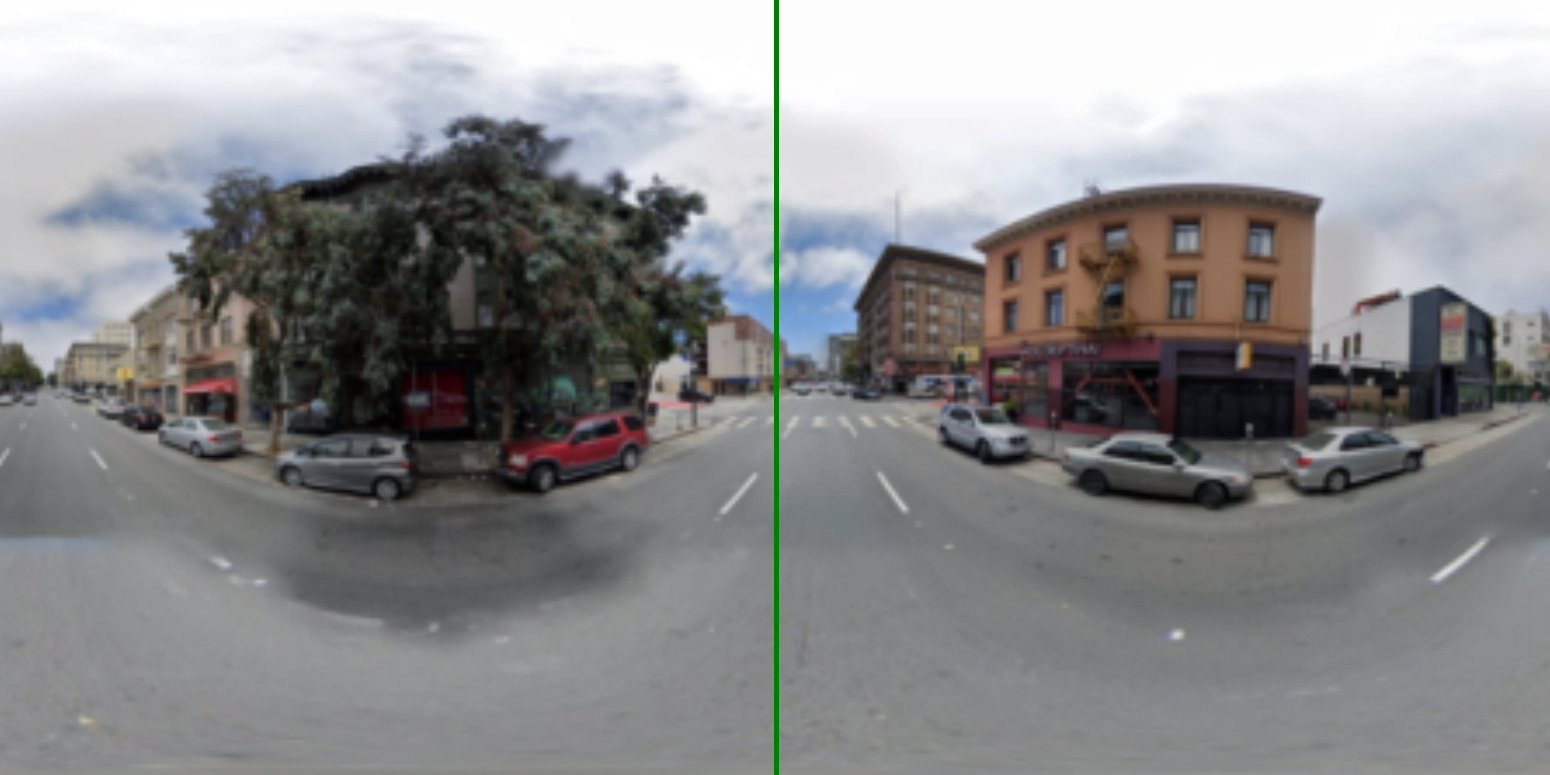} 
 \includegraphics[width=0.24\textwidth]{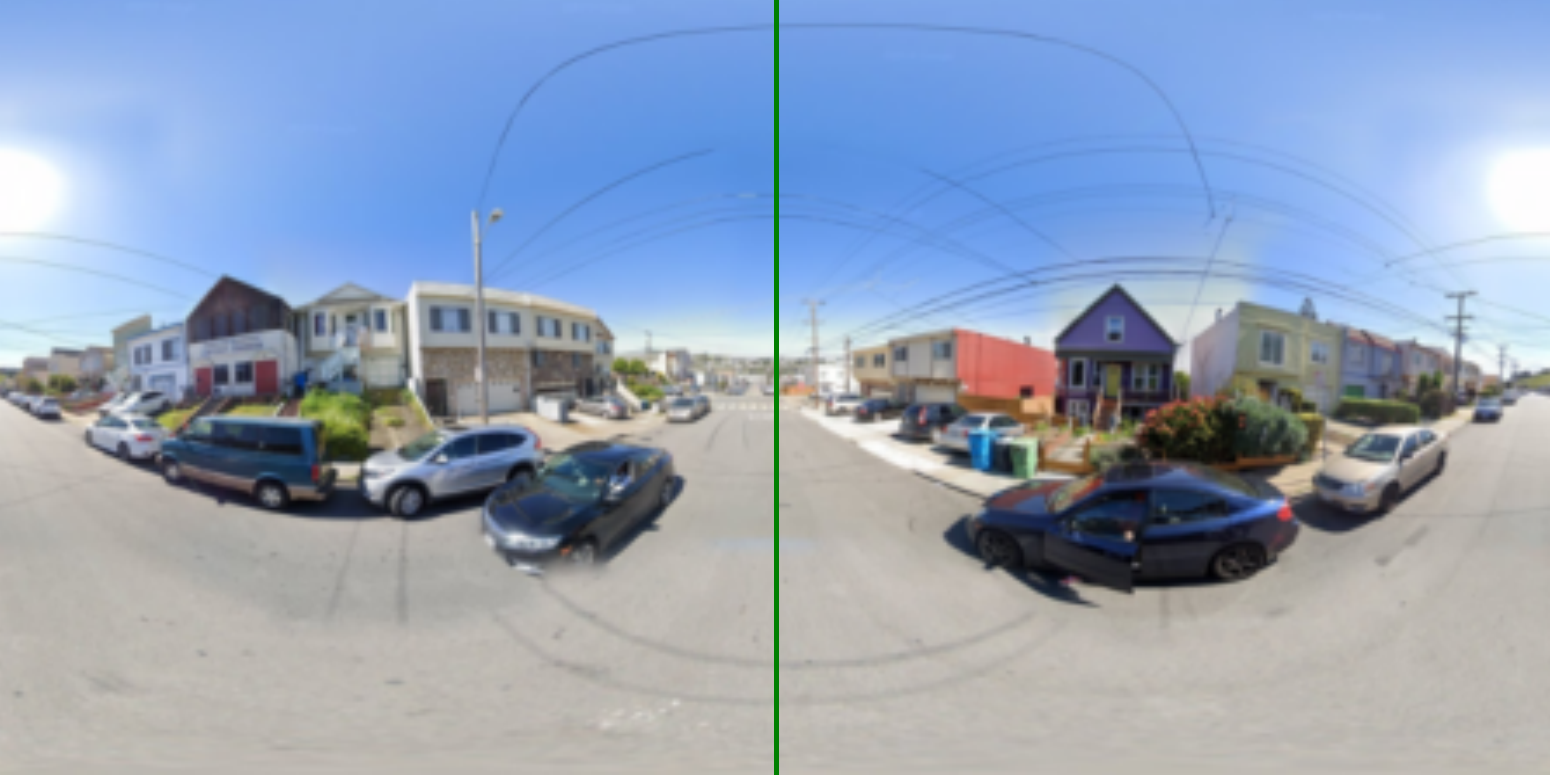}\\
 \vspace{1mm}
 \includegraphics[width=0.24\textwidth]{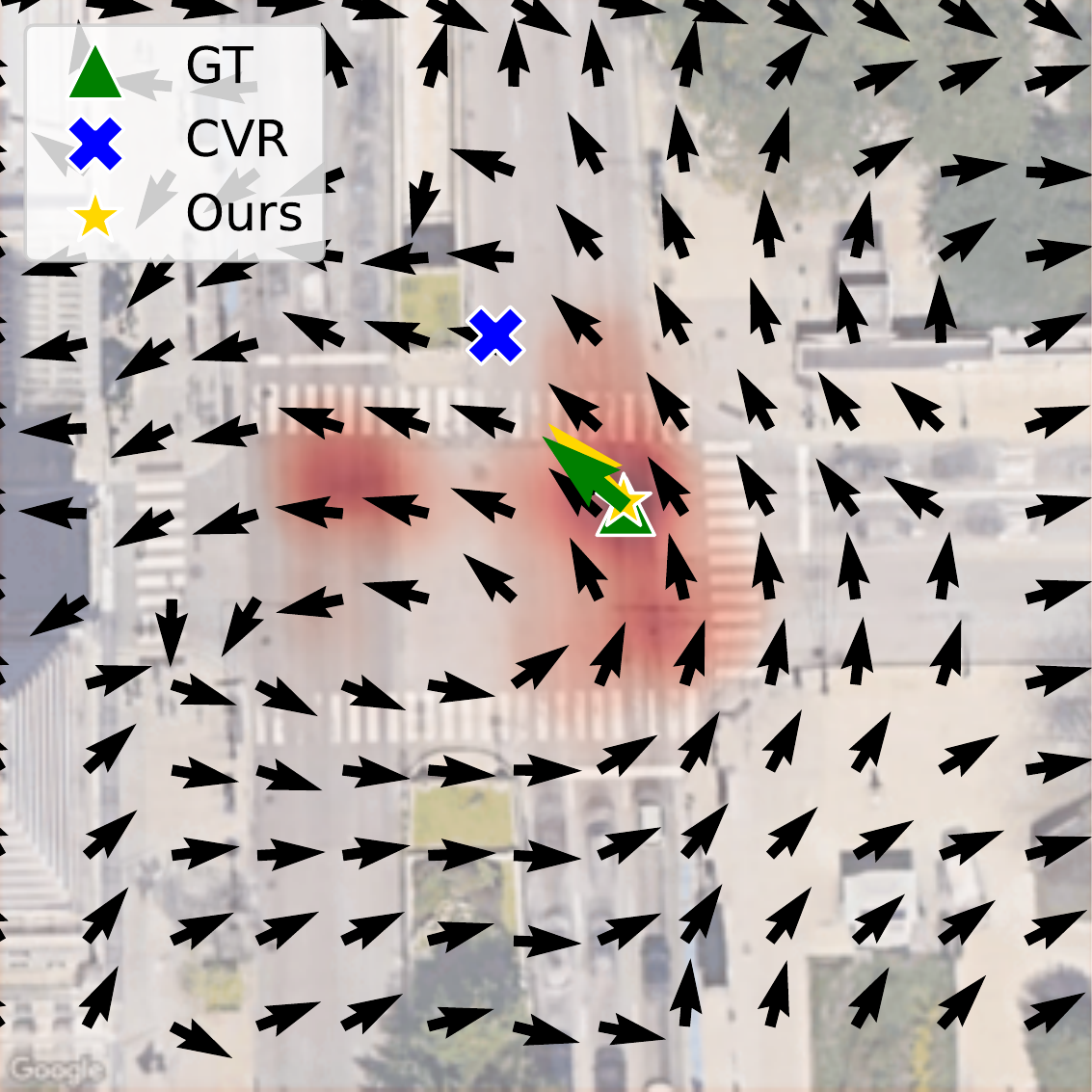}
 \includegraphics[width=0.24\textwidth]{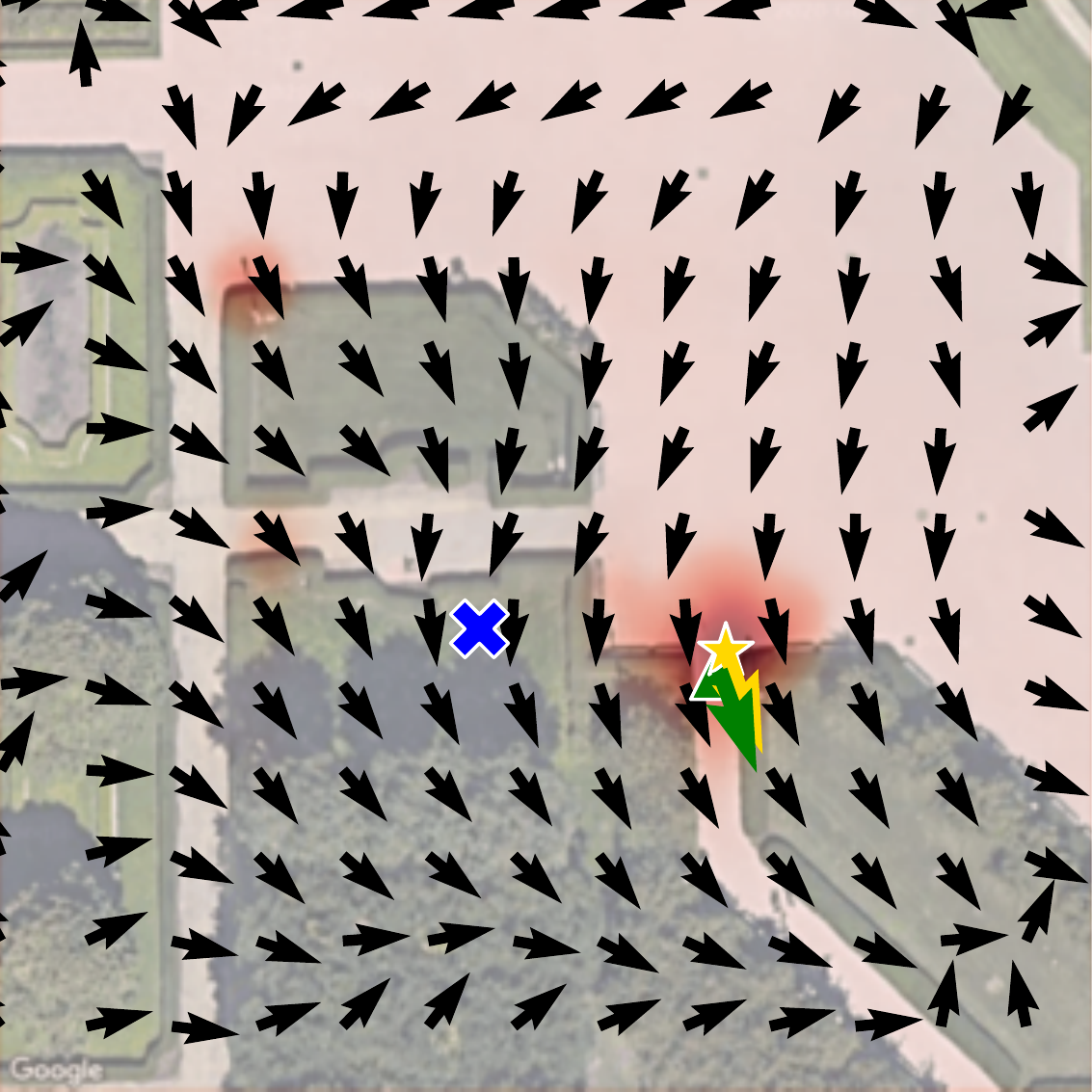}
 \includegraphics[width=0.24\textwidth]{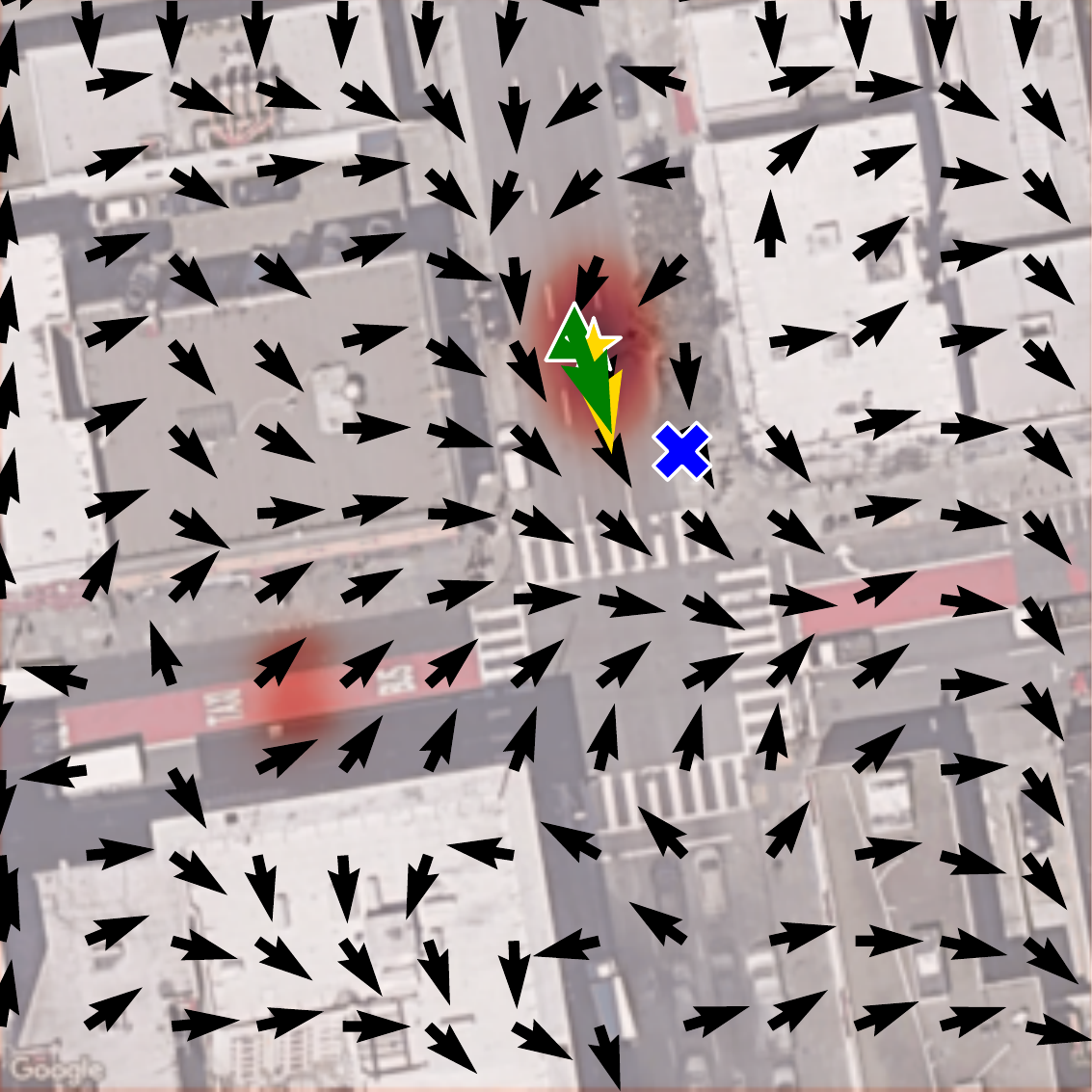}
 \includegraphics[width=0.24\textwidth]{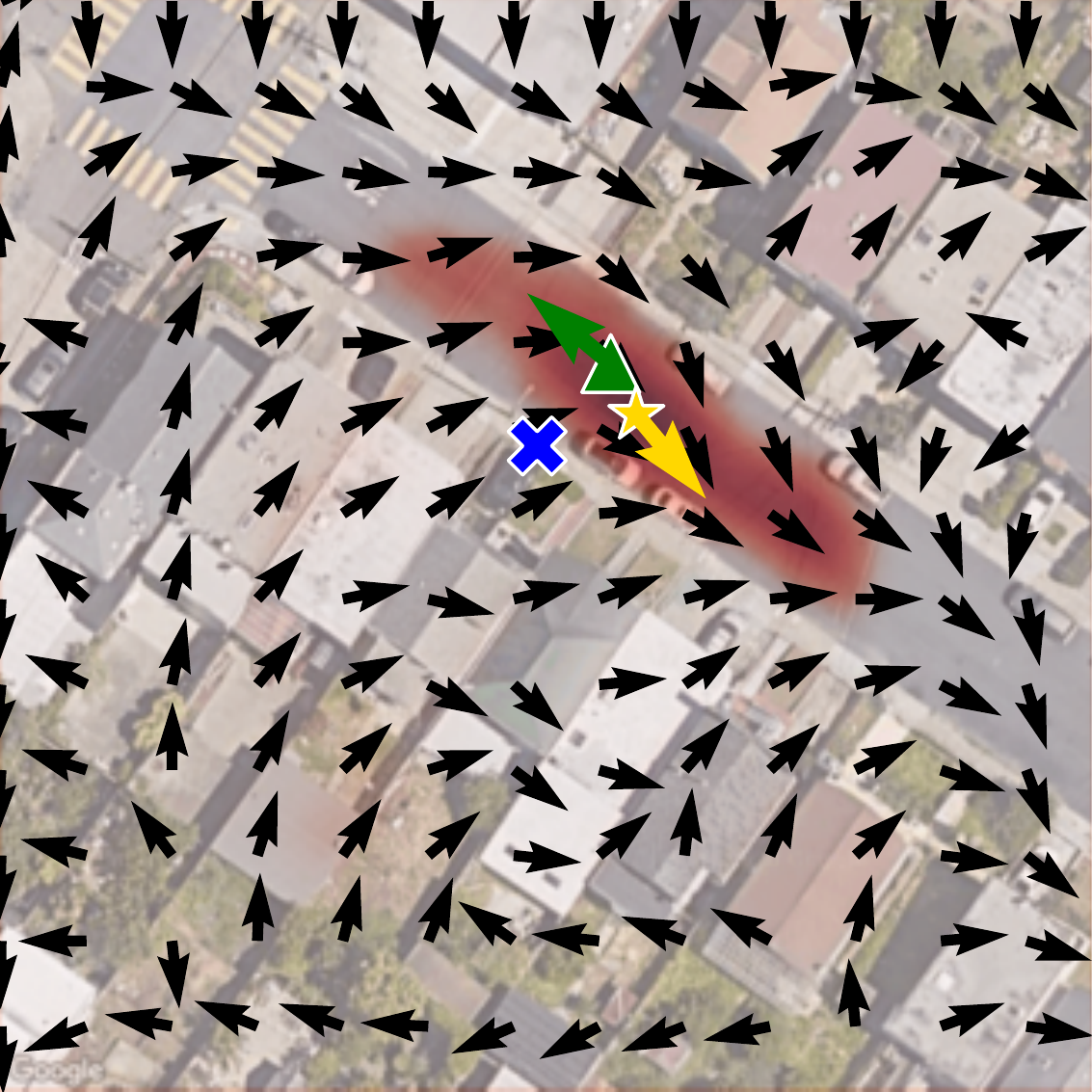} 
\caption{Qualitative results on VIGOR. First two samples are from the Same-Area test set, and the last two samples are from the Cross-Area test set. The first three samples are success cases, the fourth shows a failure case. CVR (blue cross) receives orientation-aligned ground and aerial images and does not estimate the orientation. 
CCVPE (ours) selects the orientation (yellow arrow) from the prediction location (yellow star) and dense orientation map $Y$ (black arrows). The red color shows the localization probability distribution, and the darker the color the higher the probability. The center of the ground image is always the forward direction (green vertical line), which aligns with the true orientation in the aerial view (green arrow).}
\label{fig:qualitative_results_VIGOR}
\end{figure*}

\begin{figure*}[t] 
\centering
 \includegraphics[width=0.24\textwidth]{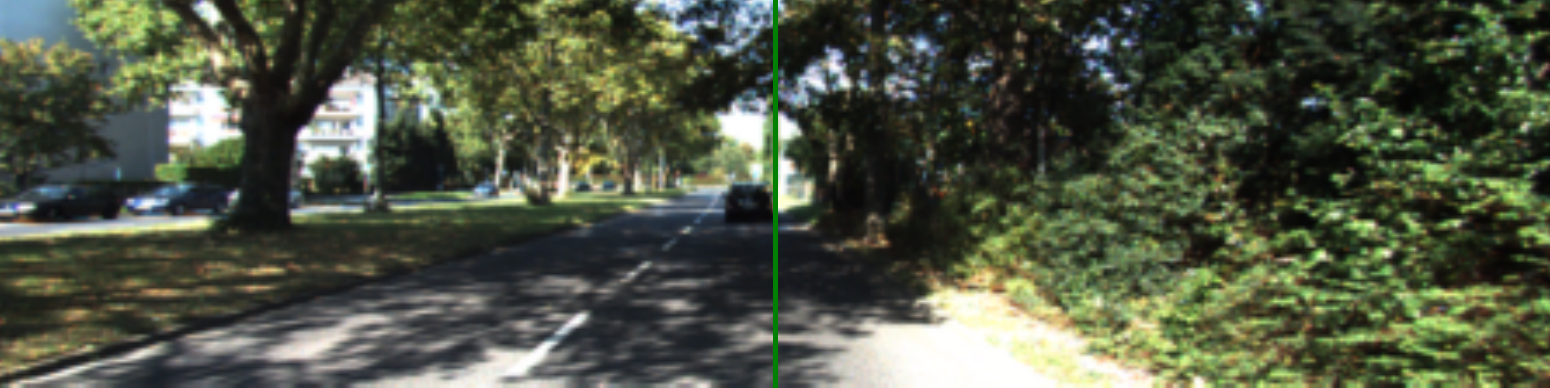}
 \includegraphics[width=0.24\textwidth]{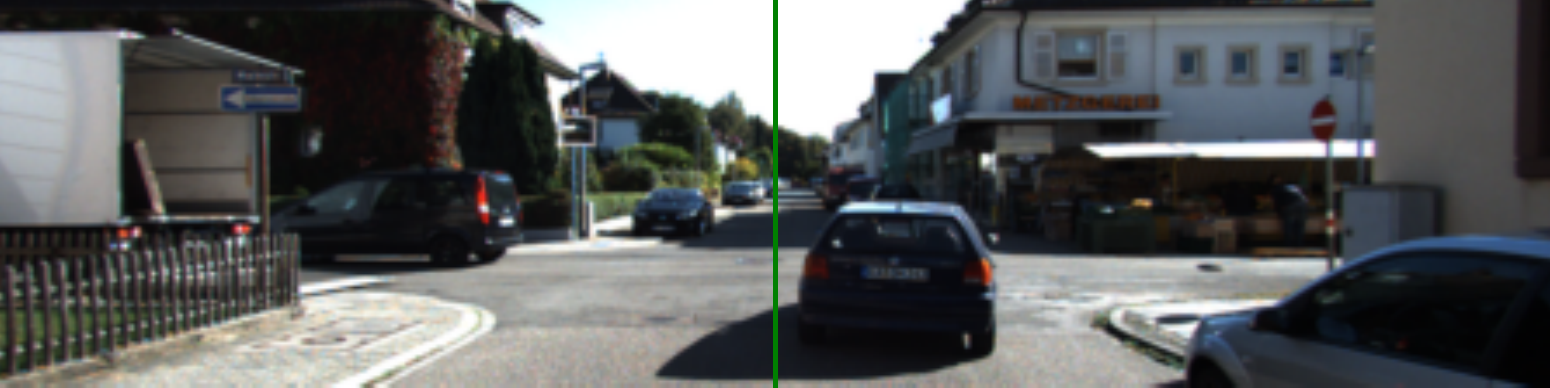}
 \includegraphics[width=0.24\textwidth]{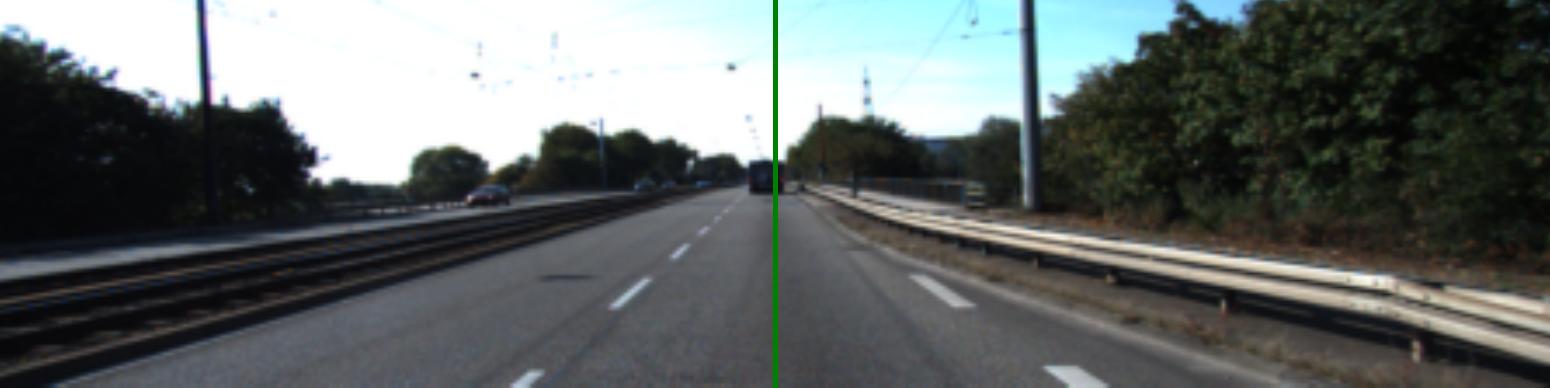} 
 \includegraphics[width=0.24\textwidth]{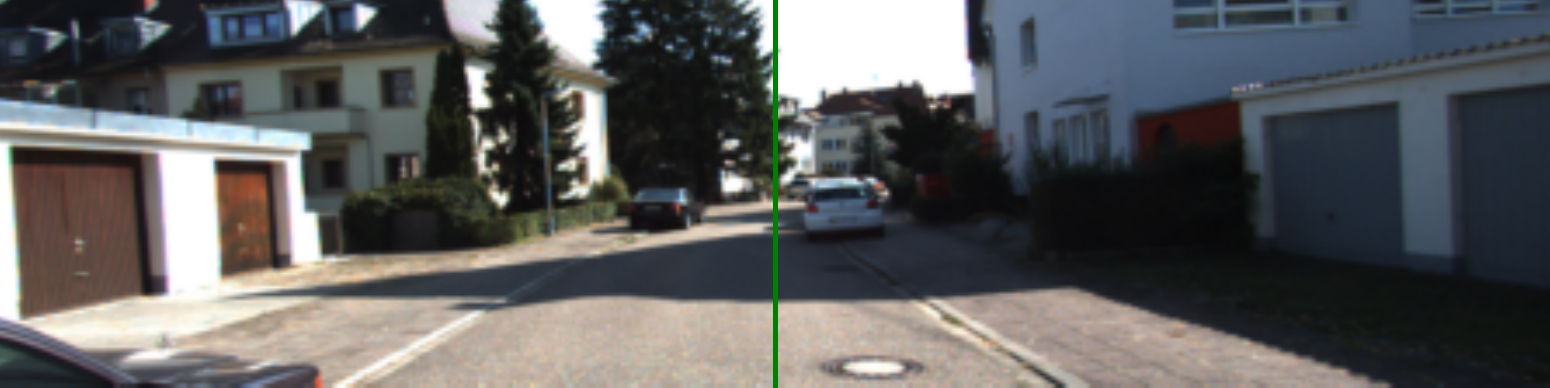}\\
 \vspace{1mm}
 \includegraphics[width=0.24\textwidth]{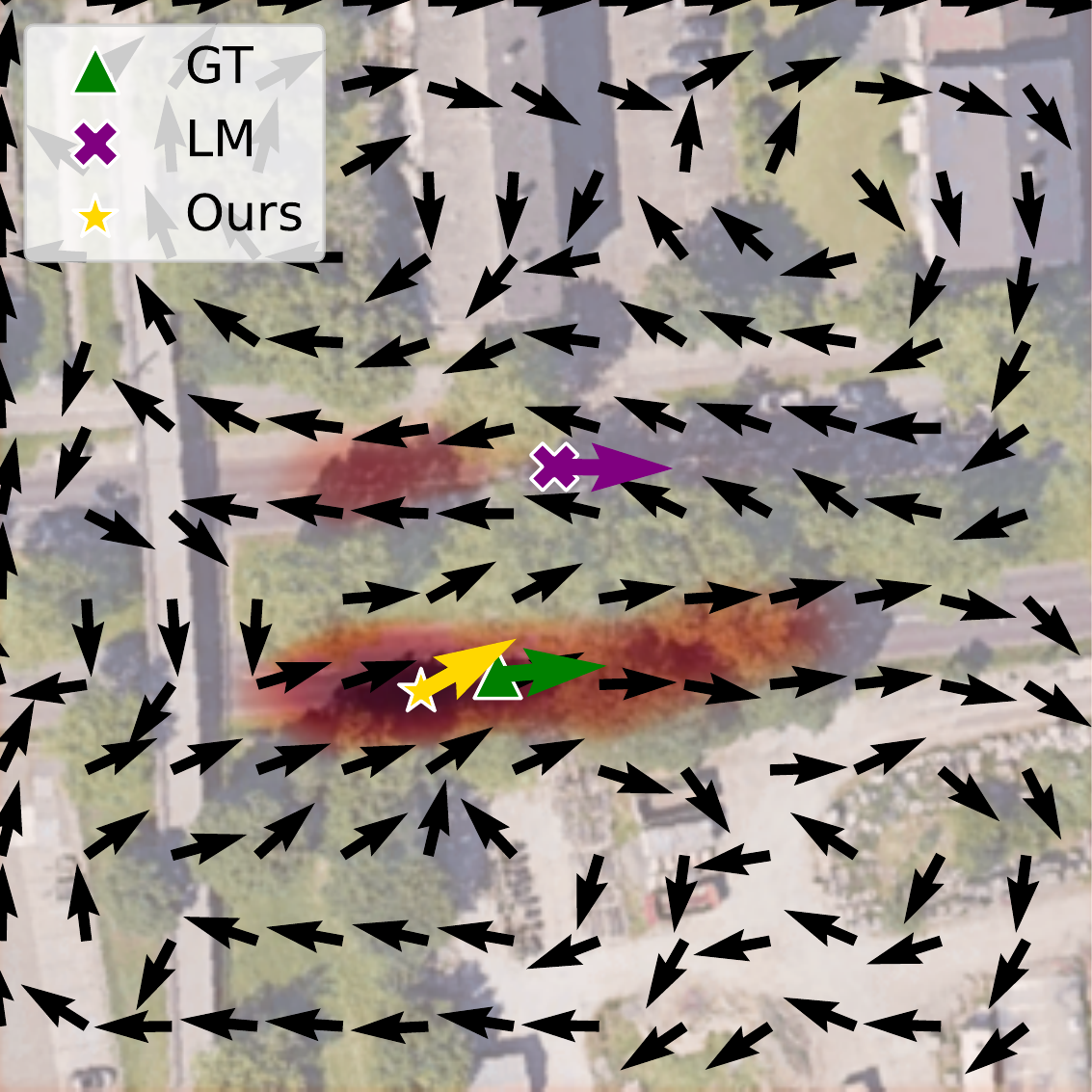}
 \includegraphics[width=0.24\textwidth]{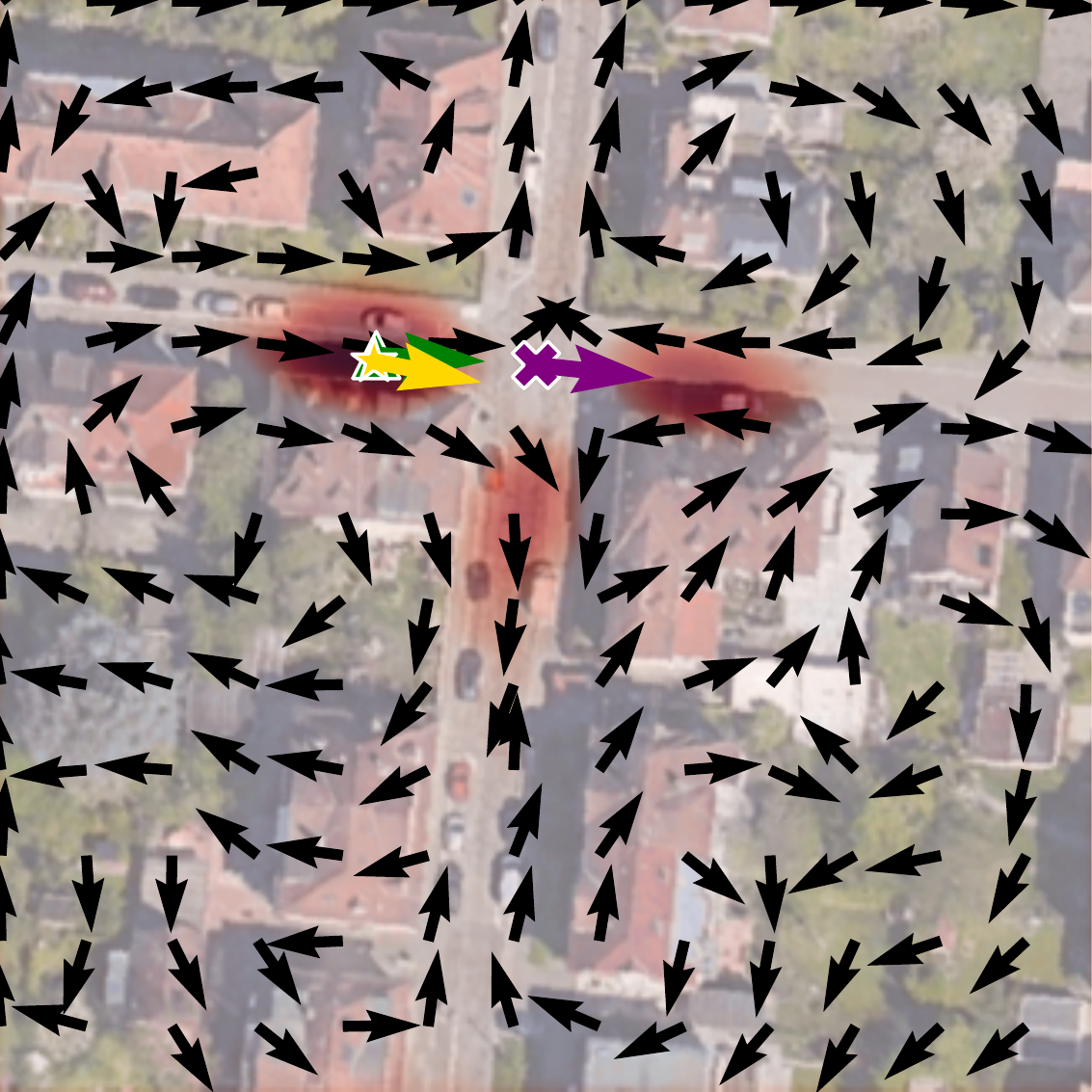}
 \includegraphics[width=0.24\textwidth]{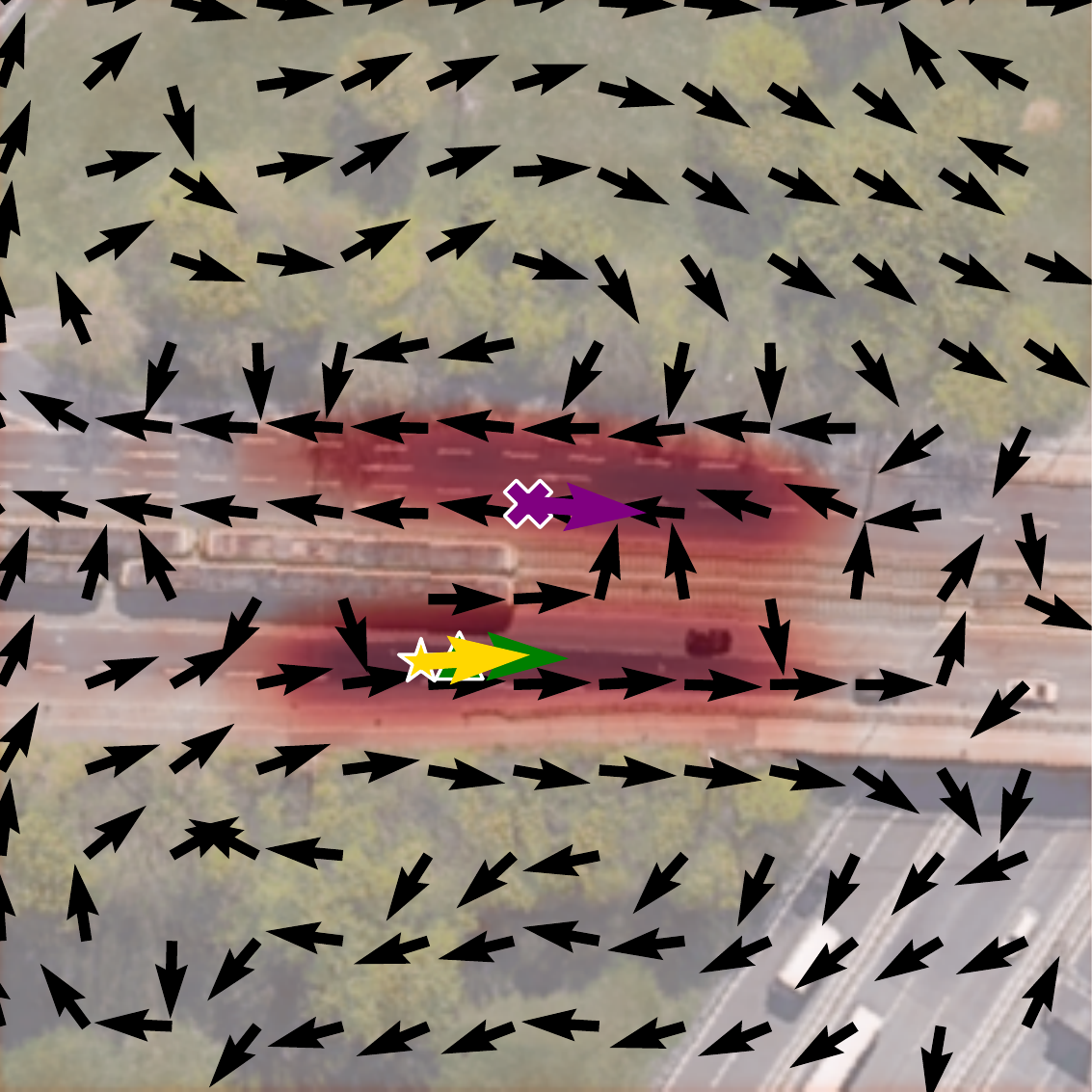}
 \includegraphics[width=0.24\textwidth]{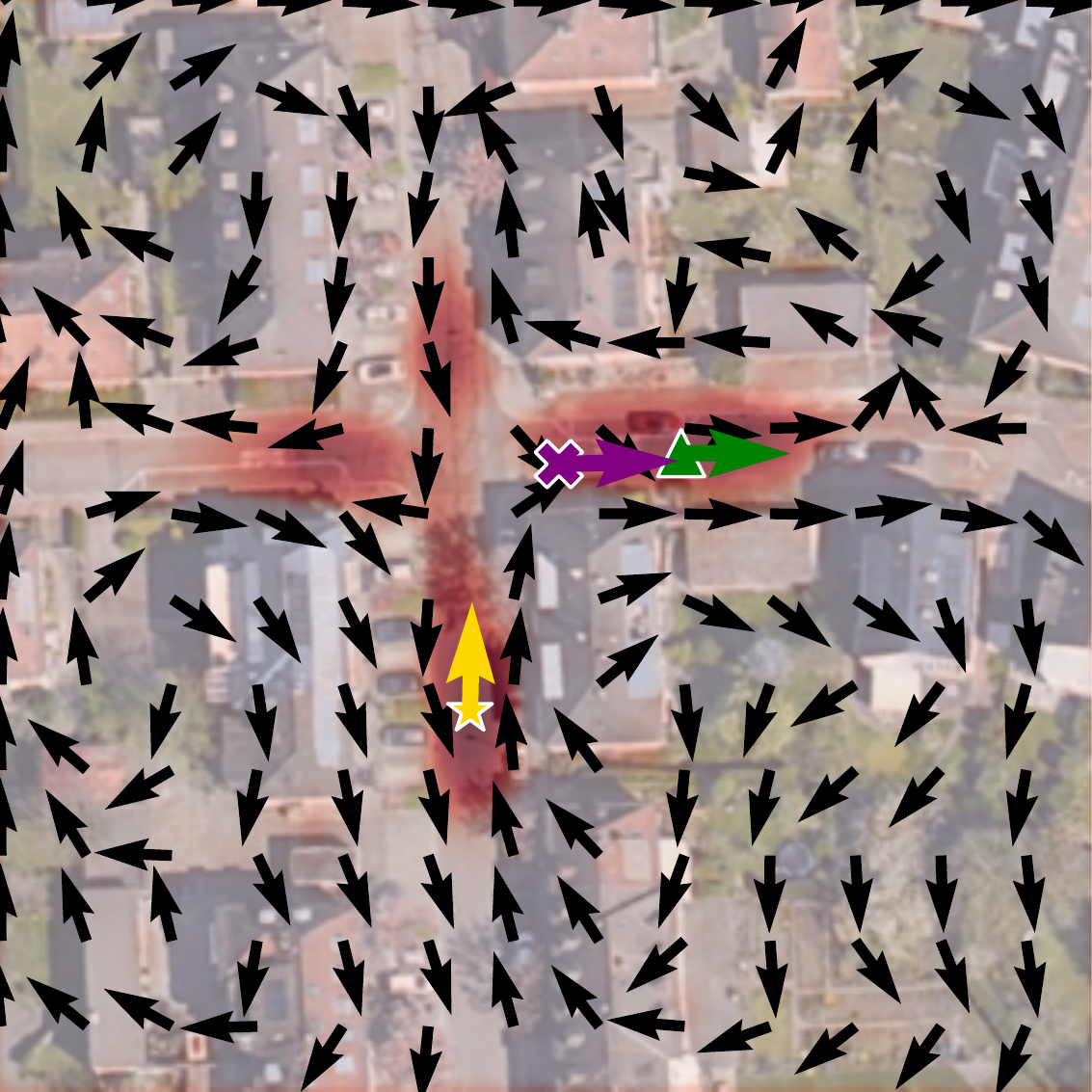} 
\caption{Qualitative results on KITTI. First two samples are from the Same-Area test set, and the last two samples are from the Cross-Area test set. The first three samples are success cases, the fourth one shows a failure case.  
We provide an orientation prior with $\pm 10^\circ$ noise to LM model, and our model does not use an orientation prior.}
\label{fig:qualitative_results_KITTI}
\end{figure*}

\textbf{Pose estimation on KITTI Same-Area:}
As shown in Tab.~\ref{tab:KITTI_test_threshold} Same-Area, camera pose estimation methods, LM~\cite{shi2022beyond}, SliceMatch~\cite{lentsch2022slicematch}, and ours, outperform image retrieval-based methods in terms of percentages of test samples with lateral and longitudinal errors within the given thresholds.
When a $\pm10^\circ$ orientation prior is considered in both training and testing, as assumed in~\cite{shi2022beyond}, our method has a lower mean/median error for both localization ($1.22$~m / $0.62$~m) and orientation estimation ($0.67^\circ / 0.54^\circ$) than LM and SliceMatch.

LM needs an orientation prior to guarantee there is an overlap in the scene between its projected aerial view and the ground view for iterative optimization.
As a result, LM does not work when such an orientation prior is absent, see Tab.~\ref{tab:KITTI_test_threshold}.
Under this more challenging setting, the performance of both SliceMatch and our method degenerates.
Our model still surpasses SliceMatch in localization performance but our model has higher errors in orientation estimation.


\textbf{Qualitative results:}
Compared to single-mode estimators, e.g. regression-based CVR~\cite{zhu2021vigor} and iterative optimization-based LM~\cite{shi2022beyond}, our model shows its advantage especially when the scene contains a symmetric layout.  
As shown in the first two samples in Fig.~\ref{fig:qualitative_results_VIGOR}, when there are multiple visually similar locations, e.g. zebra crossings or junctions, CVR regresses to a location between them.
The iterative method LM sometimes converges to a wrong local optimum, e.g. another road, see Fig.~\ref{fig:qualitative_results_KITTI}.
Our model expresses its uncertainty with its multi-modal distribution to capture all probable modes, usually identifying the correct location in all datasets.

Benefiting from our joint consideration of localization and orientation, when there are multiple probable locations in our prediction, our model predicts for each of these locations the most likely orientation.
As shown in the first image pair in Fig.~\ref{fig:qualitative_results_VIGOR}, the orientation of the camera (the green line in the ground image) roughly points to the end of a zebra crossing.
Our prediction suggests multiple locations, and each location predicts an orientation pointing to an end of a different zebra crossing.

At locations with a low localization probability, the ground-aerial descriptor matching has low similarity scores in all orientation channels.
In this case, the orientation prediction is influenced less by the descriptor matching score but appears to follow a learned prior from the aerial view.
Importantly, we show in our ablation study that explicitly providing the descriptor matching scores is still key to good orientation prediction.
\newtext{Besides, we also visualize the learned features of our model using Grad-CAM~\cite{selvaraju2017grad} in our Supplementary Material.}

\begin{table*}[ht]
\caption{Evaluation on KITTI dataset. \textbf{Best in bold.} We report mean and median localization and orientation error, as well as the percentage of test samples that have lateral/longitudinal localization or orientation error less than a threshold. The evaluation of image retrieval methods (noted with `retrieval') is collected from~\cite{shi2022accurate}.
\oldtext{In the left column, $\pm 10^\circ$ and w/o. denotes training and testing with an orientation prior with noise between $-10^\circ$ and $10^\circ$, and training and testing without an orientation prior respectively.}
\newtext{In the leftmost column, $\pm 10^\circ$ denotes training and testing with an orientation prior with noise in the range $[-10^\circ, 10^\circ]$, and $360^\circ$ means no orientation prior by using noise from the $360^\circ$ circular domain.}}
    \label{tab:KITTI_test_threshold}
    \centering
    \begin{tabular}{|P{0.5cm}|P{2.1cm}|P{0.9cm}P{0.9cm}|P{0.6cm}P{0.6cm}P{0.6cm}|P{0.6cm}P{0.6cm}P{0.6cm}|P{0.9cm}P{0.9cm}|P{0.6cm}P{0.6cm}P{0.6cm}|}
    \hline
    \multicolumn{2}{|c|}{\multirow{2}{*}{Same-Area}} & \multicolumn{2}{c|}{$\downarrow$ Localization (m)}  & \multicolumn{3}{c|}{$\uparrow$ Lateral (\%)}  & \multicolumn{3}{c|}{$\uparrow$ Longitudinal (\%)} & \multicolumn{2}{c|}{$\downarrow$ Orientation ($^\circ$)} & \multicolumn{3}{c|}{$\uparrow$ Orientation (\%)}  \\
    \multicolumn{2}{|c|}{} & mean & median & 1m&3m&5m &  1m&3m&5m & mean & median & $1^\circ$&$3^\circ$&$5^\circ$\\
    \hline
    \multirow{5}{*}{\rotatebox{90}{retrieval}} & CVM-Net \cite{CVM-Net} & - & - & 5.83 & 17.41 & 28.78 & 3.47 &11.18 & 18.42 & - & - & -& -& -\\
    \cline{2-15}
    & CVFT~\cite{shi2020optimal} & - & - & 7.71 & 22.37 & 36.28 & 3.82 &11.48 & 18.63 & - & - & -& -& -\\
    \cline{2-15}
    & SAFA~\cite{SAFA} & - & - & 9.49 & 29.31 & 46.44 & 4.35 &12.46 & 21.10 & - & - & -& -& -\\
    \cline{2-15}
    & Polar-SAFA~\cite{SAFA} & - & - & 9.57 & 30.08 & 45.83 & 4.56 &13.01 & 21.12 & - & - & -& -& -\\
    \cline{2-15}
    & DSM~\cite{shi2020looking} & - & - & 10.12 & 30.67 & 48.24 & 4.08 &12.01 & 20.14 & - & - & 3.58& 13.81& 24.44\\
    \hline
    \multirow{3}{*}{\rotatebox{90}{$\pm10^\circ$}} & LM~\cite{shi2022accurate} & 12.08 & 11.42 & 35.54 & 70.77 & 80.36 & 5.22 &15.88 & 26.13 & 3.72 & 2.83 & 19.64& 51.76& 71.72\\
    \cline{2-15}
    & SliceMatch~\cite{lentsch2022slicematch} & 7.96 & 4.39 & 49.09 & 91.76 & 98.52 & 15.19 & 49.99 & 57.35 & 4.12 & 3.65 & 13.41& 42.62 & 64.17\\
    \cline{2-15}
    & \textit{CCVPE} (ours) & \textbf{1.22} & \textbf{0.62} & \textbf{97.35} & \textbf{98.65} & \textbf{99.71} & \textbf{77.13} & \textbf{96.08} & \textbf{97.16} & \textbf{0.67} & \textbf{0.54} & \textbf{77.39} & \textbf{99.47} & \textbf{99.95} \\
    \hline
    \multirow{3}{*}{\rotatebox{90}{\newtext{$360^\circ$}}} 
    & LM~\cite{shi2022accurate} & 15.51 & 15.97 & 5.17 & 15.13 & 25.44 & 4.66 & 15.00 & 25.39 & 89.91 & 90.75 & 0.61 & 1.88 & 2.89\\
    \cline{2-15}
    & SliceMatch~\cite{lentsch2022slicematch} & 9.39 & 5.41 & 39.73 & \textbf{80.56} & \textbf{87.92} & 13.63 & 40.75& 49.22 & \textbf{8.71} & \textbf{4.42} & \textbf{11.35} & \textbf{36.23} & \textbf{55.82}\\
    \cline{2-15}
    & \textit{CCVPE} (ours) & \textbf{6.88} & \textbf{3.47} & \textbf{53.30} & 77.63 & 85.13 & \textbf{25.84} & \textbf{55.05} & \textbf{68.49} & 15.01 & 6.12 & 8.96 & 26.48 & 42.75\\
    \hline
    \hline
    \multicolumn{2}{|c|}{\multirow{2}{*}{Cross-Area}} & \multicolumn{2}{c|}{$\downarrow$ Localization (m)}  & \multicolumn{3}{c|}{$\uparrow$ Lateral (\%)}  & \multicolumn{3}{c|}{$\uparrow$ Longitudinal (\%)} & \multicolumn{2}{c|}{$\downarrow$ Orientation ($^\circ$)} & \multicolumn{3}{c|}{$\uparrow$ Orientation (\%)}  \\
    \multicolumn{2}{|c|}{} & mean & median & 1m&3m&5m &  1m&3m&5m & mean & median & $1^\circ$&$3^\circ$&$5^\circ$\\
    \hline
    \multirow{5}{*}{\rotatebox{90}{retrieval}} & CVM-Net \cite{CVM-Net} & - & - & 6.96 & 21.55 & 35.24 & 3.58 &10.45 & 17.53 & - & - & -& -& -\\
    \cline{2-15}
    & CVFT~\cite{shi2020optimal} & - & - & 7.20 & 22.05 & 36.21 & 3.63 & 11.11& 18.46 & - & - & -& -& -\\
    \cline{2-15}
    & SAFA~\cite{SAFA} & - & - & 9.15 & 27.83 & 44.27 & 4.22 &11.93 & 19.65 & - & - & -& -& -\\
    \cline{2-15}
    & Polar-SAFA~\cite{SAFA} & - & - & 10.02 & 29.09 & 46.19 & 3.82 & 11.87& 19.84 & - & - & -& -& -\\
    \cline{2-15}
    & DSM~\cite{shi2020looking} & - & - & 10.77 & 31.37 & 48.24 & 3.87 &11.73 & 19.50 & - & - & 3.53& 14.09& 23.95\\
    \hline
    \multirow{3}{*}{\rotatebox{90}{$\pm10^\circ$}} & LM~\cite{shi2022accurate} & 12.58 & 12.11 & 27.82 & 59.79 & 72.89 & 5.75 & 16.36& 26.48 & 3.95 & 3.03 & 18.42& 49.72& 71.00\\
    \cline{2-15}
    & SliceMatch~\cite{lentsch2022slicematch} & 13.50 & 9.77 & 32.43 & 78.98 & 86.44 & 8.30 & 24.48 & 35.57 & 4.20 & 6.61 & 46.82& 46.82& 46.82\\
    \cline{2-15}
    & \textit{CCVPE} (ours) & \textbf{9.16} & \textbf{3.33} & \textbf{44.06} & \textbf{81.72} & \textbf{90.23} & \textbf{23.08} & \textbf{52.85} & \textbf{64.31} & \textbf{1.55} & \textbf{0.84} & \textbf{57.72} & \textbf{92.34} & \textbf{96.19} \\
    \hline
    \multirow{3}{*}{\rotatebox{90}{\newtext{$360^\circ$}}} 
    & LM~\cite{shi2022accurate} & 15.50 & 16.02 & 5.60 & 16.02 & 25.60 & 5.64 & 15.86 & 25.76 & 89.84 & 89.85 & 0.60 & 1.60 & 2.65\\
    \cline{2-15}
    & SliceMatch~\cite{lentsch2022slicematch} & 14.85 & 11.85 & \textbf{24.00} & \textbf{62.52} & \textbf{72.89} & 7.17 & 26.11 & 33.12 & \textbf{23.64} & \textbf{7.96} & \textbf{31.69} & \textbf{31.69} & \textbf{31.69}\\
    \cline{2-15}
    & \textit{CCVPE} (ours) & \textbf{13.94} & \textbf{10.98} & 23.42 & 49.15 & 60.46 & \textbf{11.81} & \textbf{29.85} & \textbf{42.12} & 77.84 & 63.84 & 3.14 & 9.24 & 14.56\\
    \hline
    \end{tabular}
\end{table*}

\textbf{Probabilistic prediction:}
Next, we evaluate the probability estimation of the baselines and our model on the VIGOR dataset.
CVR~\cite{zhu2021vigor} and Eff-CVR regress to a single location without any probability estimation.
We fit a zero-mean Gaussian distribution on their predicted errors.
The standard deviation of this Gauss is calculated based on their localization error on the validation set.

As shown in Tab.~\ref{tab:VIGOR_test}, our model has considerably higher mean and median probability at the ground truth location than CVR~\cite{zhu2021vigor}, Eff-CVR, and SliceMatch~\cite{lentsch2022slicematch}.
During training, SliceMatch is optimized for discriminative descriptors, while our model is directly optimized for high probability at the ground truth location by our cross-entropy loss.
In general, our model is less likely to miss the ground truth location, which is an important aspect when the outputs are temporal filtered or fused with other sensor measurements.

\begin{figure*}[t] 
\centering
 \includegraphics[width=0.33\textwidth]{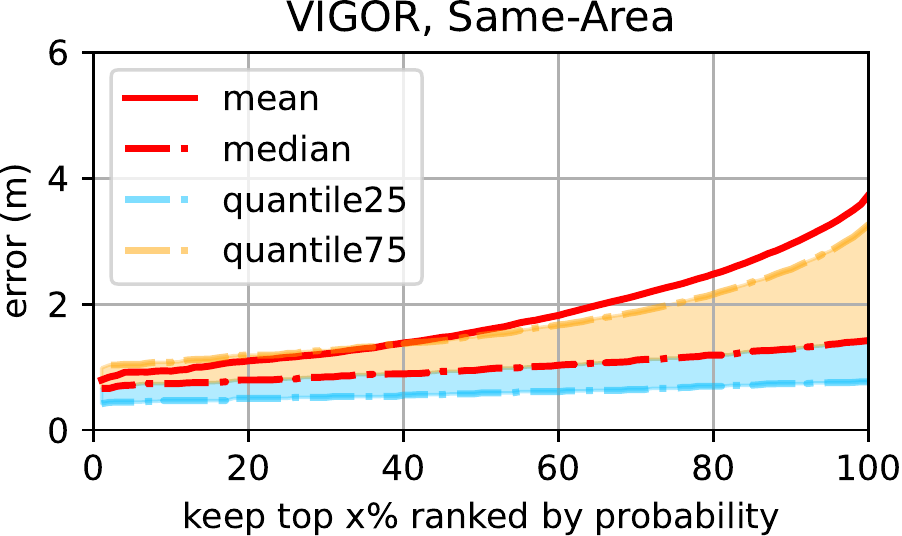}
 \includegraphics[width=0.33\textwidth]{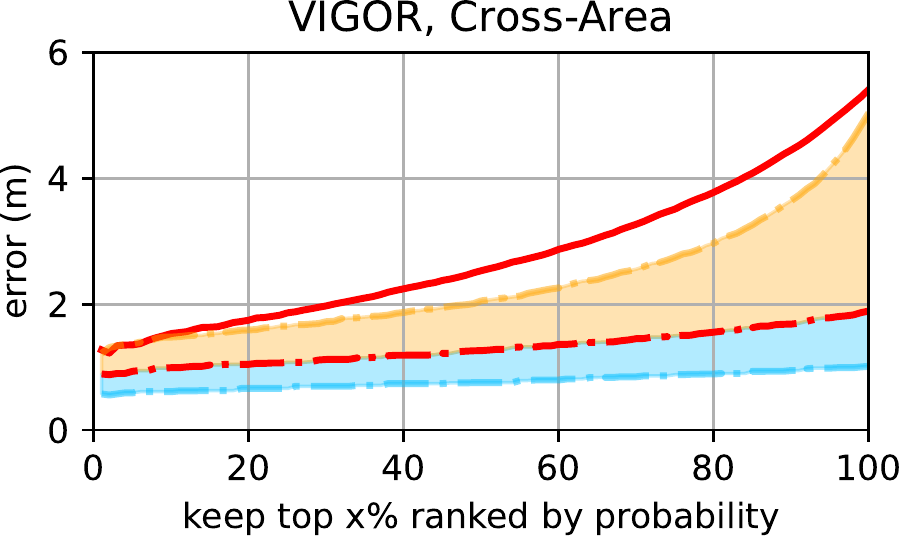} 
 \includegraphics[width=0.33\textwidth]{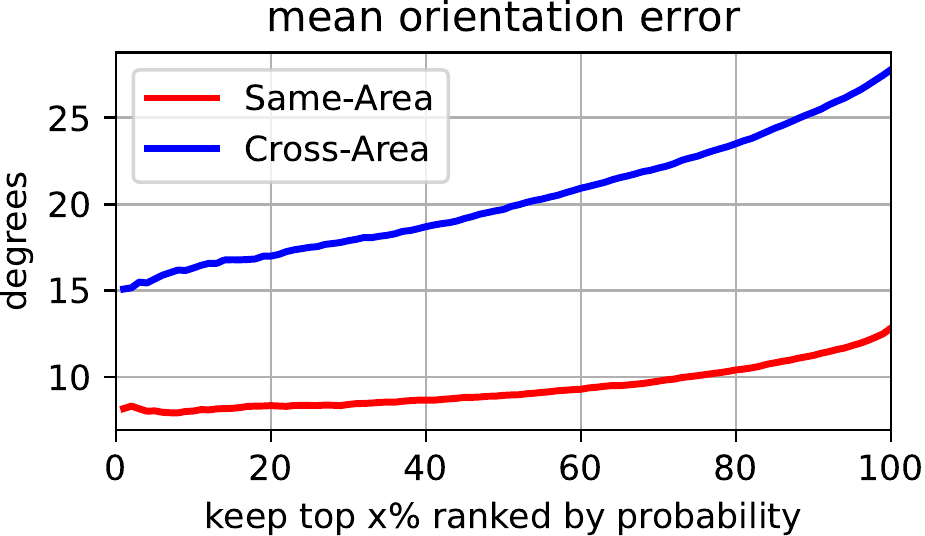} 
\caption{Ranking CCVPE's predictions based on their estimated probabilities (tested with unknown orientation). The more confident the prediction, the lower the localization and orientation error.
} 
\label{fig:error_prob_VIGOR}
\end{figure*}

Importantly, our probabilistic output can be used to identify predictions that potentially have large localization and orientation errors.
Because we construct orientation-aware descriptors, the better an aerial descriptor matches the ground descriptor, the more likely both the location and orientation of that aerial descriptor are correct. 
Therefore, the localization probability can be used to filter the orientation error as well.
As shown in Fig.~\ref{fig:error_prob_VIGOR}, when we rank the predictions based on their predicted probabilities, the more confident predictions have in general lower localization and orientation errors.
This property is important in safety-crucial applications such as autonomous driving.

\subsection{Generalization to new measurements across areas}
Here we consider use cases that target operation in areas that were not covered specifically by the training data, such as driving in different cities or suburban areas.

Overall, we see a similar trend in model comparison in the Cross-Area setting as in the Same-Area setting.
On VIGOR Cross-Area test set, see Tab.~\ref{tab:VIGOR_test}, our model surpasses the previous SOTA SliceMatch~\cite{lentsch2022slicematch} in localization by a large margin.
When orientation is unknown, our median error is $66\%$ lower than that of SliceMatch.
However, our orientation error is higher than that of SliceMatch.
On KITTI Cross-Area test set, see Tab.~\ref{tab:KITTI_test_threshold}, when an orientation prior with $\pm 10^\circ$ noise presents during training and testing, our model surpasses both LM~\cite{shi2022beyond} and SliceMatch~\cite{lentsch2022slicematch} in both localization and orientation.
Without this prior, our model has lower mean and median localization error than SliceMatch, but our orientation error is higher.

Unsurprisingly, compared to the performance on the Same-Area test set, there is a performance degradation for all models.
Our model could learn priors from the scene layout in the aerial image to guide its predictions.
It becomes more challenging when the test aerial images are unseen.
Since our predicted orientation is selected at the predicted location, the orientation error is also likely to be large when localization is wrong, see Fig.~\ref{fig:qualitative_results_KITTI} last sample.
We also observe there are more samples that have predicted orientation in the opposite direction on the Cross-Area test set than the Same-Area test set on both VIGOR and KITTI datasets.
See the last sample of Fig.~\ref{fig:qualitative_results_VIGOR} for an example.
In practice, when there is a prior in orientation, e.g. identified by the driving direction, our model can make use of the prior to improve its prediction without retraining.
We will demonstrate this in the next sub-section.

\subsection{Effects on orientation prior and image's FoV}
Next, we study our model's behavior on both VIGOR Same-Area and Cross-Area test sets for inference with an orientation prior and ground images with  different horizontal FoV.

\begin{figure}[t] 
\centering
 \includegraphics[width=0.35\textwidth]{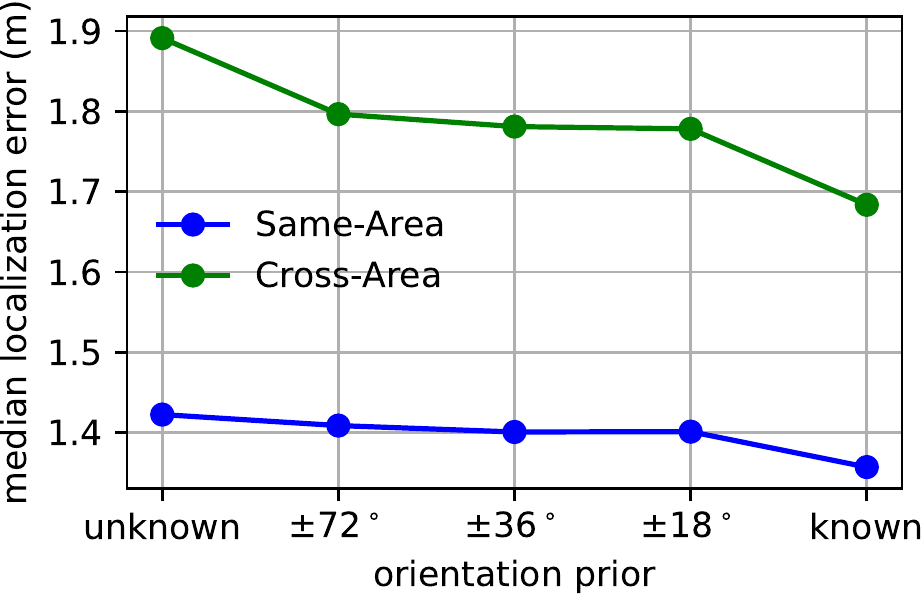} 
\caption{Localization with different orientation priors on VIGOR Same-Area and Cross-Area test set. 
} 
\label{fig:error_orientation_prior_VIGOR}
\end{figure}

\begin{figure*}[t] 
\centering
 \includegraphics[width=0.24\textwidth]{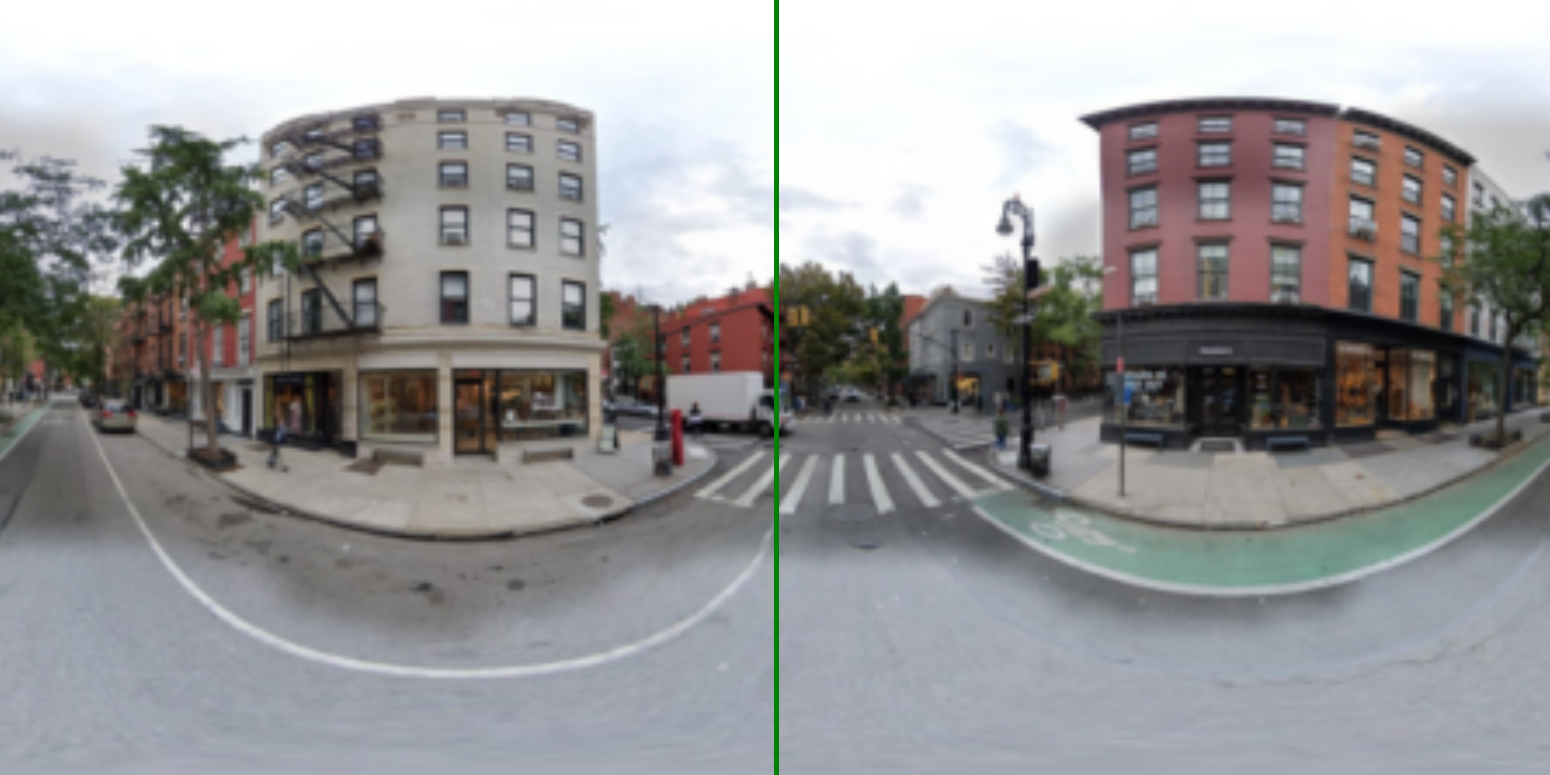} 
 \includegraphics[width=0.24\textwidth]{figures/experiments/grd_samearea_521.pdf}
 \includegraphics[width=0.24\textwidth]{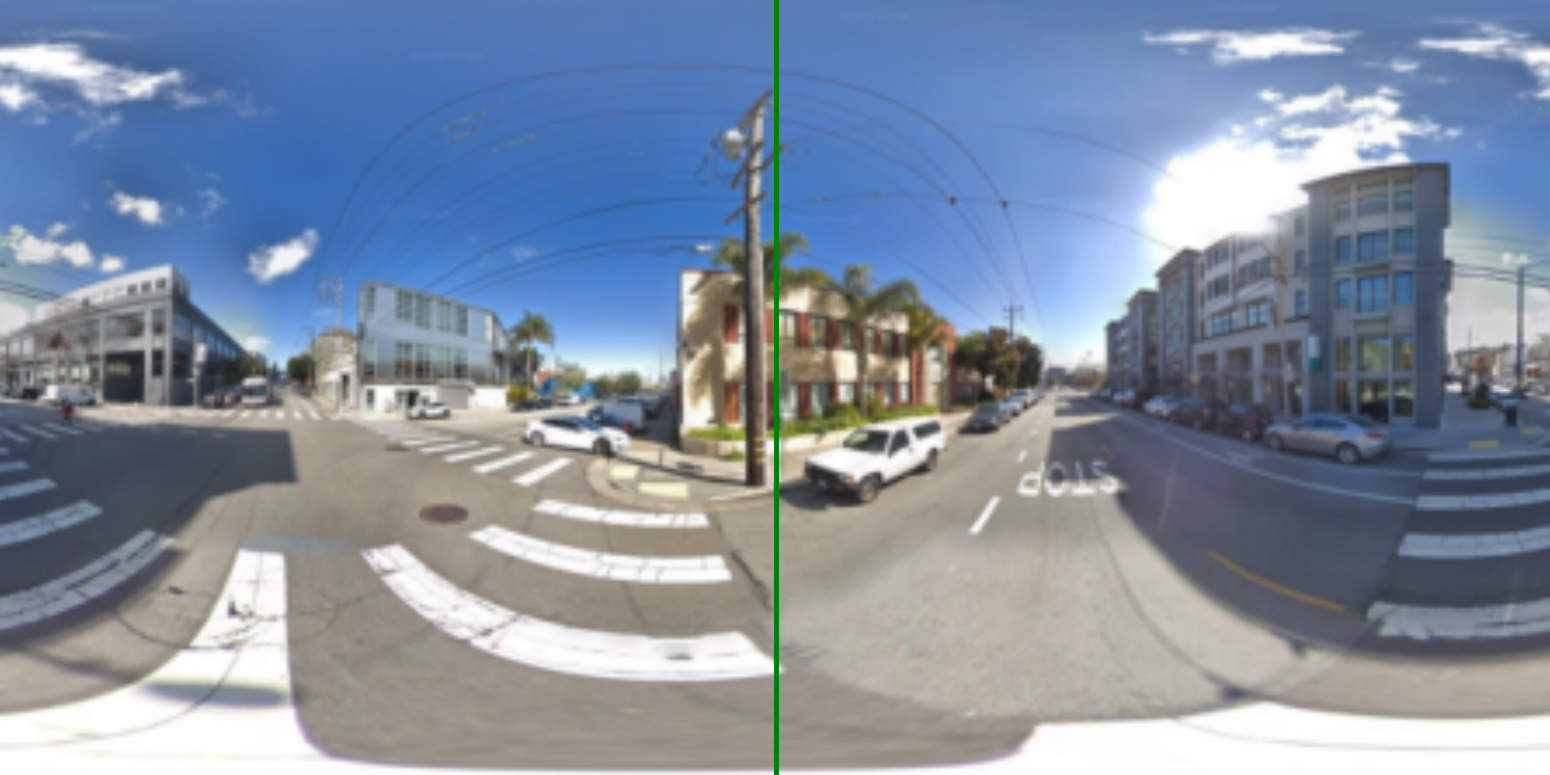} 
 \includegraphics[width=0.24\textwidth]{figures/experiments/grd_crossarea_576.pdf} \\
 \vspace{1mm}
 \includegraphics[width=0.24\textwidth]{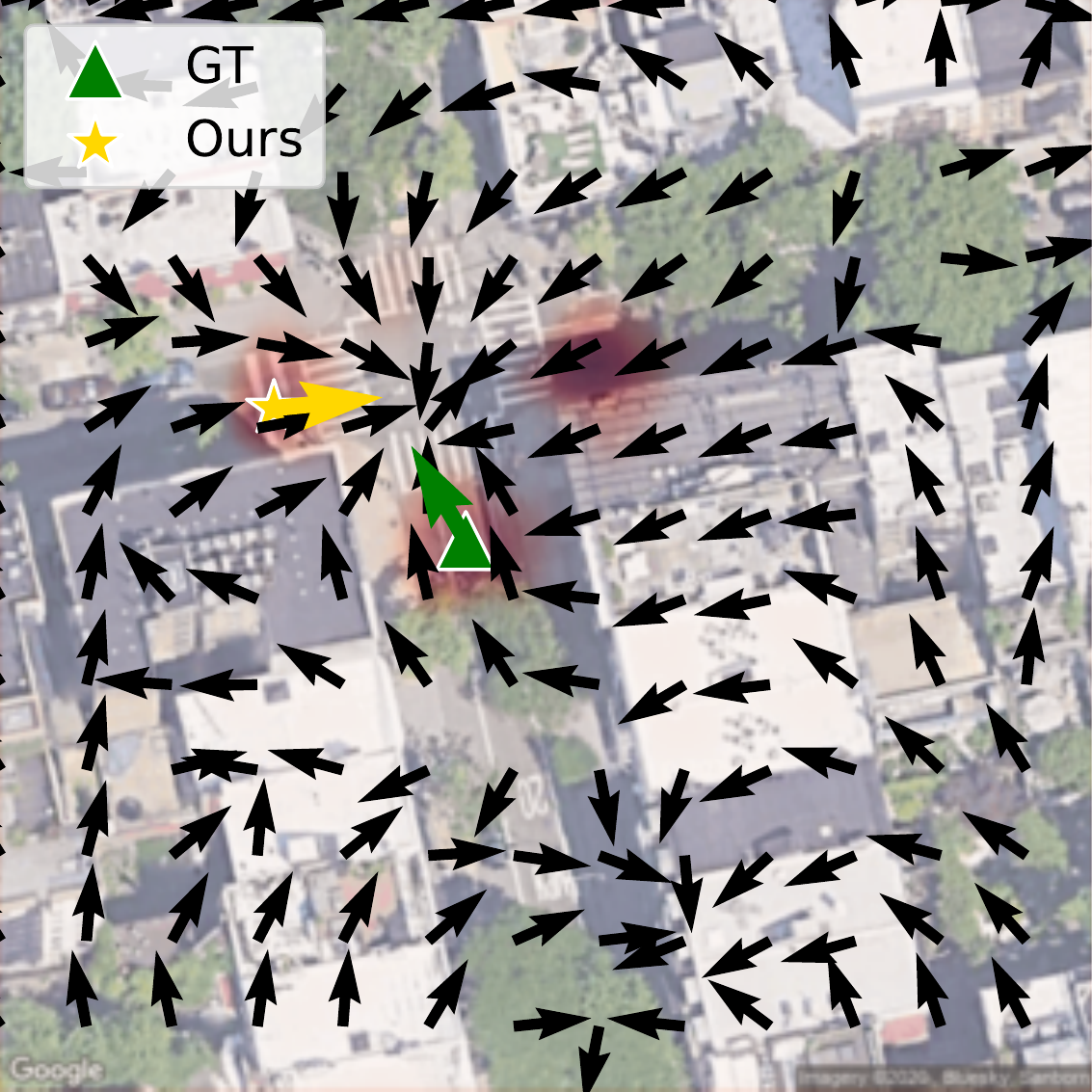}
 \includegraphics[width=0.24\textwidth]{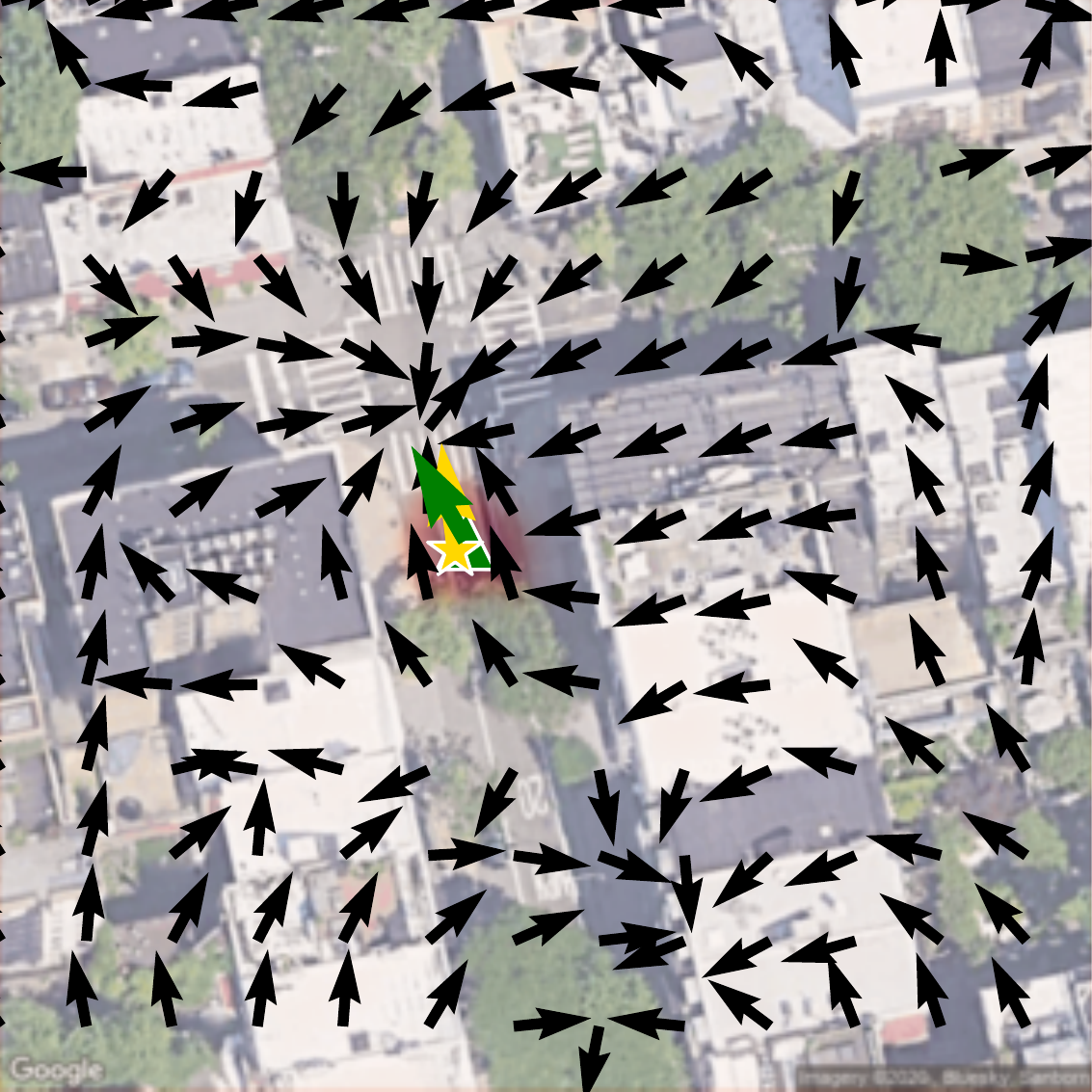} 
 \includegraphics[width=0.24\textwidth]{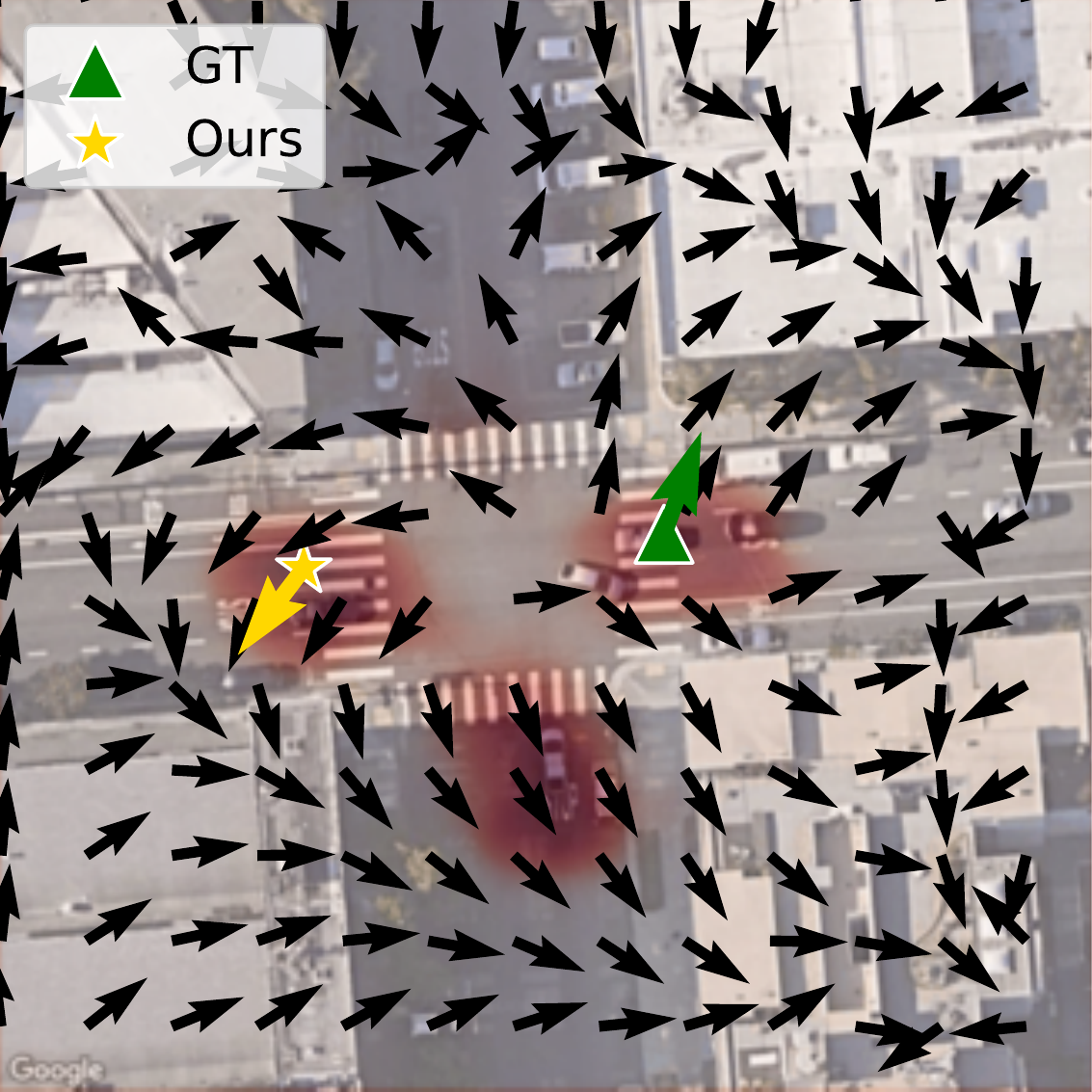}
 \includegraphics[width=0.24\textwidth]{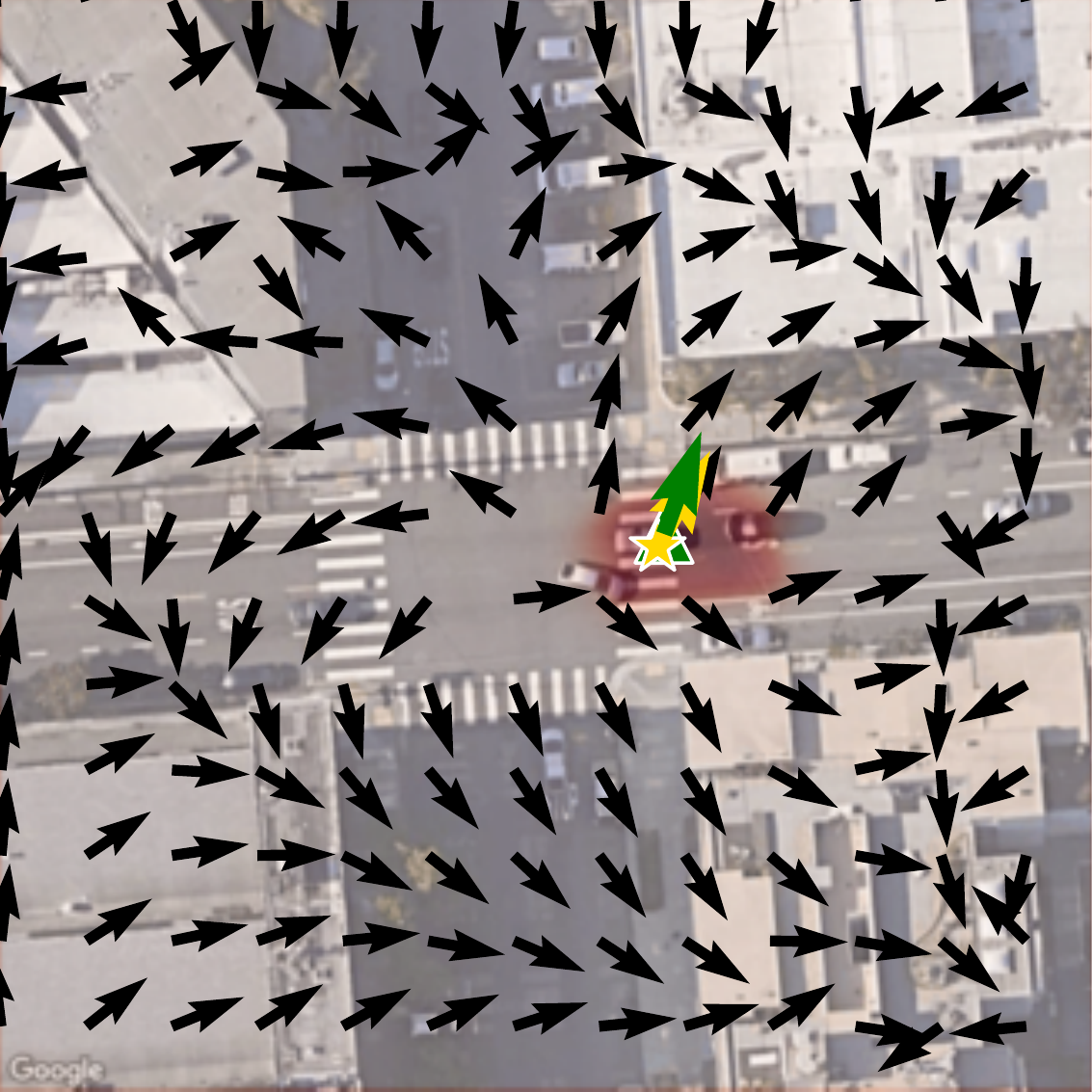} 
\caption{An orientation prior improves our localization performance on VIGOR, Same-Area (first two image pairs) and Cross-Area (last two image pairs). The first and third image: inference without an orientation prior. The second and fourth image: inference with an orientation prior containing noise between $-36^\circ$ and $36^\circ$.
With the prior, locations that expect a different orientation become improbable.
}
\label{fig:qualitative_results_VIGOR_ori_prior}
\end{figure*}

\textbf{Inference with an orientation prior:}
As described in Sec.~\ref{sec:descriptor_matching_modules}, our model can make use of an orientation prior without retraining.
Fig.~\ref{fig:error_orientation_prior_VIGOR} shows that when a more accurate orientation prior is present, the localization performance increases accordingly.
When there are multiple locations in the aerial image that match the ground image with different orientations, for example, at a crossroad, providing such an orientation prior effectively reduces the wrong matchings in our LMU and OMU modules, see examples in Fig.~\ref{fig:qualitative_results_VIGOR_ori_prior}.

\begin{figure}[t] 
\centering
 \includegraphics[width=0.35\textwidth]{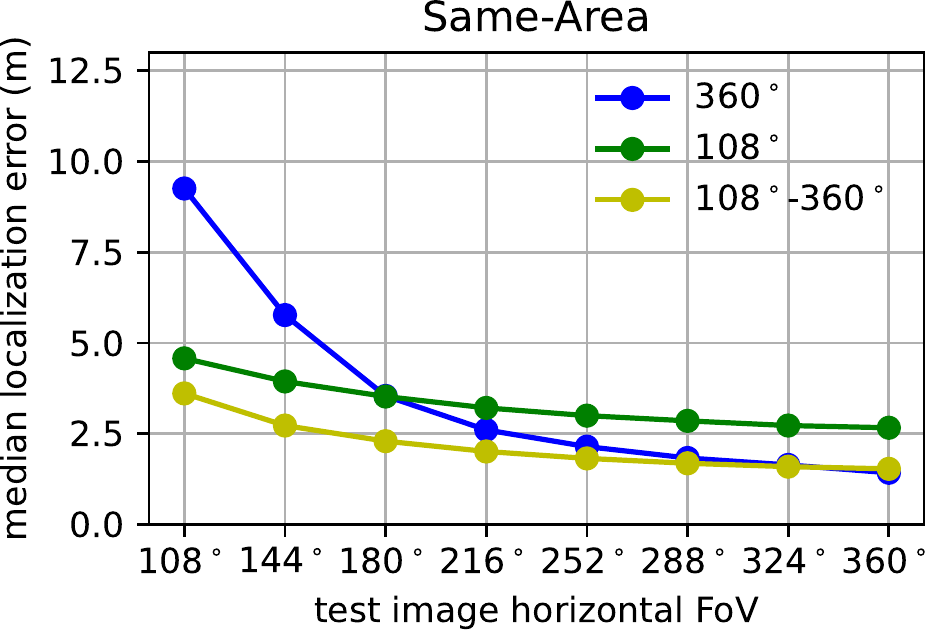} 
 \includegraphics[width=0.35\textwidth]{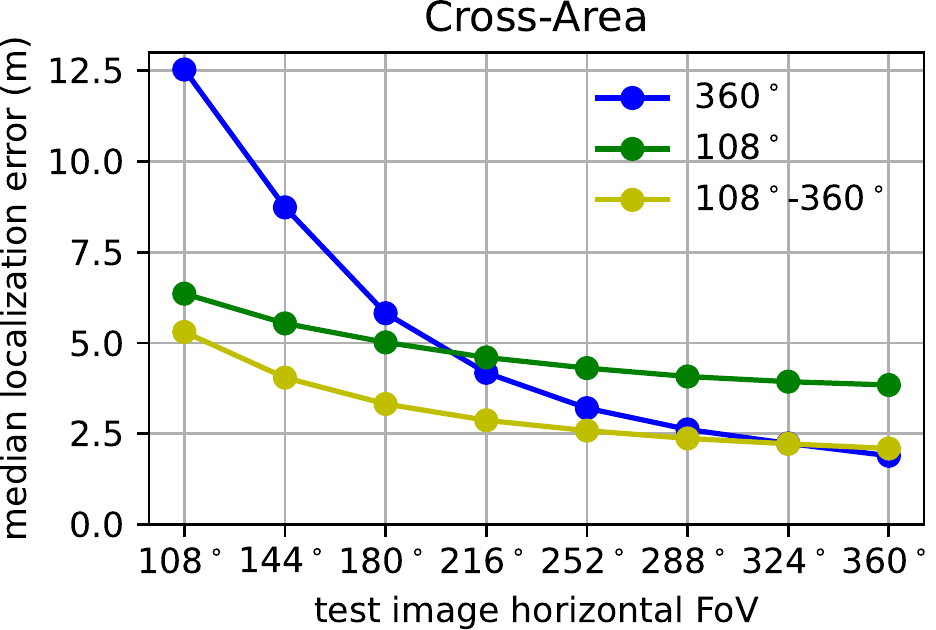} 
\caption{Localization on images with varying horizontal FoVs on VIGOR same-area and cross-area test sets. Green/blue/yellow curves represent our model trained with horizontal FoV of $108^\circ$/$360^\circ$/between $108^\circ$ and $360^\circ$.
} 
\label{fig:error_FoV_same_median_VIGOR}
\end{figure}

\textbf{Inference on images with different FoVs:}
Moreover, our model can infer on test ground images with other horizontal FoVs than in the training data.
In Fig.~\ref{fig:error_FoV_same_median_VIGOR}, we show the results of our models trained with ground images with various horizontal FoVs being tested on ground images with different horizontal FoVs.
In general, when the FoV of the ground image increases, the information contained in the image also increases.
As a result, we see a monotonic decrease in localization error when the test FoV increases for all models in Fig.~\ref{fig:error_FoV_same_median_VIGOR}.

\begin{figure*}[t] 
\centering
 \includegraphics[width=0.24\textwidth]{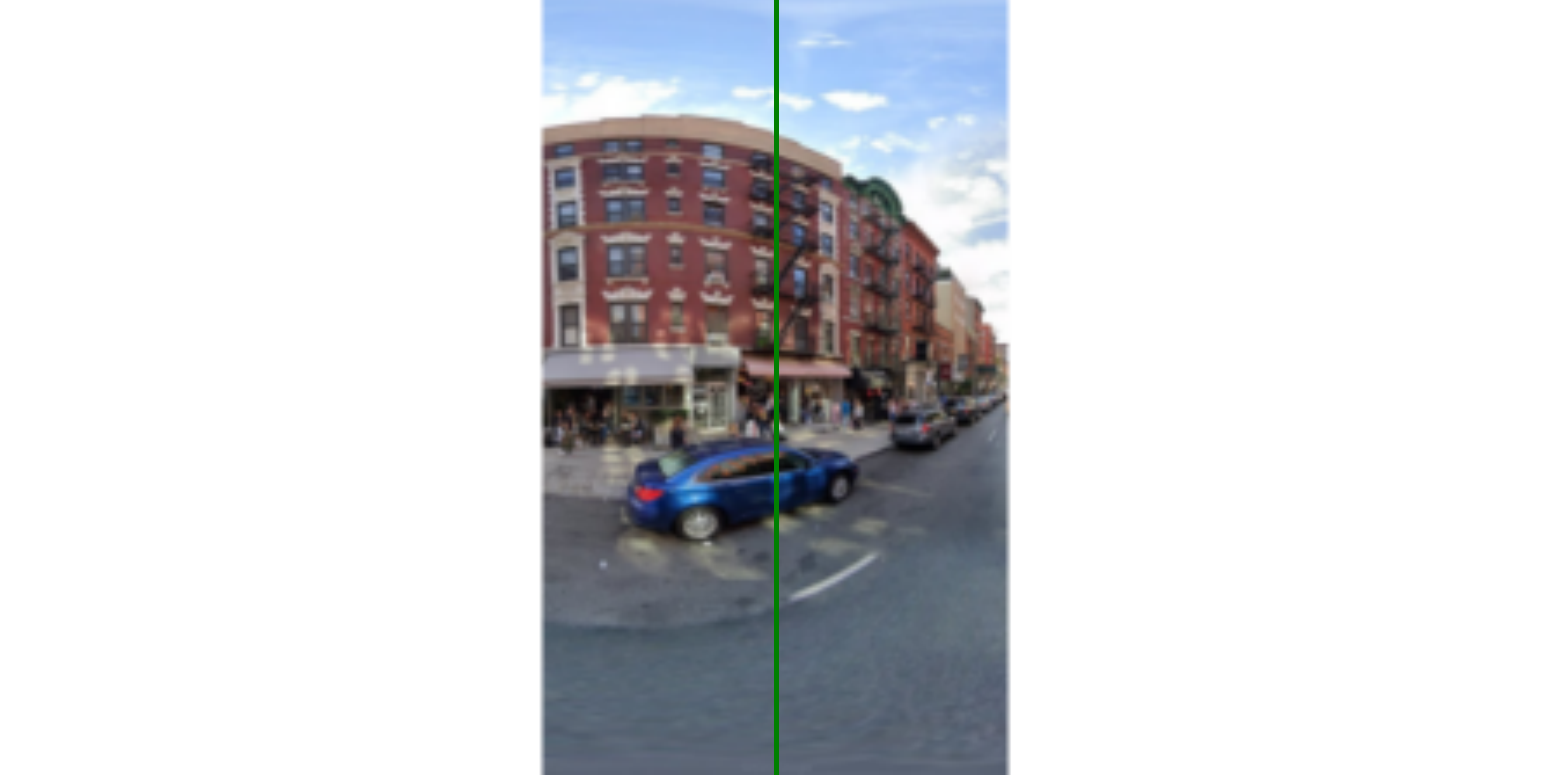} 
 \includegraphics[width=0.24\textwidth]{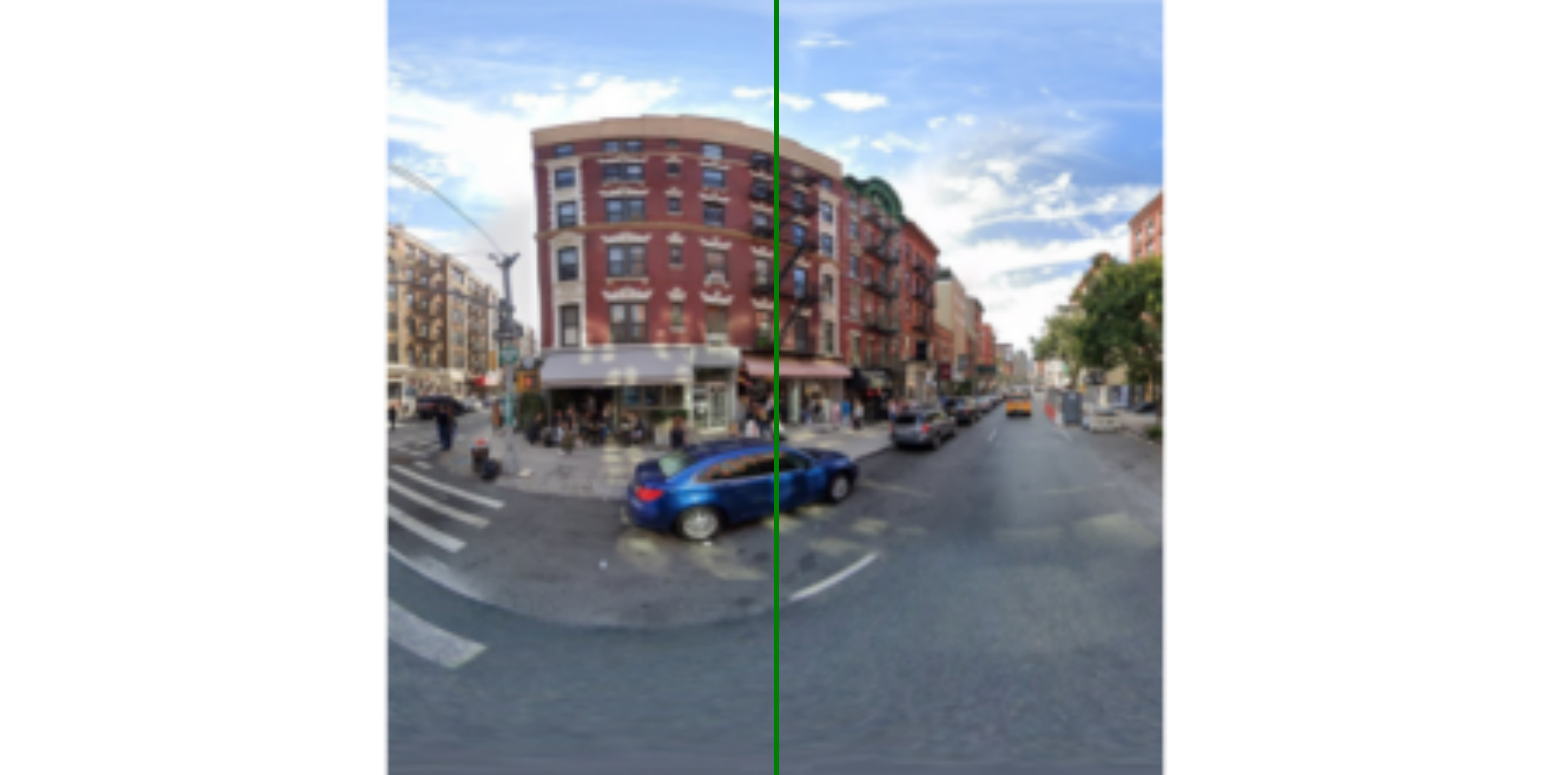} 
 \includegraphics[width=0.24\textwidth]{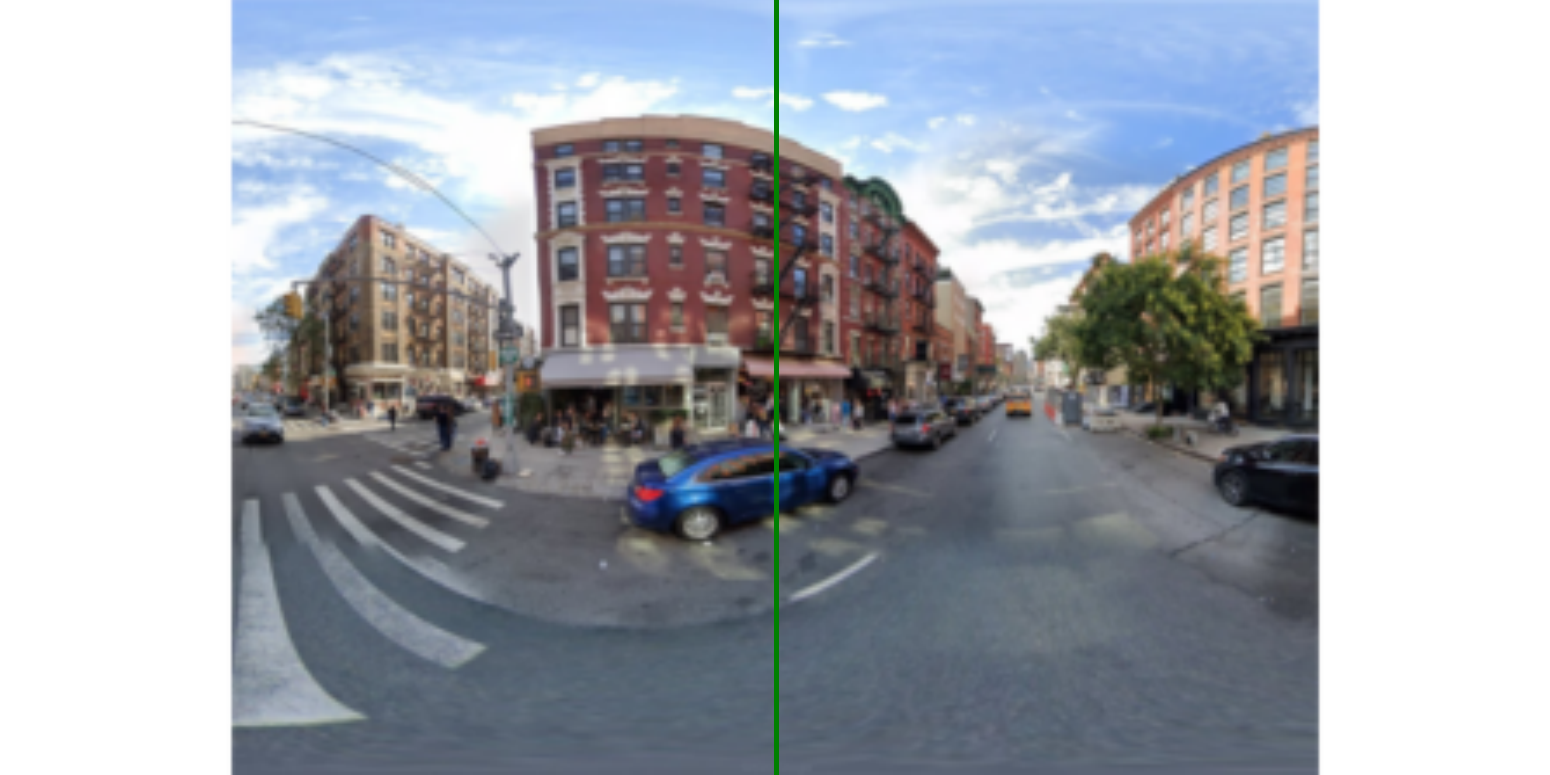} 
 \includegraphics[width=0.24\textwidth]{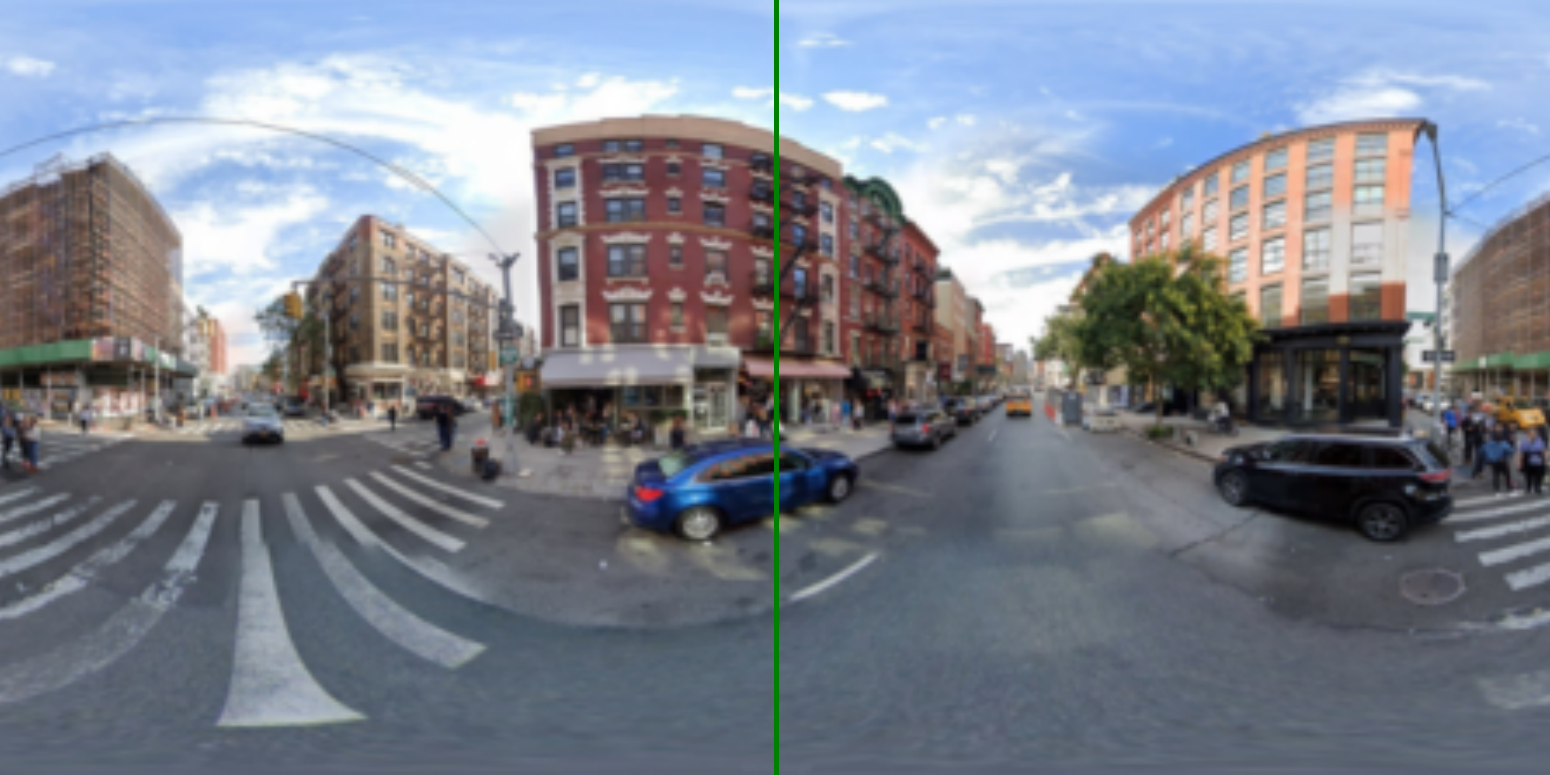}
 \vspace{1mm}
 
 \includegraphics[width=0.24\textwidth]{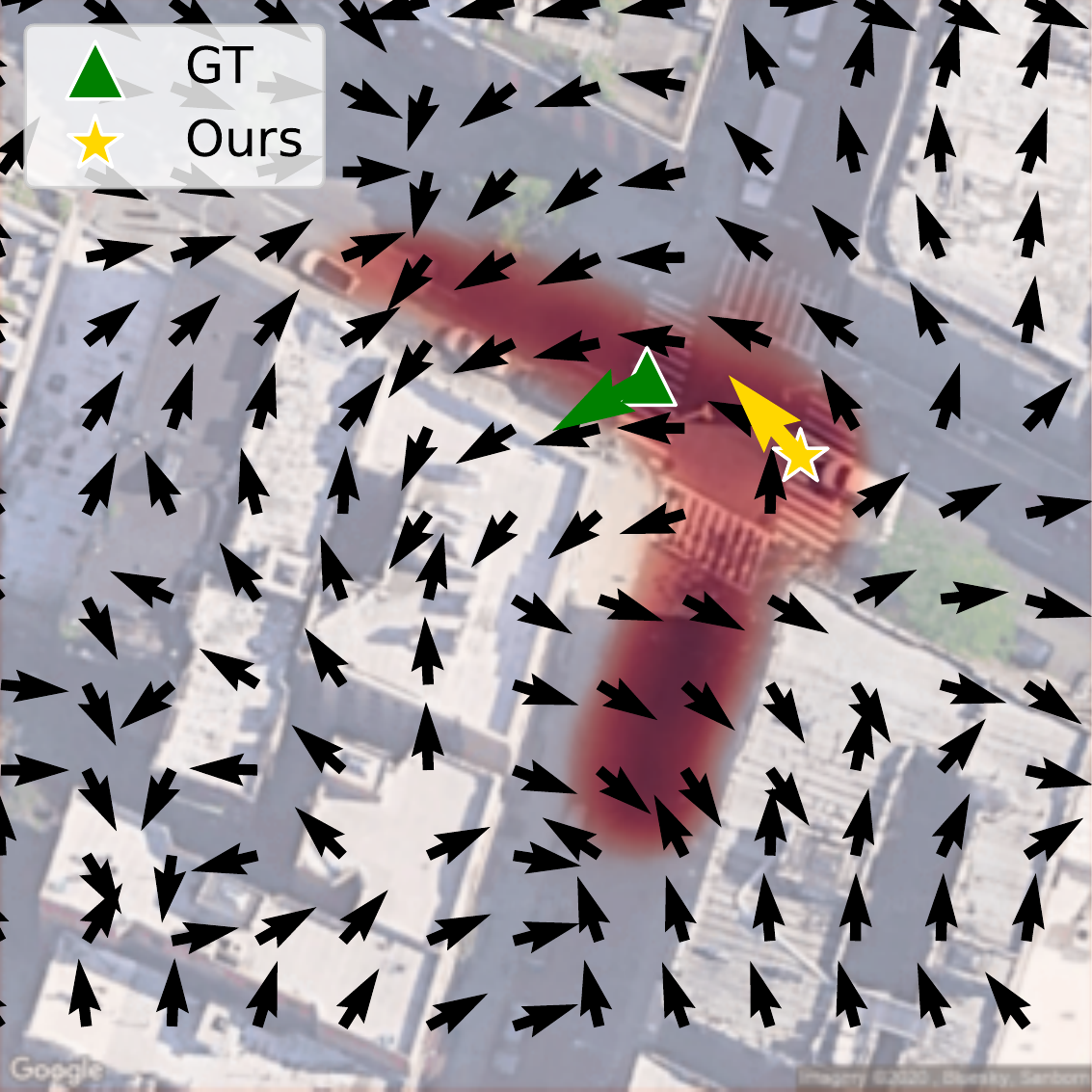}
 \includegraphics[width=0.24\textwidth]{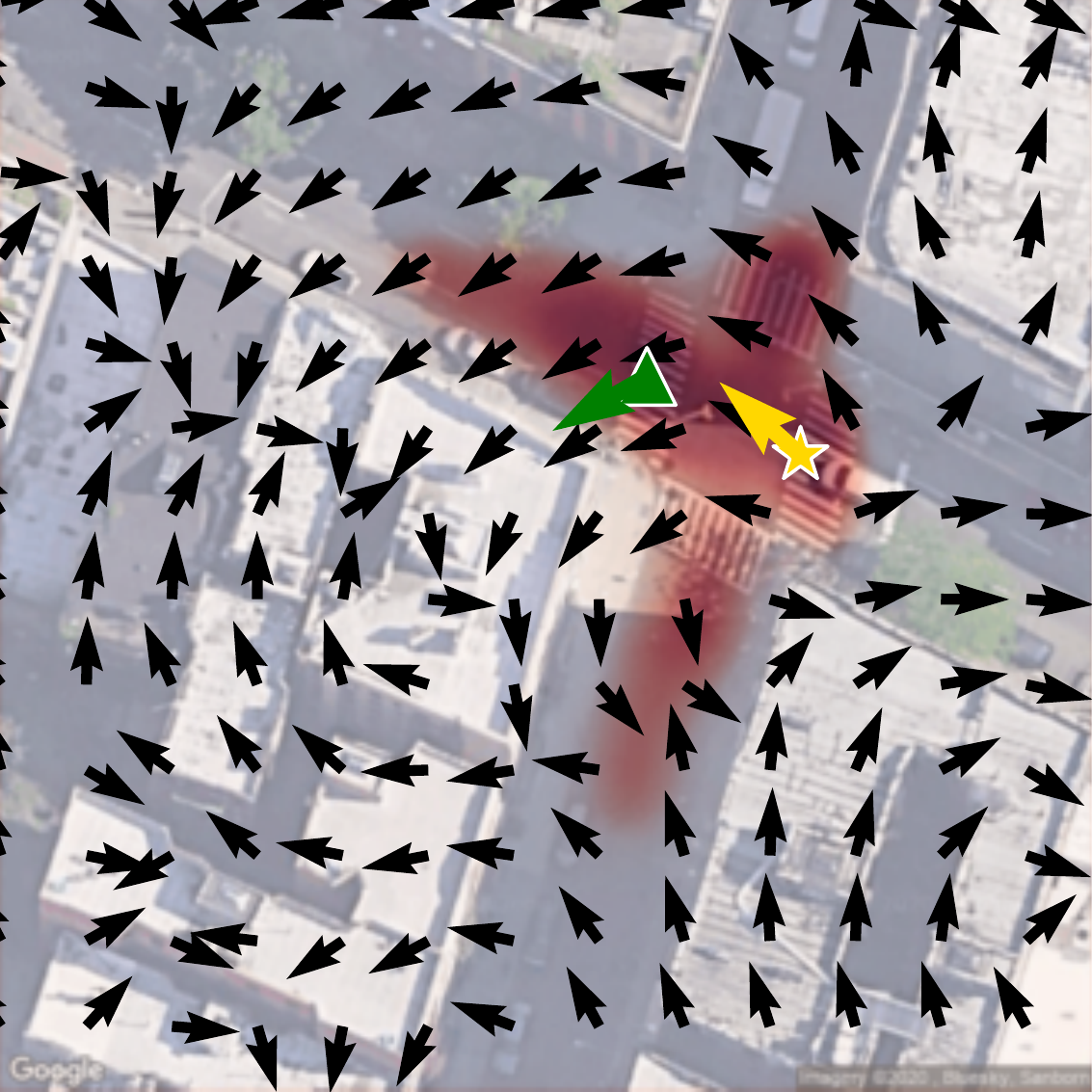}
 \includegraphics[width=0.24\textwidth]{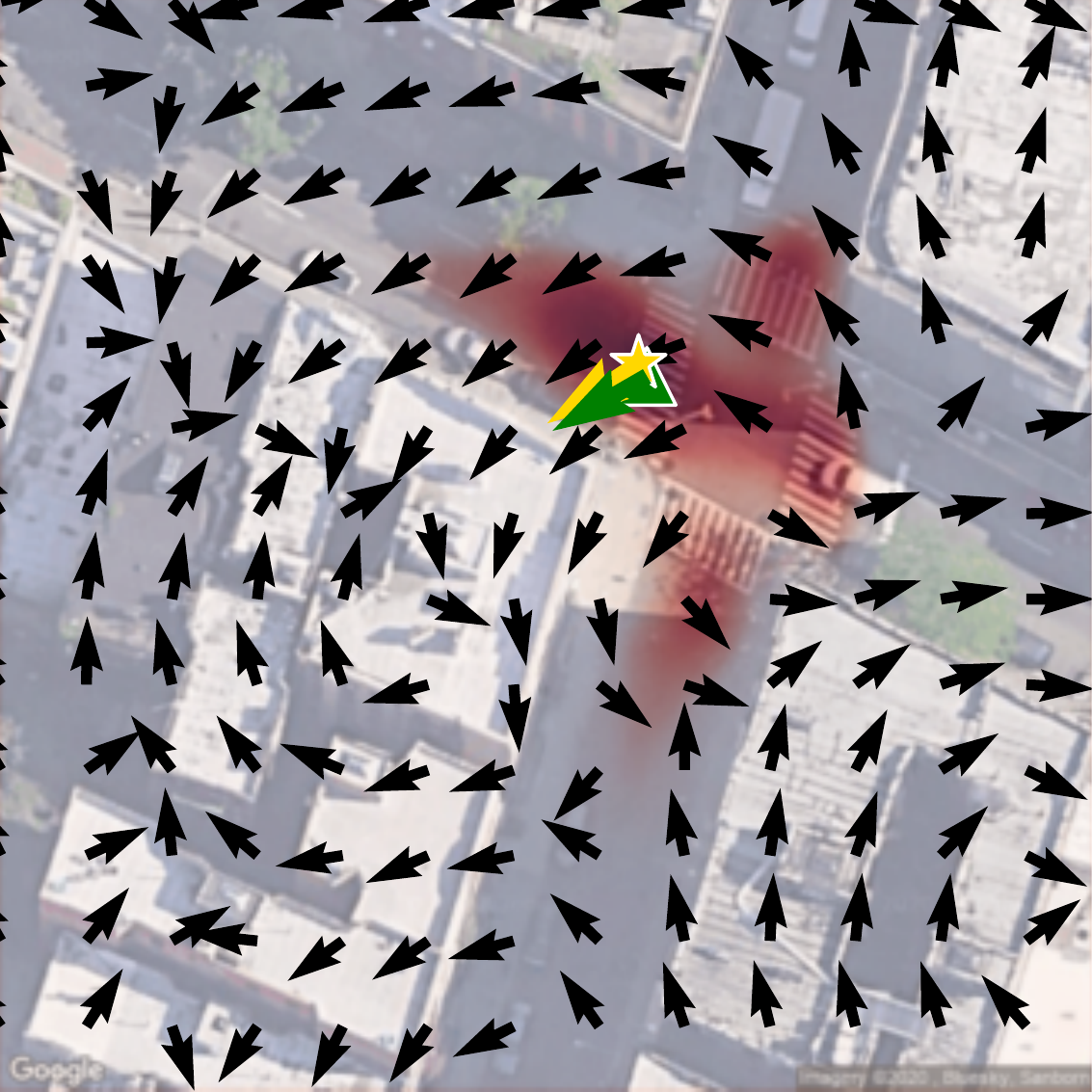}
 \includegraphics[width=0.24\textwidth]{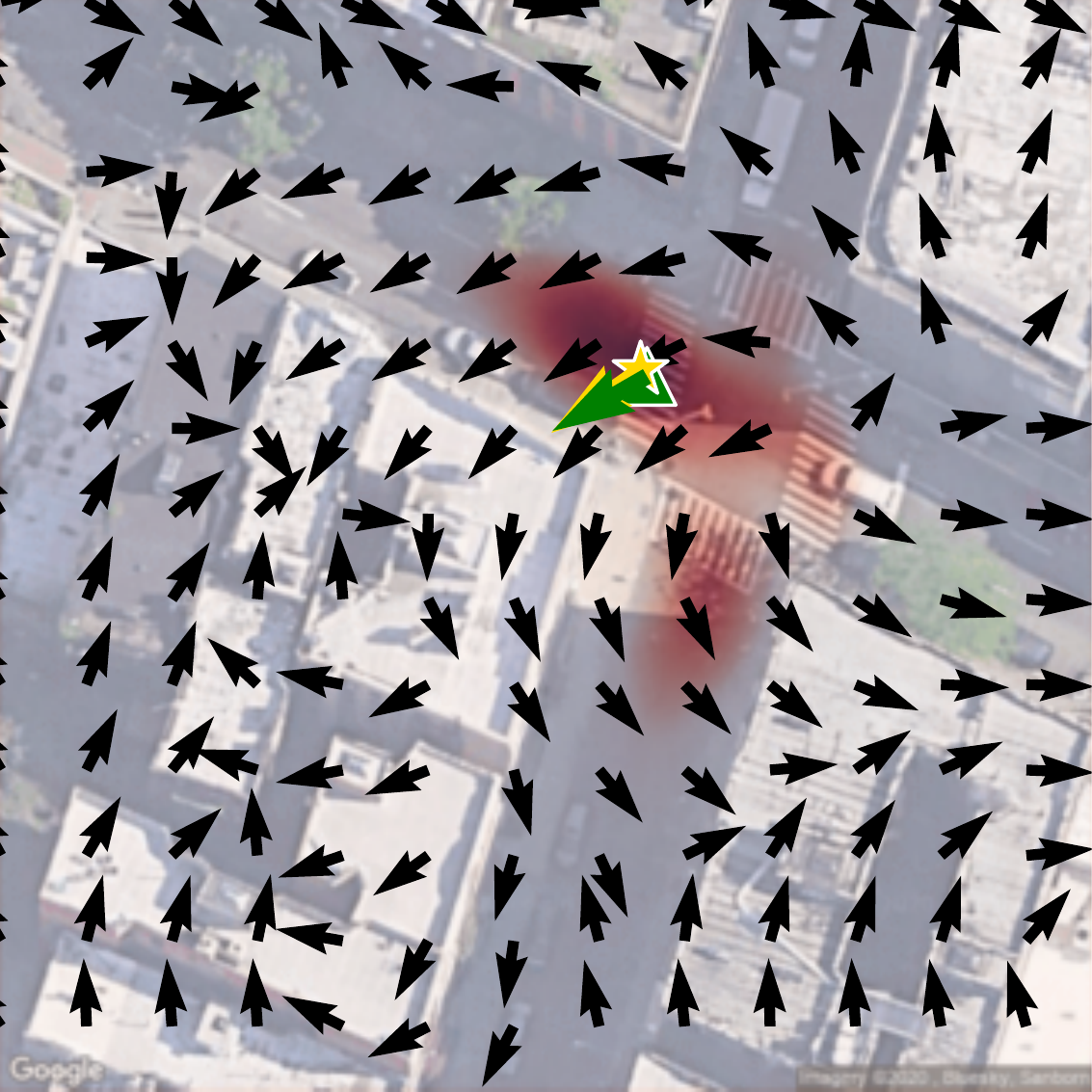}
\caption{Our model trained with panoramic image and inference on ground images with different FoVs (from left to right: $108^\circ, 180^\circ, 252^\circ, 360^\circ$) from VIGOR samearea test set.
Our dense orientation prediction is shown by black arrows. The final predicted orientation of the ground image (green vertical line) is shown by the yellow arrow.
} 
\label{fig:qualitative_results_VIGOR_different_FoV}
\end{figure*}

An example of the predictions from our model trained with panoramic images is shown in Fig.~\ref{fig:qualitative_results_VIGOR_different_FoV}.
When the FoV of the test ground image is $108^\circ$ or $180^\circ$, our model cannot distinguish different roads based on the limited content captured by the ground image, and thus predicts a multi-modal distribution to capture the probable locations.
However, the peak of the distribution is in the wrong mode and consequently, the selected orientation is also wrong.
When the test FoV increases to $252^\circ$, the peak of the output distribution is close to the ground truth location.
Further increasing the FoV reduces the localization uncertainty and improves the localization.
Notably, our model can always access the full scene layout information from the aerial view no matter what the FoV of the ground view is.
This example shows that the learned prior from the BEV layout solely is not enough for pose estimation, and our ground-aerial descriptor matching is crucial.

Because of the domain shift, the model trained with panoramas performs worse on images with small FoVs, compared to the model trained with images with a small FoV, see Fig.~\ref{fig:error_FoV_same_median_VIGOR}.
Besides, we also see a steeper decrease in localization performance when the test FoV reduces for the model trained with panoramas than the model trained with images of a horizontal FoV of $108^\circ$.
Training with images with a large FoV allows the model to use features \newtext{that} span widely in the ground image.
When those features are absent, e.g. testing with a small FoV, the performance degenerates.
On the other hand, if the training images only have a small FoV, the model would not learn to use features that span wider than the FoV of the training images.
Consequently, increasing the test FoV brings less benefit.
To tackle this trade-off, one can train the model with images whose FoVs are randomly sampled by cropping the panorama, e.g. sampled from $108^\circ, \cdots, 360^\circ$.
Consequently, the resulting model performs well for all tested FoVs.
Interestingly, this model also has slightly better localization performance than the model trained with images with FoV of $108^\circ$ when inference on images with FoV of $108^\circ$.
Note that this model is not used in our earlier comparison to other baselines for fairness since the baselines cannot include a similar data augmentation.

\subsection{Ego-vehicle pose estimation across time}
\begin{figure}[t] 
\centering
 \includegraphics[width=0.4\textwidth]{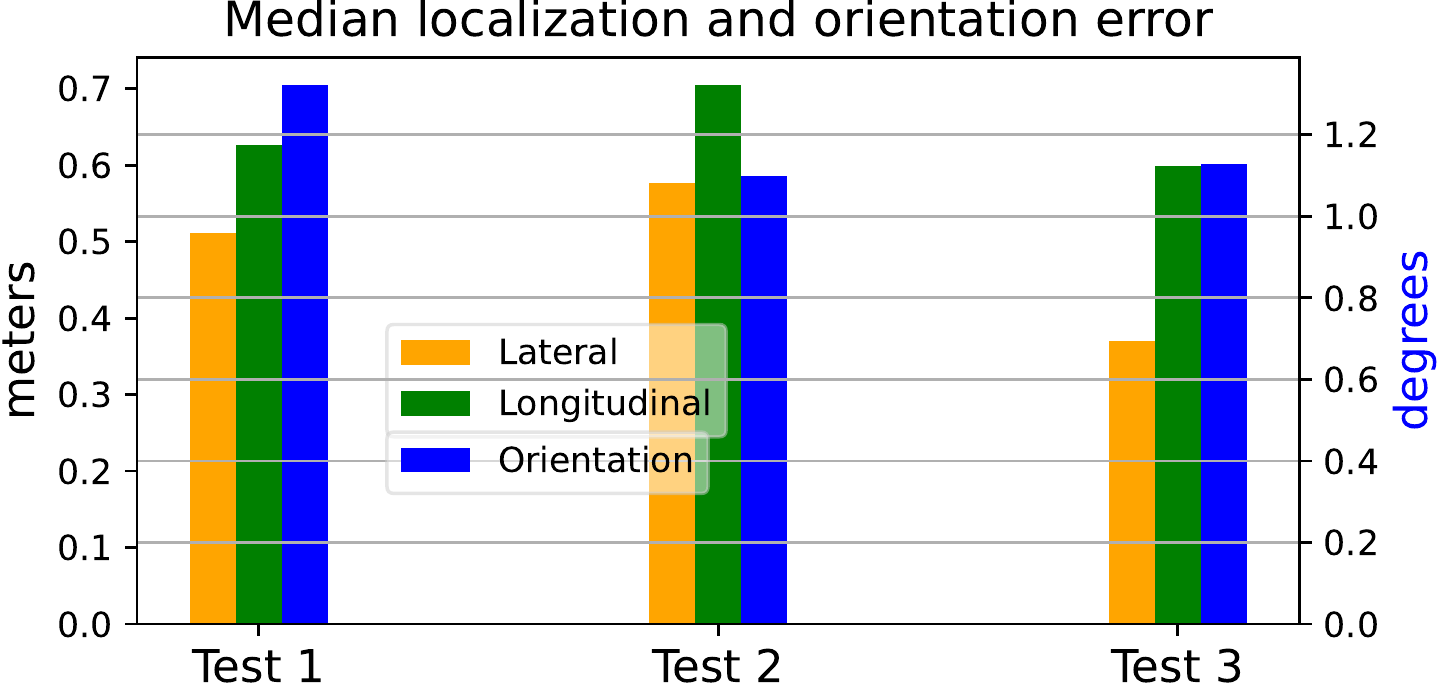} 
\caption{Median lateral and longitudinal localization error and median orientation error on Oxford RobotCar Test 1, 2, and 3 traversals.
} 
\label{fig:Oxford_median_error}
\end{figure}

On the Oxford RobotCar dataset, we deploy our model to follow the ego-vehicle over a sequence of ground-level images taken by the vehicle-mounted camera.
To process a pair of input ground and aerial images on Oxford RobotCar, CCVPE takes 0.07 seconds, i.e. $14$ FPS.
We assume there is a rough GNSS prior that identifies which aerial patch contains the location of the ground-level image.
As shown in Fig.~\ref{fig:Oxford_median_error}, on all three test traversals, our model achieves median lateral and longitudinal  localization error below $1$ meter and median orientation error around $1^\circ$.
Notably, even though the ground images in the Oxford RobotCar dataset have a small horizontal FoV compared to the panoramas in the VIGOR dataset, our model generalizes better to new ground images along the same route across time on the former dataset than to new panoramas on the latter dataset.
On the Oxford RobotCar dataset, the aerial view can provide a strong prior as the vehicle always drives along the same route and the test area is seen during training.
In contrast, the panoramas in the VIGOR dataset are not always captured on the road, plus its scenes are more diverse because of its broad coverage.

\begin{figure}[t] 
\centering
 \includegraphics[width=0.24\textwidth]{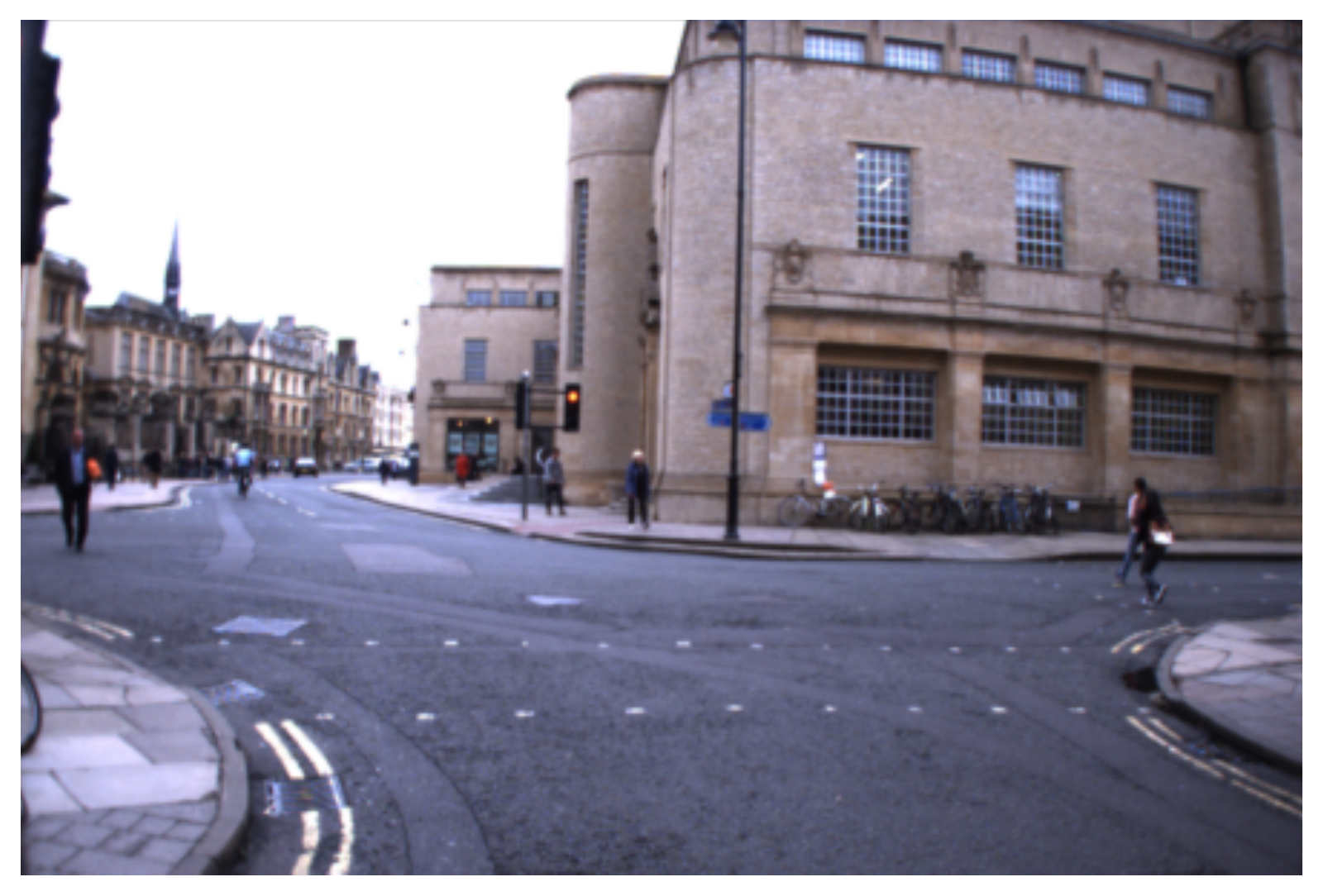} 
 \includegraphics[width=0.24\textwidth]{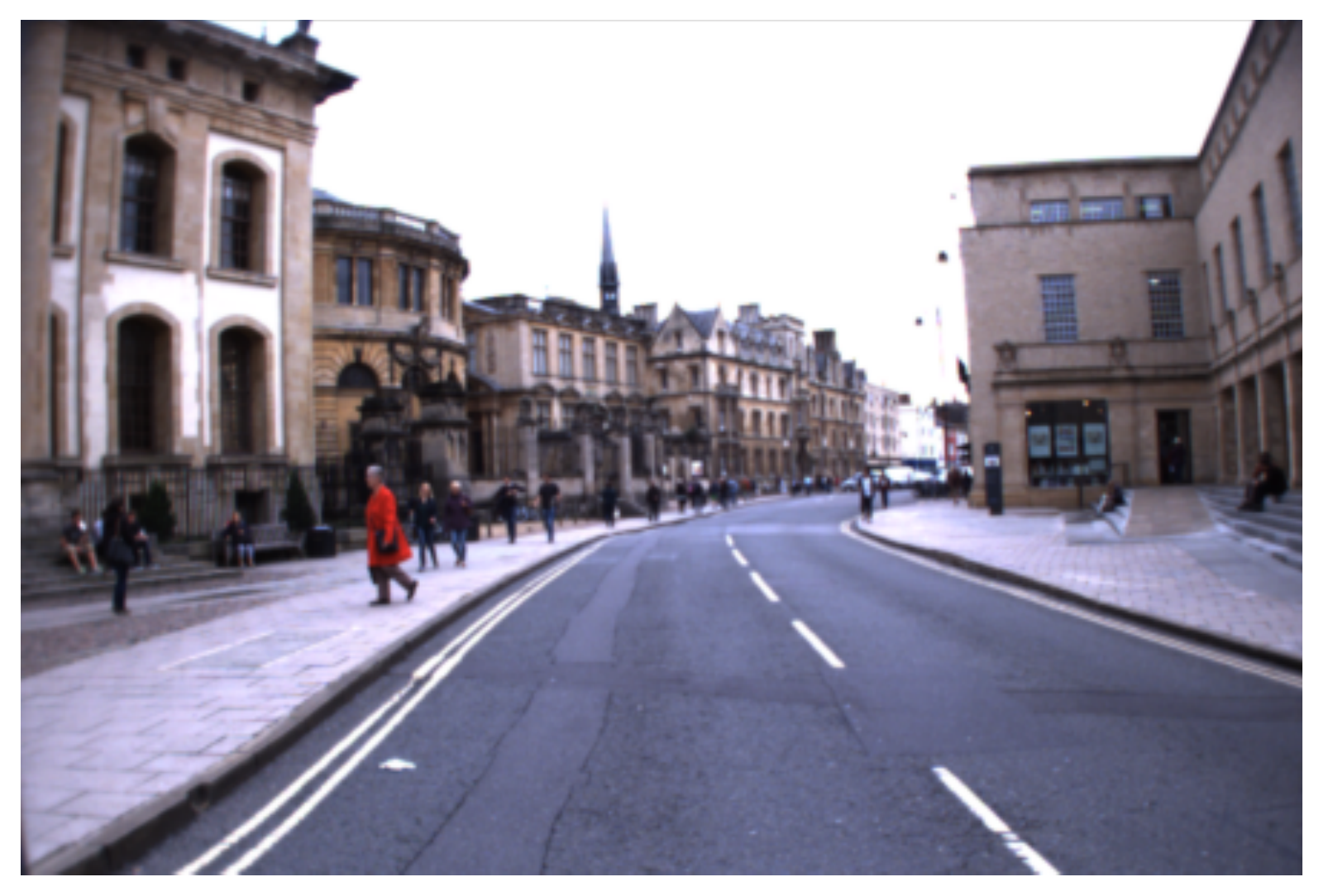} \\
 \vspace{0.1mm}
 \begin{annotatedFigure}
	{\includegraphics[width=0.24\textwidth]{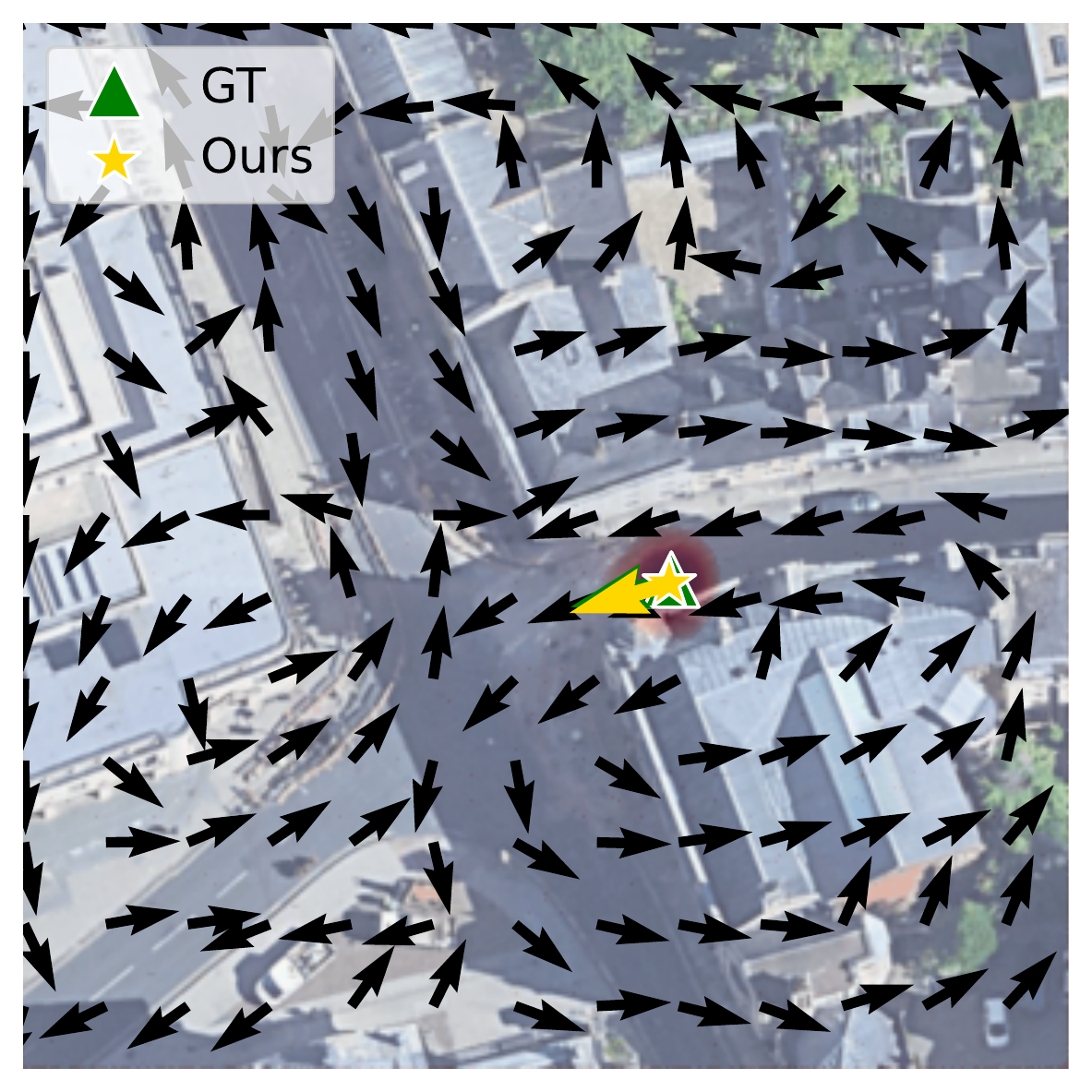}}
	\annotatedFigureBox{0.466,0.402}{0.725,0.596}{A}{0.466,0.596}
	\annotatedFigureBox{0.201,0.154}{0.445,0.41}{B}{0.201,0.154}
 \end{annotatedFigure}
 \begin{annotatedFigure}
	{\includegraphics[width=0.24\textwidth]{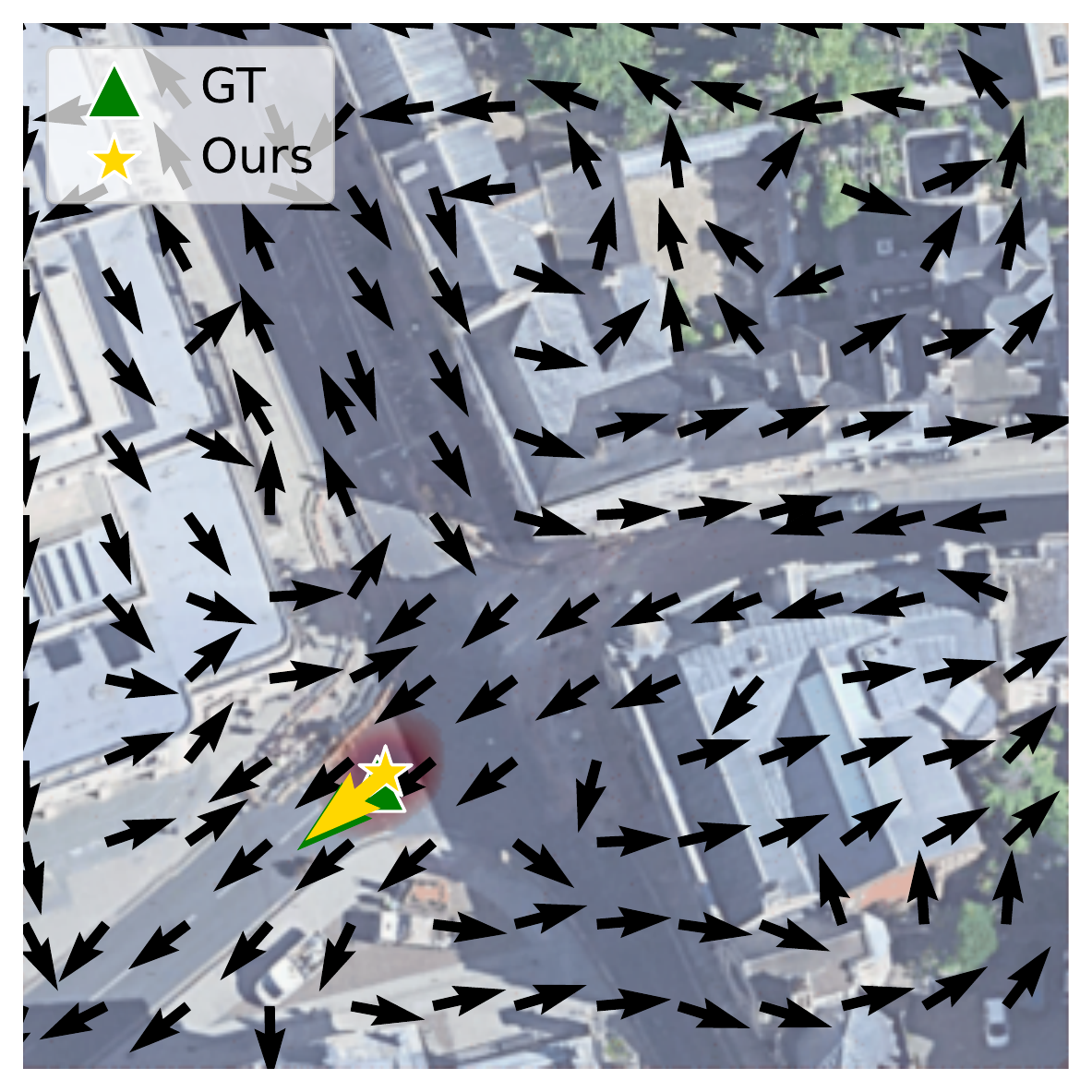}}
	\annotatedFigureBox{0.466,0.402}{0.725,0.596}{A}{0.466,0.596}
	\annotatedFigureBox{0.201,0.154}{0.445,0.41}{B}{0.201,0.154}
\end{annotatedFigure}

\caption{Localization and orientation estimation on two frames in a sequence on Oxford RobotCar Test 1 traversal. Right frame: ${\sim}13$ seconds after the left frame. Because the two ground-level images capture a different scene, the output orientation field in the same region A (or B) in the identical aerial images is different.
}
\label{fig:Oxford_qualitative}
\end{figure}

On the Oxford RobotCar sequences, we can observe how the predicted orientation map adapts to ground images at different nearby locations within the same aerial view,
as seen in Fig.~\ref{fig:Oxford_qualitative}.
While the model learned a prior from the aerial view on the driving direction of the roads,
the orientation predictions do respond to the ground image content at the high-probability locations.
E.g.~ when the vehicle is in area A, the local orientation field points towards the junction seen in the ground view. The orientations in area A reflect a different orientation (a prior) once the vehicle moved on to area B.
We provide a Supplementary Video of ego-vehicle pose estimation across time on this dataset.

\subsection{Ablation study}
\oldtext{Finally}
\newtext{Next}, we present an ablation study on the VIGOR  Same-Area validation set.

\textbf{Number of LMU modules:}
As mentioned in Sec.~\ref{sec:implementation_details}, our model has $K=6$ LMU modules for coarse-to-fine descriptor matching for localization.
Here, we study the effect of LMU modules on localization performance by removing them at low or high levels.
When removing an LMU module, we modify the corresponding convolutional layer in the Localization Decoder such that it directly processes the aerial feature without any matching scores.

\begin{table*}[ht]
\centering
\caption{Effect of LMU modules on mean localization error on VIGOR same-area validation set. \textbf{Best in bold.}}
\label{table:ablation_GGFP}
\begin{tabular}{|p{2.2cm}|p{0.6cm}|p{0.6cm}|p{0.6cm}|p{1cm}|p{1.2cm}|p{0.7cm}|p{0.6cm}|p{0.6cm}|p{1cm}|p{1.2cm}|p{1.2cm}|} 
\hline
\multicolumn{12}{|c|}{VIGOR Same-Area, validation set}  \\
\hline
K =  & 1 & 1,2 & 1,2,3 & 1,2,3,4 & 1,2,3,4,5 & 6 & 5,6 & 4,5,6 & 3,4,5,6 & 2,3,4,5,6 & 1,2,3,4,5,6 \\
\hline
median error (m) & 2.58 & 1.89 & 1.58 & 1.45 & 1.44 & 2.68 & 1.99 & 1.58 & 1.47 & \textbf{1.42} & \textbf{1.42} \\
\hline
\end{tabular}
\end{table*}

\begin{figure*}[t] 
\centering
\begin{subfigure}[t]{0.3\textwidth}
     \centering
     \includegraphics[width=1\textwidth]{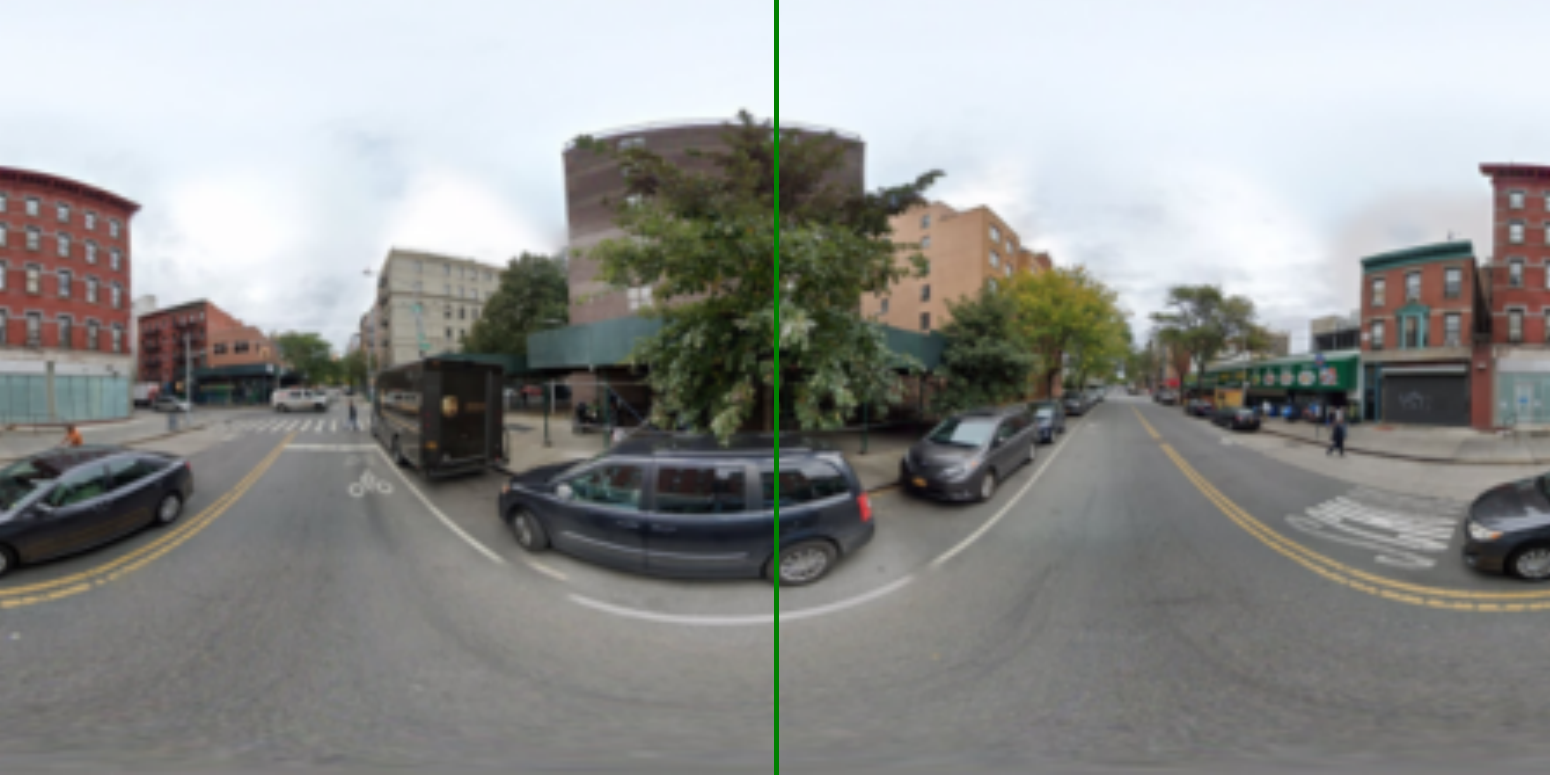}
     \caption{ground image}
\end{subfigure}
\begin{subfigure}[t]{0.195\textwidth}
    \centering
    \includegraphics[width=1\textwidth]{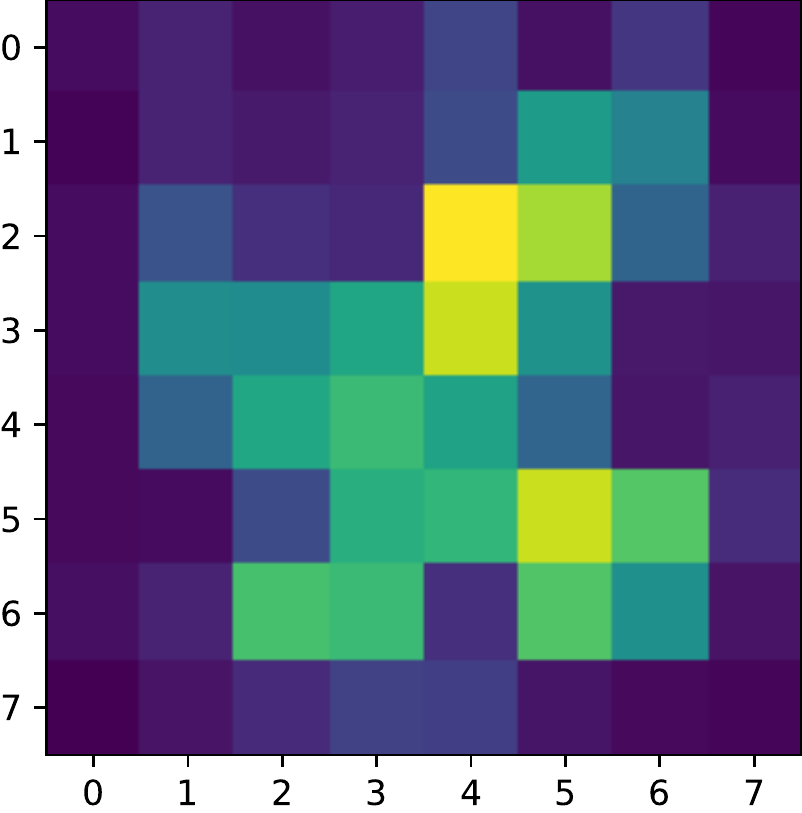}
    \caption{$k=1$}
\end{subfigure}
\begin{subfigure}[t]{0.2\textwidth}
    \centering
    \includegraphics[width=1\textwidth]{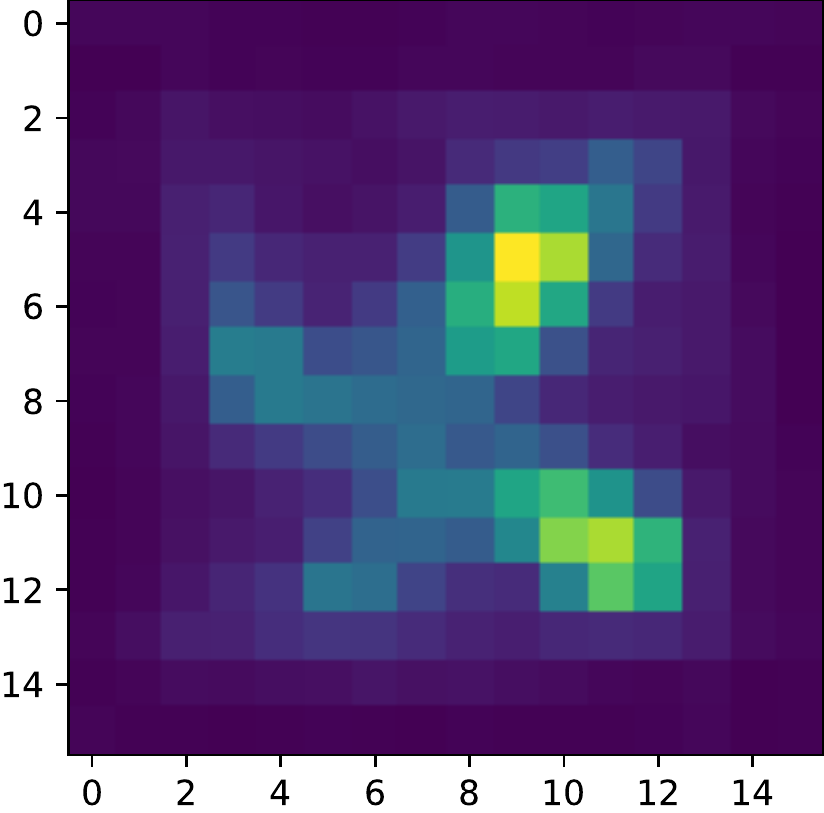}
    \caption{$k=2$}
\end{subfigure}
\begin{subfigure}[t]{0.2\textwidth}
    \centering
    \includegraphics[width=1\textwidth]{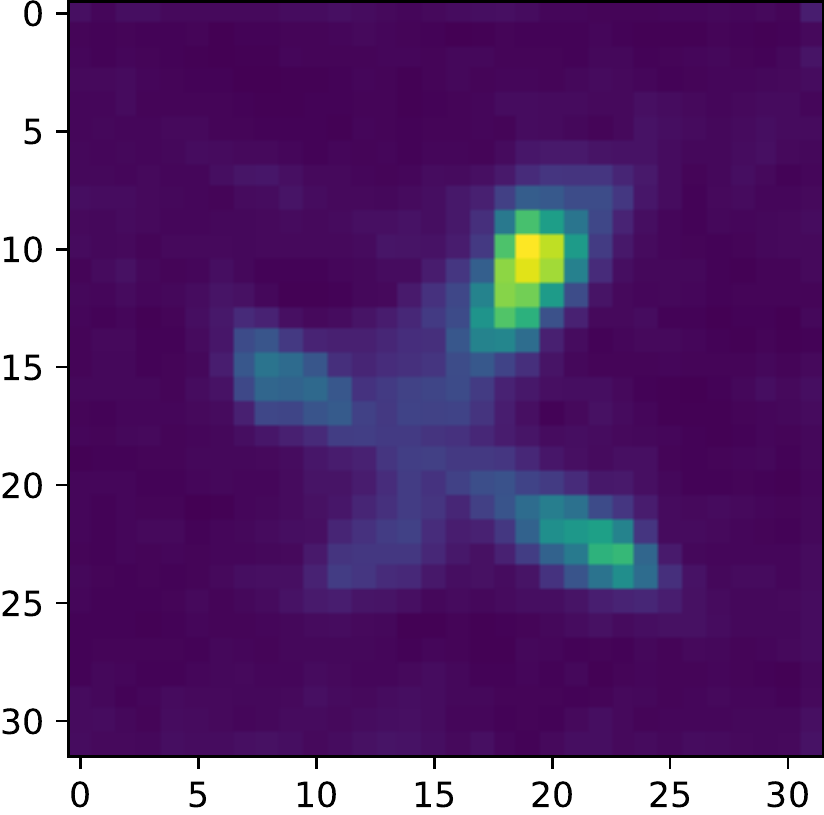}
    \caption{$k=3$}
\end{subfigure}
\begin{subfigure}[t]{0.3\textwidth}
     \centering
     \includegraphics[width=0.7\textwidth]{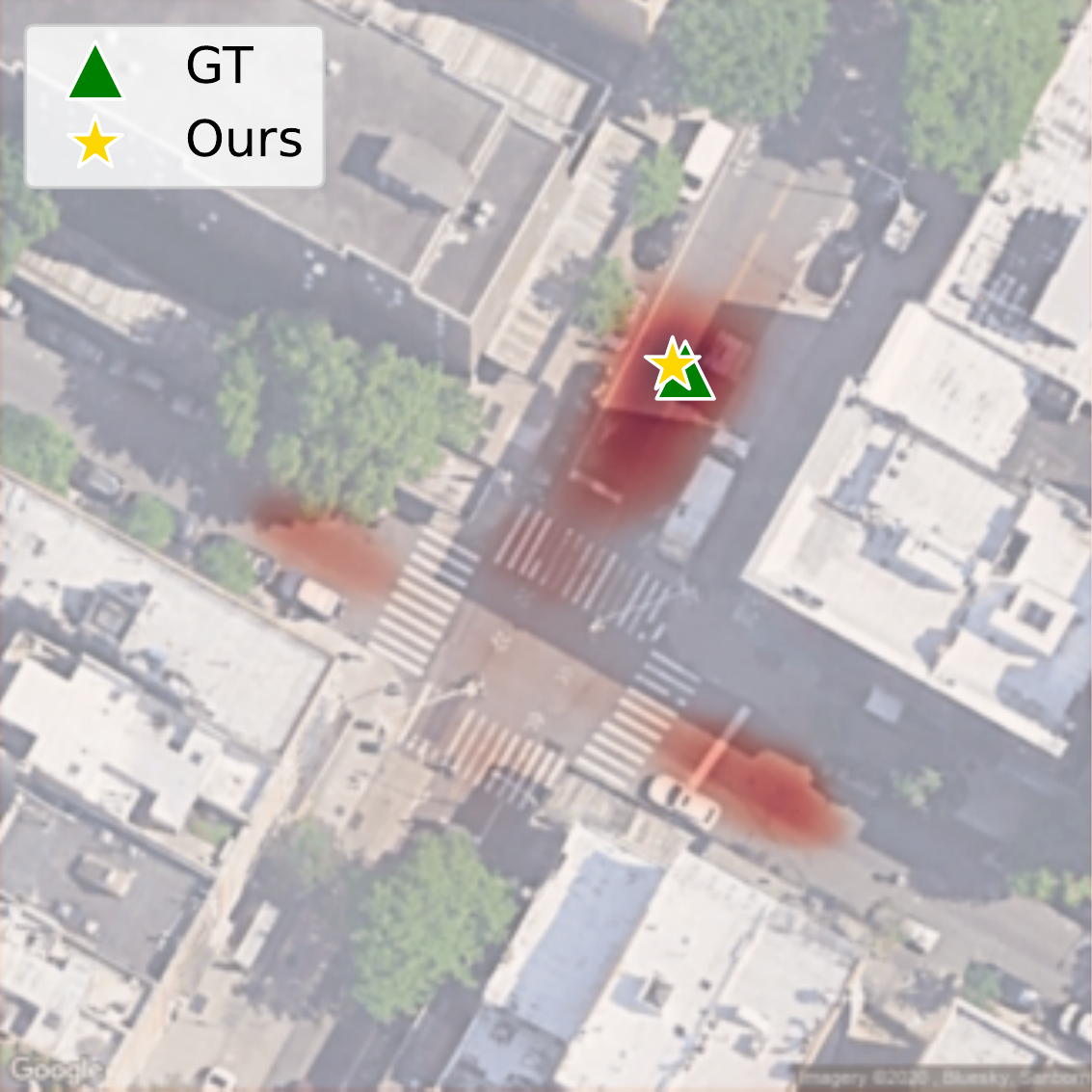}
     \caption{localization on aerial image}
\end{subfigure}
\begin{subfigure}[t]{0.195\textwidth}
    \centering
    \includegraphics[width=1\textwidth]{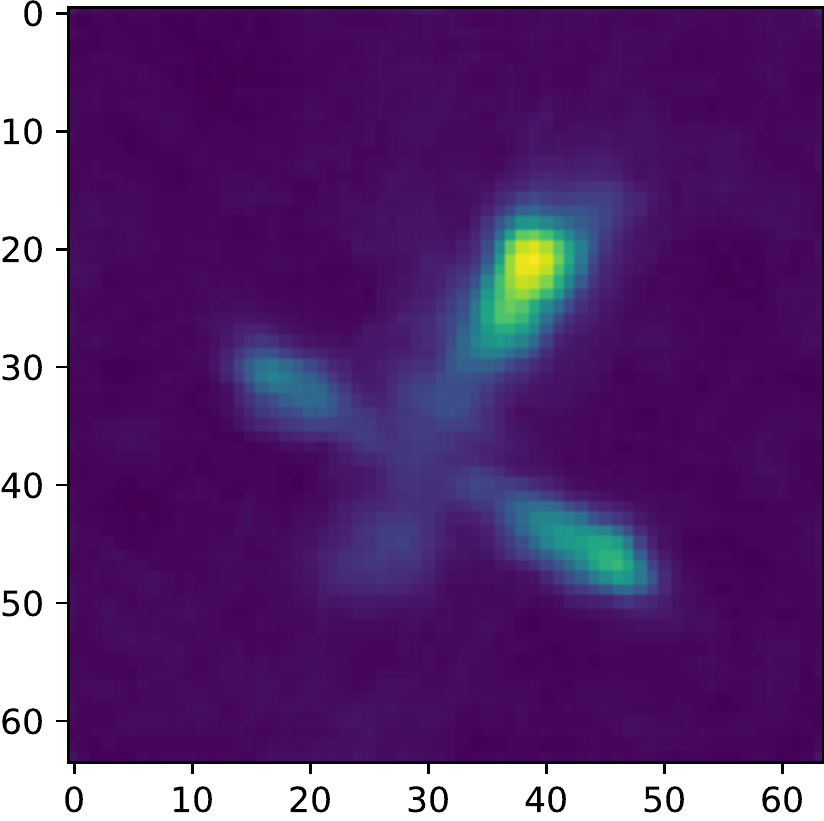}
    \caption{$k=4$}
\end{subfigure}
\begin{subfigure}[t]{0.2\textwidth}
    \centering
    \includegraphics[width=1\textwidth]{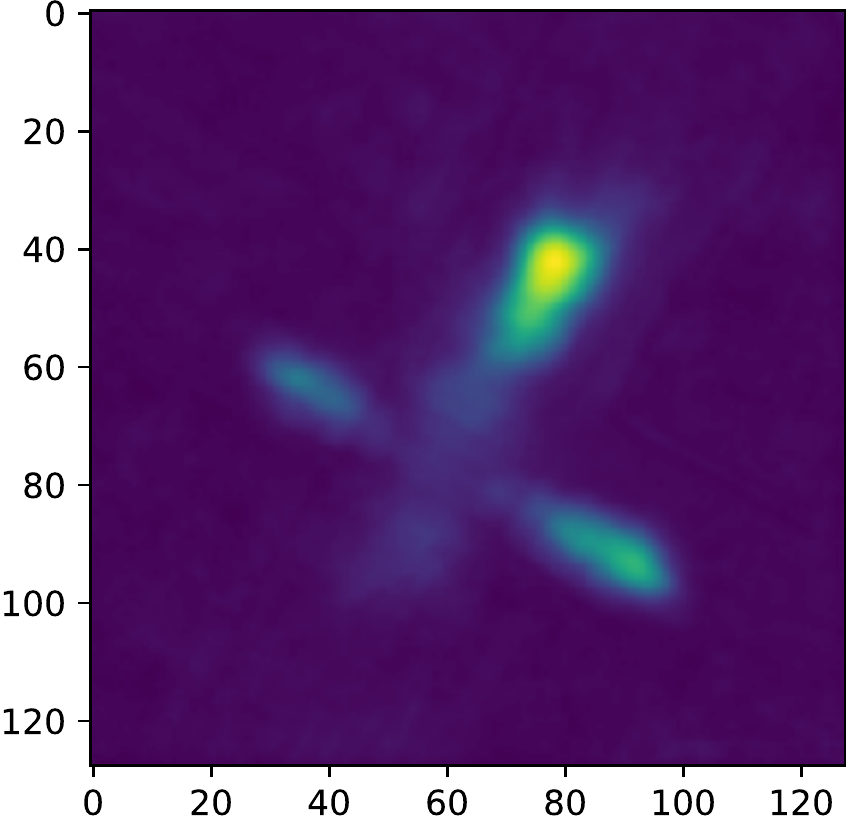}
    \caption{$k=5$}
\end{subfigure}
\begin{subfigure}[t]{0.2\textwidth}
    \centering
    \includegraphics[width=1\textwidth]{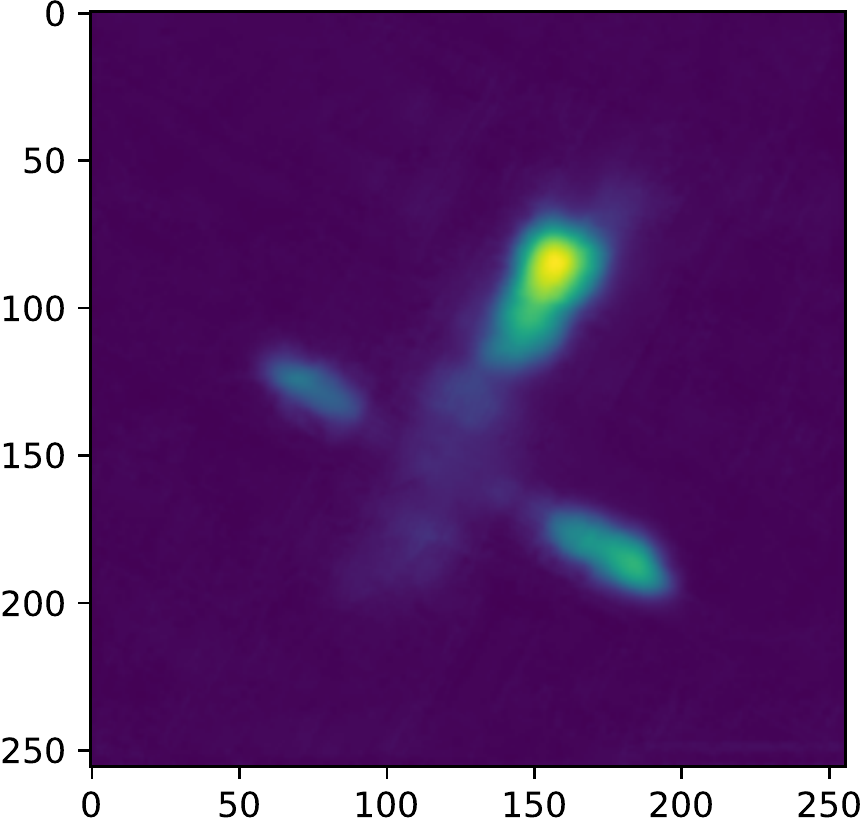}
    \caption{$k=6$}
\end{subfigure}
\caption{Visualization of the max matching score map in LMU at levels $1$ to $6$.}
\label{fig:effects_of_GGFPs}
\end{figure*}

As shown in Tab.~\ref{table:ablation_GGFP}, the model with LMU modules at all $6$ levels outperforms other variants in terms of the median localization error.
When excluding the LMU modules at low or high levels, we see a consistent decrease in localization performance.
Importantly, using LMU modules only at high levels, e.g. $K=6$, does not provide equally good localization performance as the models that also have LMU modules at lower levels.
Directly contrasting aerial descriptors at a fine resolution is a difficult learning task.
Using LMU modules at lower levels can provide a better starting point for descriptor matching at higher levels, leading to better localization performance.

Qualitatively, we see in Fig.~\ref{fig:effects_of_GGFPs} that the max matching score map inside the LMU module becomes sharper when the level $k$ increases, but does not improve noticeably anymore after level $4$, i.e. resolution $64 \times 64$, on the selected example.
Quantitatively, the increase in localization performance is also less when $k$ increases.
For our main experiments, we include LMU modules at all levels, i.e. $K=6$.
This setting also aligns with the commonly used coarse-to-fine formulation in other computer vision tasks~\cite{mayer2016large,dosovitskiy2015flownet}.

\textbf{Other architectural variations: }
\oldtext{Next,}
\newtext{We first compare our model to our conference work~\cite{xia2022visual}}.
\newtext{Then,} we study the effect of \newtext{the backbone,} the OMU module, the number of orientation channels $R$, and the L2-normalization before feature concatenation in the LMU and OMU modules.
\newtext{Finally, we test replacing the Rolling \& Matching with a simple concatenation of ground and aerial features, as well as removing the contrastive learning loss $\mathcal{L}_\mathcal{M}$}.

\newtext{In our conference work~\cite{xia2022visual}, VGG~\cite{VGG} and SAFA modules~\cite{SAFA} are used as feature extractors and feature projectors, and the ground-to-aerial matching is only conducted at the bottleneck.
}
\oldtext{We compare the pose estimation performance of our model to our conference work~\cite{xia2022visual}, where VGG~\cite{VGG} and SAFA modules~\cite{SAFA} are used as feature extractors and feature projectors.
Note that, in~\cite{xia2022visual}, the ground-to-aerial matching is only conducted at the bottleneck.}
Thus, for a fair comparison, we include a \newtext{CCVPE} model with VGG~\cite{VGG} backbone and use the LMU module only at the bottleneck, i.e. \oldtext{$K=1$}\newtext{`$K=1$, VGG'}.
In \cite{xia2022visual}, pose estimation is achieved by comparing the model's probability estimations on differently rotated samples, while we construct orientation-aware descriptors and estimate the pose in a single forward pass.
Validation results of models with different settings are summarized in Tab.~\ref{table:ablation_architecture}.
Our model with VGG~\cite{VGG} backbone and $K=1$ beats the \oldtext{method}\newtext{approach} in~\cite{xia2022visual} in both localization and orientation \oldtext{estimation}\newtext{accuracy}.

\oldtext{Comparing}
\newtext{The comparison between} the row `$K=1$, VGG' \oldtext{to} \newtext{and} the row `$K=1$' in Tab.~\ref{table:ablation_architecture}, shows that using EfficientNet-B0~\cite{tan2019efficientnet} as the \newtext{default} backbone improves both localization and orientation estimation.
Interestingly, we find that if we do not normalize the aerial descriptor before the feature concatenation in the LMU and OMU modules, both localization and orientation estimation performance decrease significantly. 
In particular, the orientation estimation becomes no better than a random guess.
The magnitude of cosine similarity matching score $\mathcal{M}_k$ is between $-1$ and $1$. 
Normalizing the aerial descriptors makes the magnitude of their elements stay in a similar range as $\mathcal{M}_k$.
If we do not normalize the concatenated aerial descriptors, the model might not effectively use the information in matching score $\mathcal{M}_k$ and treat $M_k$ as noise.

Similar to concatenating aerial descriptors without normalization, excluding the OMU module \newtext{and only processing the aerial descriptors} also makes the orientation prediction fail.
\newtext{Since we randomly change the ground images' orientation during training, there is no useful prior on the orientation when only considering the aerial image.}
\newtext{Next, we study the effect of different numbers of orientation channels $R$.}
When increasing the granularity in the orientation space, i.e. using a larger $R$ when rolling aerial descriptors in the LMU and OMU modules, both localization and orientation estimation performance increases.
Constructing aerial descriptors for more global orientation intervals not only provides more fine-grained orientation matching scores but also improves the orientation-aware features for localization.
Limited by the width $W'$ of encoded ground feature $g_e(G)$, the maximum $R$ we can use is 20 on the VIGOR dataset.
\oldtext{Besides, we also observe a small increase in mean orientation error comparing $R=10$ to $R=20$.}
\newtext{We observe a small increase in mean orientation error when increasing $R$ from $10$ to $20$.}
Overall, $R=20$ provides the best localization result, and therefore we used it in our main experiments (we used $R=16$ on KITTI because the input image has a different resolution).
Similar to the Localization Decoder, we tested including 6 OMU modules for all 6 levels in the Orientation Decoder.
Although this setting reduces the median orientation error, we observe an increase in the localization error and mean orientation error.
LMU at higher levels has finer spatial resolutions, while the granularity in orientation space is fixed, e.g. $R=20$, in all OMU modules.
Thus, we do not expect the same benefit here as in the Localization Decoder, and we use the OMU module only at the first level.

\newtext{When replacing the proposed Rolling \& Matching by straightforward ground-aerial feature concatenation, there is a large drop in both localization and orientation estimation performance, see the comparison between `$K=1$' and `Concat@1' and the comparison between `$K=6, R=20$' and `Concat@6' in Tab.~\ref{table:ablation_architecture}.
When directly concatenating the ground feature with the aerial feature, the model has the additional challenge of learning that different rotated versions of the same panoramic image should be located at the same place.
In contrast, our Rolling \& Matching design injects inductive biases into the model by using the translational equivariant ground encoder, and by forcing corresponding ground and aerial descriptors to be similar.
Specifically, we ensure this rotational equivariance is kept by the OMU for orientation estimation.
Therefore, when inputting different rotated versions of the same panorama, the same matching score pattern would re-occur in different orientation channels.
Our Orientation Decoder still needs to learn how different permutations of matching scores translate to the orientation vector field, but this matching volume has a relatively low number of channels compared to concatenated ground and aerial features.
The LMU's inductive bias is to be \textit{invariant} to different ground camera's orientations, which is achieved by taking the maximum over orientation channels.
None of these orientation and localization-specific inductive biases are present in the concatenation approach, which explains the large difference in performance.
}

\newtext{
Importantly, our Rolling \& Matching is empowered by the orientation-aware ground and aerial descriptors. 
If we do not enforce the orientation awareness for the aerial descriptor by removing the infoNCE losses, both localization and orientation prediction performance of the model decreases significantly, see rows with `No infoNCE' in Tab.~\ref{table:ablation_architecture}.
}

\begin{table}[ht]
\centering
\caption{CCVPE architecture comparisons on VIGOR Same-Area validation set. \textbf{Best in bold.} `No norm' means we do not normalize the aerial descriptors before feature concatenation in the LMU and OMU modules. `6 OMUs' means we include the OMU module at all levels. \newtext{`Concat' denotes direct concatenating of the ground and aerial features instead of conducting Rolling \& Matching.}}
\label{table:ablation_architecture}
\begin{tabular}{|p{2.5cm}|p{1cm}|p{1cm}|p{1cm}|p{1cm}|} 
\hline
\multirow{2}{*}{VIGOR, val.} & \multicolumn{2}{c|}{$\downarrow$ Localization (m)} & \multicolumn{2}{c|}{$\downarrow$ Orientation ($^\circ$)}\\
\cline{2-5}
 & mean  & median & mean  &median \\
\hline
\cite{xia2022visual} & 9.76 & 6.15 & 55.91 & 15.66 \\
\hline
K=1, VGG & 7.06 & 3.90 & 18.12 & 7.91 \\
\hline
K=1, No norm & 5.44 & 2.95 & 89.72 & 89.17 \\
\hline
\newtext{K=1, No infoNCE} & \newtext{9.37} & \newtext{6.04} & \newtext{90.59} & \newtext{90.53} \\
\hline
K=1 & 4.93 & 2.61 & 14.68 & 7.50 \\
\hline
\newtext{Concat@1} & \newtext{13.57} & \newtext{11.87} & \newtext{89.29} & \newtext{89.17} \\
\hline
K=6, No OMU & 3.73 & \textbf{1.40} & 89.92 & 89.86 \\
\hline
\newtext{K=6, No infoNCE} & \newtext{9.11} & \newtext{5.67} & \newtext{90.22} & \newtext{90.17} \\
\hline
K=6, R=2 & 5.35 & 2.40 & 36.74 & 25.39 \\
\hline
K=6, R=4 & 4.51 & 1.80 & 16.06 & 8.57\\
\hline
K=6, R=5 & 4.21 & 1.74 & 15.11 & 8.42\\
\hline
K=6, R=10 & 3.85 & 1.51 & \textbf{13.06} & 6.88\\
\hline
K=6, R=20 \newtext{(default)} & \textbf{3.63} & 1.42 & 13.11 & 6.61\\
\hline
K=6, 6 OMUs & 3.78 & 1.43 & 15.34 & \textbf{3.81}\\
\hline
\newtext{Concat@6} & \newtext{9.66} & \newtext{6.63} & \newtext{89.60} & \newtext{89.03} \\
\hline
\end{tabular}
\end{table}

\textbf{Loss on localization heat map:}
Using our best model architecture, we compare the cross-entropy loss and Wasserstein distance-based loss for localization and uncertainty estimation.
In Tab.~\ref{table:ablation_loss}, we show the mean and median localization error and the predicted probability at the ground truth pixel of models trained with different losses.

\begin{table}[ht]
\centering
\caption{Evaluation of CCVPE with different localization losses on VIGOR Same-Area validation set. \textbf{Best in bold.}}
\label{table:ablation_loss}
\begin{tabular}{|p{2cm}|p{1cm}|p{1cm}|p{1cm}|p{1cm}|} 
\hline
\multirow{2}{*}{VIGOR, val.} & \multicolumn{2}{c|}{$\downarrow$ Localization (m)} & \multicolumn{2}{c|}{$\uparrow$ P@GT (\num{e-3})}\\
\cline{2-5}
 & mean  & median & mean  &median \\
\hline
Cross-entropy & \textbf{3.63} & 1.42 & 1.48 & \textbf{1.00} \\
\hline
Wasserstein & 3.75 & \textbf{1.41} & \textbf{3.97} & 0.00\\
\hline
\end{tabular}
\end{table}

The model trained with cross-entropy loss has a lower mean localization error than the model trained with Wasserstein distance-based loss.
Notably, the model trained with Wasserstein distance-based loss outputs localization distributions that are very sharp.
Biased by a few accurate predictions, the mean probability at the ground truth pixel of this model is higher than that of the model trained with cross-entropy.
However, the median probability at the ground truth pixel is near zero, indicating many of the ground truth locations receive little probability mass.
In temporal filtering or multi-sensor fusion, fusing such predictions might make the system miss the ground truth location.
Besides, we also observe that training with Wasserstein distance-based loss makes the output distribution less indicative of localization and orientation errors.
This reduces its practicality in safety crucial applications where the outliers in prediction should be filtered out.
Thus, we used the cross-entropy loss as our localization loss $\mathcal{L}_D$.

\subsection{\newtext{Runtime analysis}}
\newtext{
First, we study how the proposed Rolling \& Matching influences the runtime of our method.
On the VIGOR dataset, when increasing the number of orientation bins ($R=2, 4, 5, 10, 20$) for Rolling \& Matching in all LMU and OMU modules, the inference speed of our method decreases slightly (18, 17, 17, 17, 15 FPS).
Since the Rolling \& Matching is a convolution process between ground and aerial descriptors, it can be done efficiently.
}

\newtext{
Next, we compare the runtime of our method to that of previous state-of-the-art methods.
To include more baselines, the comparison is done on the KITTI dataset.
On the same device (a single V100 GPU), our method takes 0.042s to process a pair of input images (24 FPS) on the KITTI dataset, which is slower than SliceMatch's 156 FPS~\cite{lentsch2022slicematch} but faster than LM's 0.59 FPS~\cite{shi2022beyond}. 
Importantly, even though SliceMatch runs faster, CCVPE is considerably more accurate in localization.
Note that the authors of~\cite{fervers2022uncertainty} evaluated the runtime of their method on the KITTI-360 dataset with a more advanced GPU (RTX6000), and their method runs at approximately 2-3 Hz~\cite{fervers2022uncertainty}, which is slower than CCVPE.
}

%% file: 5-conclusion.tex
\section{Conclusion}
In this work, the novel Convolutional Cross-View Pose Estimation method (CCVPE) was proposed.
CCVPE exploits the strength of a translational equivariant feature encoder and of contrastive learning to learn orientation-aware descriptors for joint localization and orientation estimation.
Instead of estimating a single location, its Localization Decoder outputs a multi-modal distribution to capture the underlying localization uncertainty.
The Localization Matching Upsampling (LMU) and Orientation Matching Upsampling (OMU) modules were devised to summarize orientation invariant localization cues and orientation-dependent information from the descriptor matching result when upsampling the aerial feature maps inside two separate decoders.
The Orientation Decoder outputs a dense orientation vector field that is conditioned on the localization distribution. 
Thus, CCVPE's orientation prediction becomes multi-modal when there are multiple modes in the localization distribution.

CCVPE achieves 72\% and 36\% lower median localization errors (1.42~m and 3.47~m) than the previous SOTA (5.07~m and 5.41~m) on the VIGOR and KITTI datasets, and it has comparable orientation estimation accuracy.
Importantly, CCVPE can work with ground images with different horizontal FoVs and incorporate an orientation prior to improve the localization without re-training.
Its probabilistic output can be used to filter out predictions that potentially have large localization and orientation errors, yielding better practicality than the baselines that do not have a probability estimate.
We demonstrated on traversals collected at different times in the Oxford RobotCar dataset that CCVPE can estimate the pose of ego-vehicle at $14$ FPS with a median lateral and longitudinal error below $1$ meter and a median orientation error around $1^\circ$, bringing cross-view pose estimation methods closer to the requirement of autonomous driving of $<0.3$ m lateral and longitudinal localization accuracy.
Future work will address applying temporal filtering on the single frame estimates and multi-sensor fusion to further increase the pose estimation accuracy.
Besides, more efficient descriptor construction and matching strategies will be studied.